\pdfoutput=1

\documentclass[11pt]{article}

\usepackage{acl}

\usepackage{times}
\usepackage{latexsym}

\usepackage{graphicx}
\usepackage{subcaption}
\usepackage{xcolor}
\usepackage{amsmath}
\usepackage{amssymb}
\usepackage{booktabs}
\usepackage{makecell}
\usepackage{cleveref}

\usepackage[T1]{fontenc}
\usepackage[utf8]{inputenc}

\usepackage{microtype}

\usepackage{todonotes}
\makeatletter
\newcommand*\iftodonotes{\if@todonotes@disabled\expandafter\@secondoftwo\else\expandafter\@firstoftwo\fi}  %
\makeatother

\usepackage{colortbl}%
\usepackage[normalem]{ulem}

\title{Weaker Than You Think: A Critical Look at Weakly Supervised Learning}

\author{Dawei Zhu$^{1}$ \, Xiaoyu Shen$^{2}$\thanks{\hspace{1.5 mm}Work done outside Amazon.} \, Marius Mosbach$^{1}$ \, Andreas Stephan$^{3}$ \, Dietrich Klakow$^{1}$ \\
$^1$Saarland University, Saarland Informatics Campus\\
$^2$Amazon Alexa AI\\
$^3$University of Vienna\\
\texttt{dzhu@lsv.uni-saarland.de}}

\begin{document}
\maketitle
\begin{abstract}
Weakly supervised learning is a popular approach for training machine learning models in low-resource settings. Instead of requesting high-quality yet costly human annotations, it allows training models with noisy annotations obtained from various weak sources. Recently, many sophisticated approaches have been proposed for robust training under label noise, reporting impressive results. In this paper, we revisit the setup of these approaches and find that \emph{the benefits brought by these approaches are significantly overestimated}. Specifically, we find that the success of existing weakly supervised learning approaches heavily relies on the availability of clean validation samples which, as we show, can be leveraged much more efficiently by simply training on them. 
After using these clean labels in training, the advantages of using these sophisticated approaches are mostly wiped out. This remains true even when reducing the size of the available clean data to just five samples per class, making these approaches impractical. To understand the true value of weakly supervised learning, we thoroughly analyze diverse NLP datasets and tasks to ascertain when and why weakly supervised approaches work. Based on our findings, we provide recommendations for future research.\footnote{Our code is available at: \url{https://github.com/uds-lsv/critical_wsl}}

\end{abstract}

\section{Introduction}

Weakly supervised learning (WSL) is one of the most popular approaches for alleviating the annotation bottleneck in machine learning. Instead of collecting expensive clean annotations, it leverages weak labels from various weak labeling sources such as heuristic rules, knowledge bases or lower-quality crowdsourcing~\cite{ratner2017_snorkel}. These weak labels are inexpensive to obtain, but are often noisy and inherit biases from their sources. Deep learning models trained on such noisy data without regularization can easily overfit to the noisy labels~\cite{zhang2017_rethinking, tanzer2022memorisation}. 
Many advanced WSL techniques have recently been proposed to combat the noise in weak labels, and significant progress has been reported. On certain datasets, they even manage to match the performance of fully-supervised models~\cite{Liang2020_bond, ren2020_denoising, Yu2021_cosine}.

\begin{figure}[t]
    \centering
    \includegraphics[width=\columnwidth]{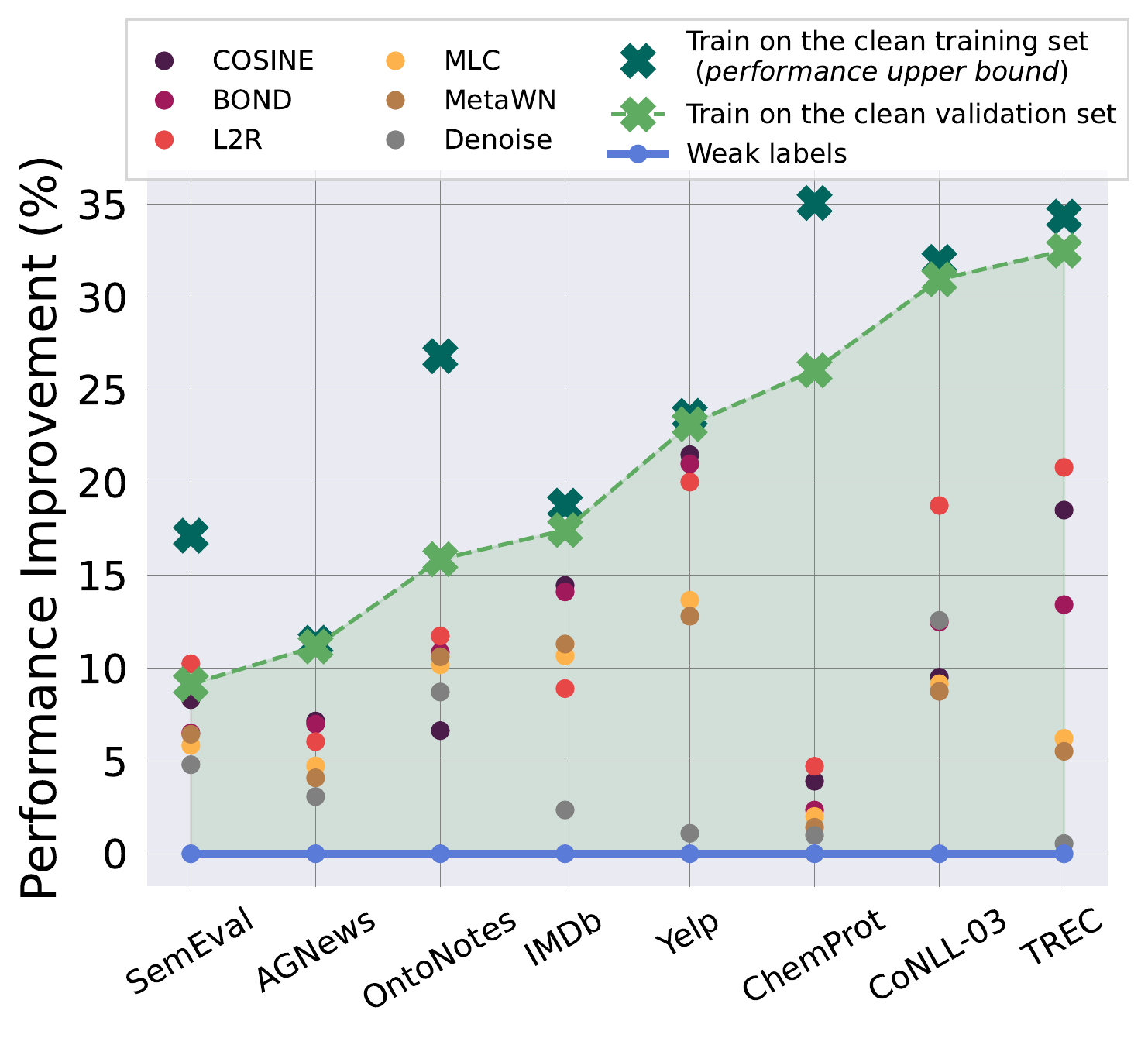}
    \caption{\textbf{Performance improvement over weak labels on the test sets.} Each point represents the average performance improvement of one approach over five runs. On various NLP datasets, weakly supervised methods (dots) outperform weak labels (blue line) on the test sets. However, \emph{simply fine-tuning on the available clean validation data (light green crosses) outperforms all sophisticated weakly supervised methods in almost all cases}. See Appendix \ref{sec:appendix:training_on_the_full_validation_sets} for experimental details.
    }
    \label{fig:train_on_full_clean_validation}
\end{figure}

In this paper, we take a close look at the claimed advances of these WSL approaches and find that the \emph{benefits of using them are significantly overestimated}. Although they appear to require only weak labels during training, a substantial number of clean validation samples are used for various purposes such as early-stopping \cite{Liang2020_bond, Yu2021_cosine} and meta-learning \cite{ren2018learning, shu2019meta, zheng2021meta}. We cast doubt on this practice: in real-world applications, these clean validation samples could have instead been used for training. To address our concern, we explore fine-tuning models directly on the validation splits of eight datasets provided by the WRENCH benchmark \cite{zhang2021_wrench} and compare it to recent WSL algorithms. The results are shown in Figure \ref{fig:train_on_full_clean_validation}. Interestingly, although all WSL models generalize better than the weak labels, \textbf{simply fine-tuning on the validation splits outperforms all WSL methods in almost all cases}, sometimes even by a large margin. This suggests that existing WSL approaches are not evaluated in a realistic setting and the claimed advances of these approaches may be overoptimistic. In order to determine the true benefits of WSL approaches in a realistic setting, we conduct extensive experiments to investigate the role of clean validation data in WSL. Our findings can be summarized as follows:
\begin{itemize}
  \item Without access to any clean validation samples, all WSL approaches considered in this paper \emph{fail to} work, performing similarly to or worse than the weak labels (\S\ref{sec:clean_data_is_necessary}).
  \item Although increasing the amount of clean validation samples improves WSL performance (\S\ref{sec:how_much_clean_data_does_wsl_need}), these validation samples can be more efficiently leveraged by directly training on them, which can outperform WSL approaches when there are more than 10 samples per class for most datasets (\S\ref{sec:little_clean_data_for_peft}).
  \item Even when enabling WSL models to continue training on clean validation samples, they can barely beat an embarrassingly simple baseline which directly fine-tunes on weak labels followed by fine-tuning on clean samples. This stays true with as few as 5 samples per class (\S\ref{sec:cft}).
  \item The knowledge encoded in pre-trained language models biases them to seek linguistic correlations rather than shallow rules from the weak labels; further fine-tuning the pre-trained language models with contradicting examples helps reduce biases from weak labels (\S\ref{sec:cft_ablation}).
\end{itemize}

\indent Altogether, we show that existing WSL approaches significantly overestimate their benefits in a realistic setting. We suggest future work to (1) fully leverage the available clean samples instead of only using them for validation and (2) consider the simple baselines discussed in this work when comparing WSL approaches to better understand WSL's true benefits.

\section{Related work}

\paragraph{Weak supervision. }
Weak supervision is proposed to ease the annotation bottleneck in training machine learning models. It uses weak sources to automatically annotate the data, making it possible to obtain a large amount of annotated data at a low cost. A comprehensive survey is done in \citet{zhang2022_weak_survey}. \citet{ratner2017_snorkel} propose to label data programmatically using heuristics such as keywords, regular expressions or knowledge bases.
One drawback of weak supervision is that its annotations are noisy, i.e., some annotations are incorrect. Training models on such noisy data may result in poor generalization \cite{zhang2017_rethinking, tanzer2022memorisation, zhang2022_weak_survey}.
One option to counter the impact of wrongly labeled samples is to re-weight the impact of examples in loss computation \citep{ren2018learning,shu2019meta, zheng2021meta}. Another line of research leverages the knowledge encoded in pre-trained language models \cite{ren2020_denoising, Jiang2021_needle, Stephan2022}. 
Methods such as BOND \citep{Liang2020_bond}, ASTRA \cite{karamanolakis2021astra} and COSINE \citep{Yu2021_cosine} apply teacher-student frameworks to train noise-robust models. \citet{Zhu2023} show that teacher-student frameworks may still be fragile in challenging situations and propose incorporating meta-learning techniques in such cases.
Multiple benchmarks are available to evaluate weak supervision systems, e.g., WRENCH \citep{zhang2021_wrench}, Skweak \citep{lison-etal-2021-skweak}, and WALNUT \citep{Zheng2022_walnut}. In this paper, we take representative datasets from WRENCH and reevaluate existing WSL approaches in more realistic settings.

\paragraph{Realistic evaluation.} Certain pitfalls have been identified when evaluating machine learning models developed for low-resource situations. Earlier work in semi-supervised learning (SSL) in computer vision, for example, often trains with a few hundred training examples while retaining thousands of validation samples for model selection \cite{tarvainen2017mean, Miyato2018}. \citet{oliver2018realistic} criticize this setting and provide specific guidance for realistic SSL evaluation. Recent work in SSL has been adapted to discard the validation set and use a fixed set of hyperparameters across datasets \cite{xie2020unsupervised, zhang2021flexmatch, li2021comatch}. In NLP, it has been shown that certain (prompt-based) few-shot learning approaches are sensitive to prompt selection which requires separate validation samples \cite{perez2021true}. This defeats the purported goal of few-shot learning, which is to achieve high performance even when collecting additional data is prohibitive. Recent few-shot learning algorithms and benchmarks have adapted to a more realistic setting in which fine-grained model selection is either skipped \cite{gao2021making, alex2021raft, bragg2021flex, schick2022true, lu2022fantastically} or the number of validation samples are strictly controlled \cite{bragg2021flex, zheng2021fewnlu}. To our knowledge, no similar work exists exploring the aforementioned problems in the context of weak supervision. This motivates our work.

\section{Overall setup}
\label{sec:overall_setp}
\paragraph{Problem formulation.} Formally, let $\mathcal{X}$ and $\mathcal{Y}$ be the feature and label space, respectively. In standard supervised learning, we have access to a training set $D=\{(x_i,y_i )\}_{i=1}^{N}$ sampled from a clean data distribution $\mathcal{D}_{c}$ of random variables $(X,Y)\in \mathcal{X} \times \mathcal{Y}$. In weak supervision, we are instead given a weakly labeled dataset $D_{w} = \{(x_i, \hat{y}_{i})\}_{i=1}^{N}$ sampled from a noisy distribution $\mathcal{D}_{n}$, where $\hat{y}_{i}$ represents labels obtained from weak labeling sources such as heuristic rules or crowd-sourcing.\footnote{Majority voting can be used to resolve conflicting weak labels from different labeling sources.} $\hat{y}_{i}$ is noisy, i.e., it may be different from the ground-truth label $y_{i}$. The goal of WSL algorithms is to \emph{obtain a model that generalizes well on $D_{test} \sim \mathcal{D}_{c}$ despite being trained on $D_{w}\sim \mathcal{D}_{n}$}. In recent WSL work, a set of clean samples, $D_{v} \sim \mathcal{D}_{c}$, is also often included for model selection.\footnote{We refer to model selection as the process of finding the best set of hyperparameters via a validation set, including the optimal early-stopping time. Prior work has shown that early-stopping is crucial for learning with noisy labels \cite{arpit2017closer, Yu2021_cosine, zhu2022bert, tanzer2022memorisation}.}

\paragraph{Datasets.} We experiment with eight datasets covering different NLP tasks in English. Concretely, we include four text classification datasets: (1) AGNews~\cite{zhang2015character}, (2) IMDb~\cite{maas2011_imdb}, (3) Yelp~\cite{zhang2015character}, (4) TREC~\cite{li2002_trec}, two relation classification datasets: (5) SemEval~\cite{hendrickx2010_semeval} and (6) ChemProt~\cite{krallinger2017_chemprot}, and two Named-Entity Recognition (NER) datasets: (7) CoNLL-03~\cite{tjong2003_conll} and (8) OntoNotes~\cite{pradhan2013_ontonotes}. The weak annotations are obtained from WRENCH \cite{zhang2021_wrench}. Table \ref{tab:dataset} summarizes the basic statistics of the datasets.

\begin{table}[h!]
\centering
	\resizebox{\columnwidth}{!}{
		\begin{tabular}{llcccccc}
			\toprule \bf Dataset & \bf Task & \bf \# Class  &\bf \# Train & \bf \# Val & \bf \# Test \\ \midrule
            AGNews &Topic &4 &96K & 12K & 12K  \\ %
            IMDb &Sentiment &2 &20K & 2.5K & 2.5K \\ %
            Yelp &Sentiment& 2 &30K & 3.8K & 3.8K  \\ %
            TREC & Question & 6 & 4,965 & 500 & 500   \\ %
            SemEval & Relation & 9 & 1,749 & 178 & 600     \\
            ChemProt & Relation & 10 &  13K & 1.6K &  1.6K  \\
            CoNLL-03 & NER & 4 & 14K & 3.2K & 3.4K     \\
            OntoNotes 5.0 & NER & 18 &  115K & 5K &  23K  \\
            \bottomrule
		\end{tabular}
	}%
	\caption{\textbf{Dataset statistics.} Additional details on datasets are provided in Appendix \ref{sec:appendix:dataset_details}.
    }
	\label{tab:dataset}

\end{table}

\paragraph{WSL baselines.} We analyze popular WSL approaches including: (1) \textbf{FT\textsubscript{W}} represents the standard fine-tuning approach\footnote{We use the subscript ``W'' to emphasize that this fine-tuning is done on the weakly annotated data and to distinguish it from the fine-tuning experiments in Section \ref{sec:little_clean_data_for_peft} which are done on clean data.} \cite{Howard2018, devlin2019bert}. \citet{ren2020_denoising}, \citet{zhang2021_wrench} and \citet{Zheng2022_walnut} show that a pre-trained language model (PLM) fine-tuned on a weakly labeled dataset often generalizes better than the weak labels synthesized by weak labeling sources. (2) \textbf{L2R} \citep{ren2018learning} uses meta-learning to determine the optimal weights for each (noisy) training sample so that the model performs best on the (clean) validation set. Although this method was originally proposed to tackle artificial label noise, we find it performs on par with or better than recent weak supervision algorithms on a range of datasets. (3) \textbf{MLC} \cite{zheng2021meta} uses meta-learning as well, but instead of weighting the noisy labels, it uses the meta-model to correct them. The classifier is then trained on the corrected labels. (4) \textbf{BOND} \cite{Liang2020_bond} is a noise-aware self-training framework designed for learning with weak annotations. (5) \textbf{COSINE} \cite{Yu2021_cosine} underpins self-training with contrastive regularization to improve noise robustness further and achieves state-of-the-art performance on the WRENCH \cite{zhang2021_wrench} benchmark.

To provide a fair comparison, we use RoBERTa-base \cite{liu2019roberta} as the common backbone PLM for all WSL approaches (re)implemented in this paper.

\section{Is clean data necessary for WSL?}
\label{sec:clean_data_is_necessary}

Recent best-performing WSL approaches rely on a clean validation set for model selection. Figure~\ref{fig:train_on_full_clean_validation} reveals that they fail to outperform a simple model that is directly fine-tuned on the validation set. Therefore, a natural question to ask is: ``\emph{Will WSL still work without accessing the clean validation set?}”. If the answer is yes, then we can truly reduce the burden of data annotation and the benefits of these WSL approaches would be undisputed. This section aims to answer this question.

\begin{figure}[]
    \centering
    \includegraphics[width=\columnwidth]{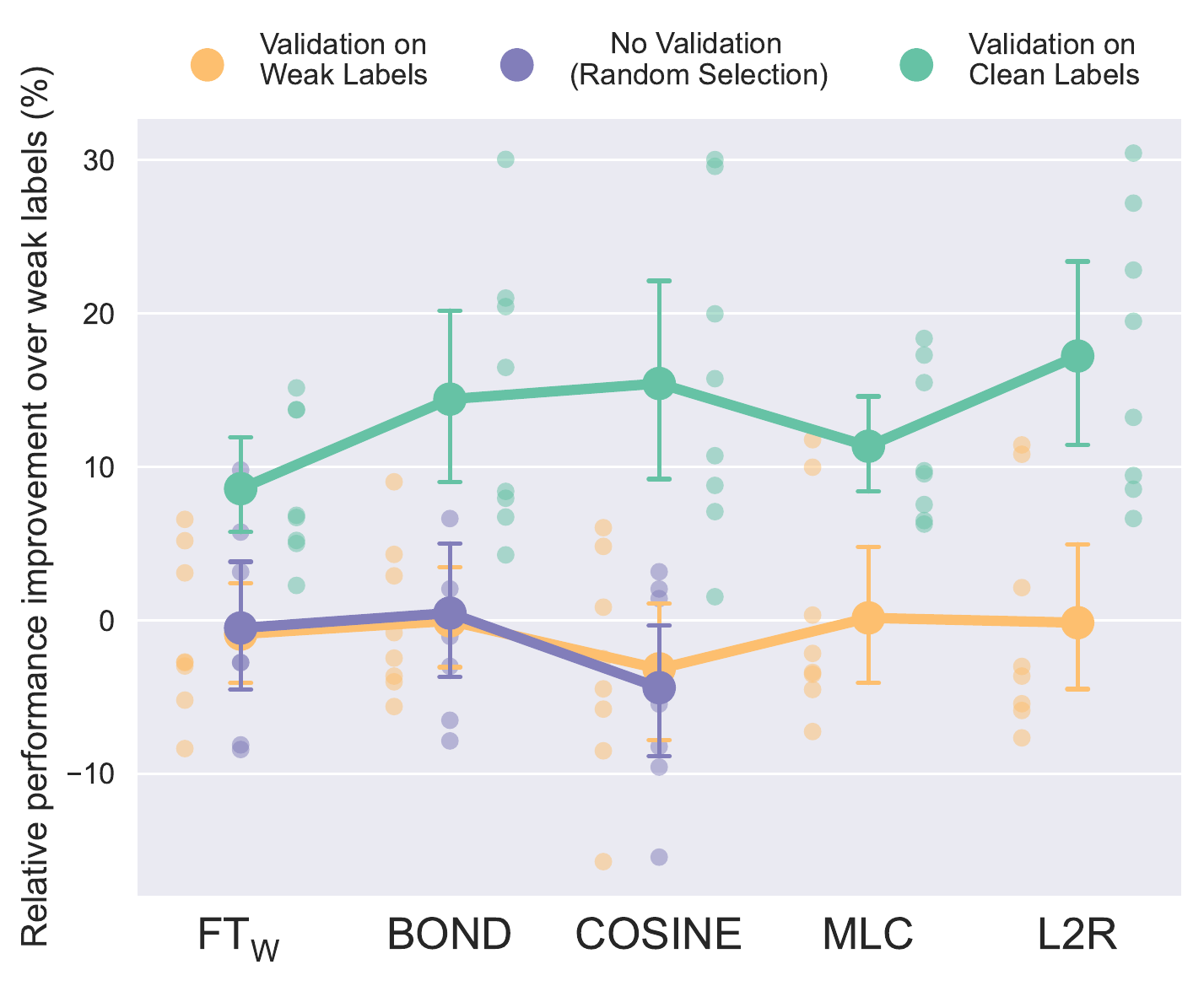}
    \caption{\textbf{Relative performance gain over weak labels when varying validation conditions.} The dots show the average performance gain across 5 runs for each of the 8 datasets. The curves show the average gain across datasets. WSL baselines achieve noticeable performance gains only if a clean validation set is used. Performing model selection on a weakly labeled validation set does not help generalization. Note that L2R and MLC are not applicable without validation data.}
    \label{fig:zcd_fine_tuning}
\end{figure}

\paragraph{Setup.} We compare three different validation choices for model selection using either (1) a clean validation set from $D_v$ as in prior work, (2) weak labels from $\tilde{D}_v$ obtained by annotating the validation set via weak labeling sources (the same procedure used to construct training annotations), or (3) no validation data at all. In the last setting, we randomly select 5 sets of hyperparameters from our search space (see Appendix~\ref{sec:appendix:implementation_details}). We run the WSL approaches introduced in Section~\ref{sec:overall_setp} on all eight datasets with different validation choices and measure their test performance. Each experiment is repeated 5 times with different seeds.

While one may expect a certain drop in performance when switching from $D_v$ to $\tilde{D}_v$, the absolute performance of a model does not determine the usefulness of a WSL method. We are more interested in whether a trained model generalizes better than the weak labels.\footnote{Weak labeling sources are typically applied to the training data to synthesize a weakly annotated training set. However, it is also possible to synthesize the weak labels for the test set following the same procedure and measure their performance. In other words, weak labeling sources can be regarded as the most basic classification model, and the synthesized weak labels are its predictions.} In realistic applications, it is only worth deploying trained models if they demonstrate clear advantages over the weak labels. Therefore, we report the relative performance gain of WSL approaches over the weak labels. Formally, let $P_{WL}, P_{\alpha}$ denote the performance (accuracy, F1-score, etc.) achieved by the weak labels and a certain WSL method $\alpha$, respectively. The the relative performance gain is defined as $G_{\alpha} = (P_{\alpha}-P_{WL})/P_{WL}$. We consider a WSL approach to be \emph{effective} and practically useful only if $G_{\alpha}>0$.

\begin{figure*}[t]
        \centering
        
     \begin{subfigure}[b]{0.5\columnwidth}
         \centering
         \includegraphics[width=\columnwidth]{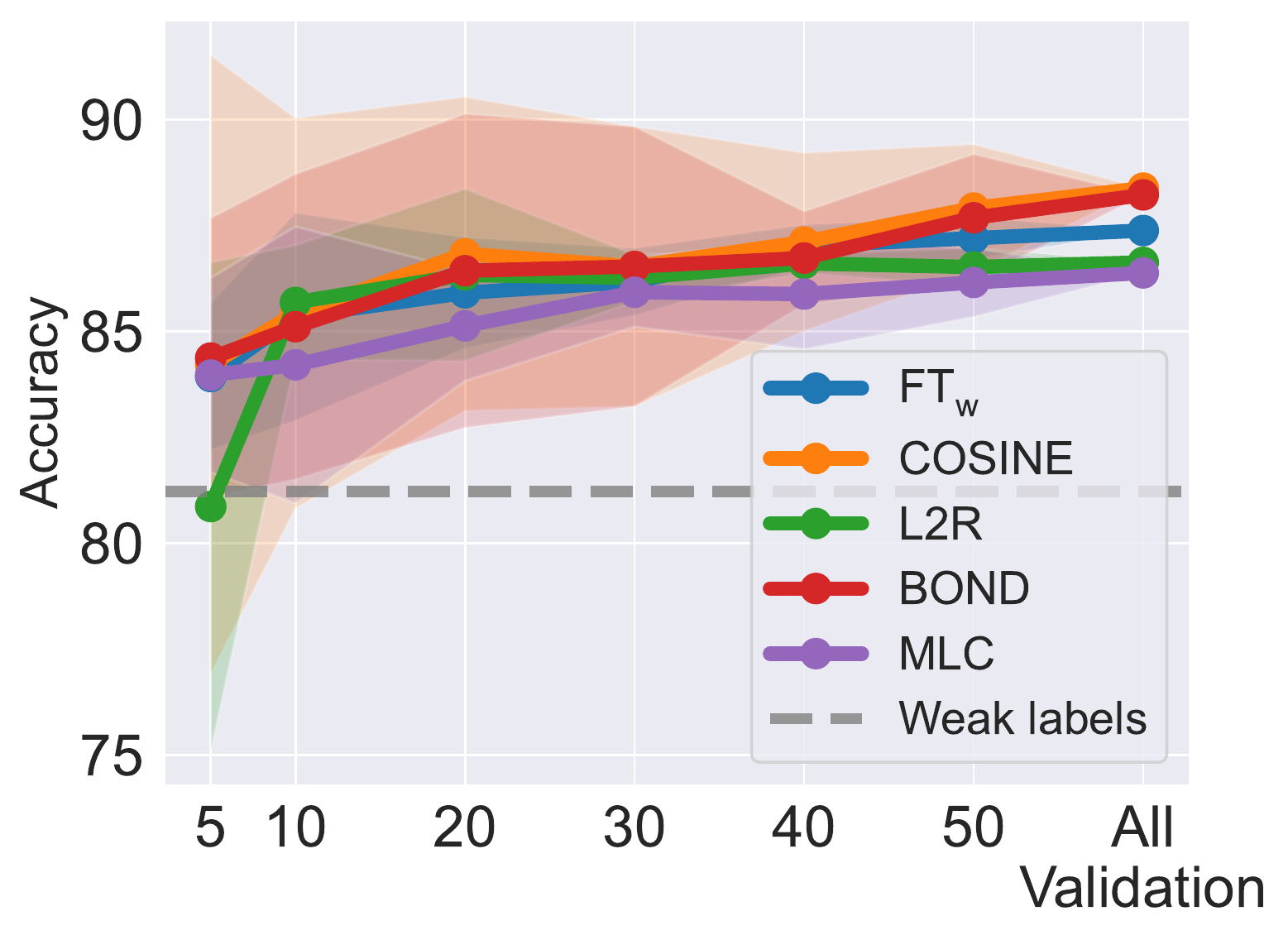}
         \caption{AGNews}
     \end{subfigure}\hfill
          \begin{subfigure}[b]{0.5\columnwidth}
         \centering
         \includegraphics[width=\columnwidth]{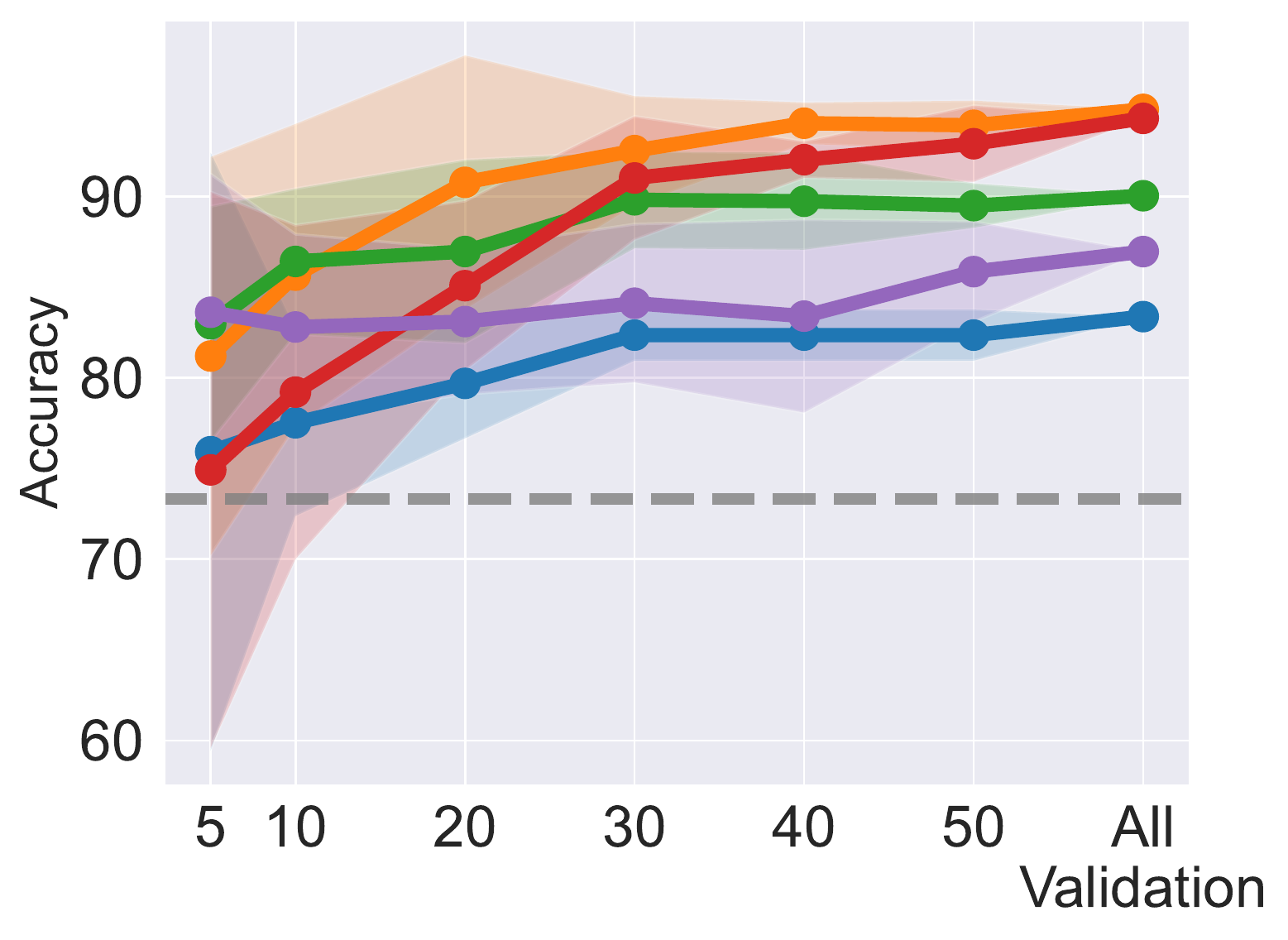}
         \caption{Yelp}

     \end{subfigure}\hfill
          \begin{subfigure}[b]{0.5\columnwidth}
         \centering
         \includegraphics[width=\columnwidth]{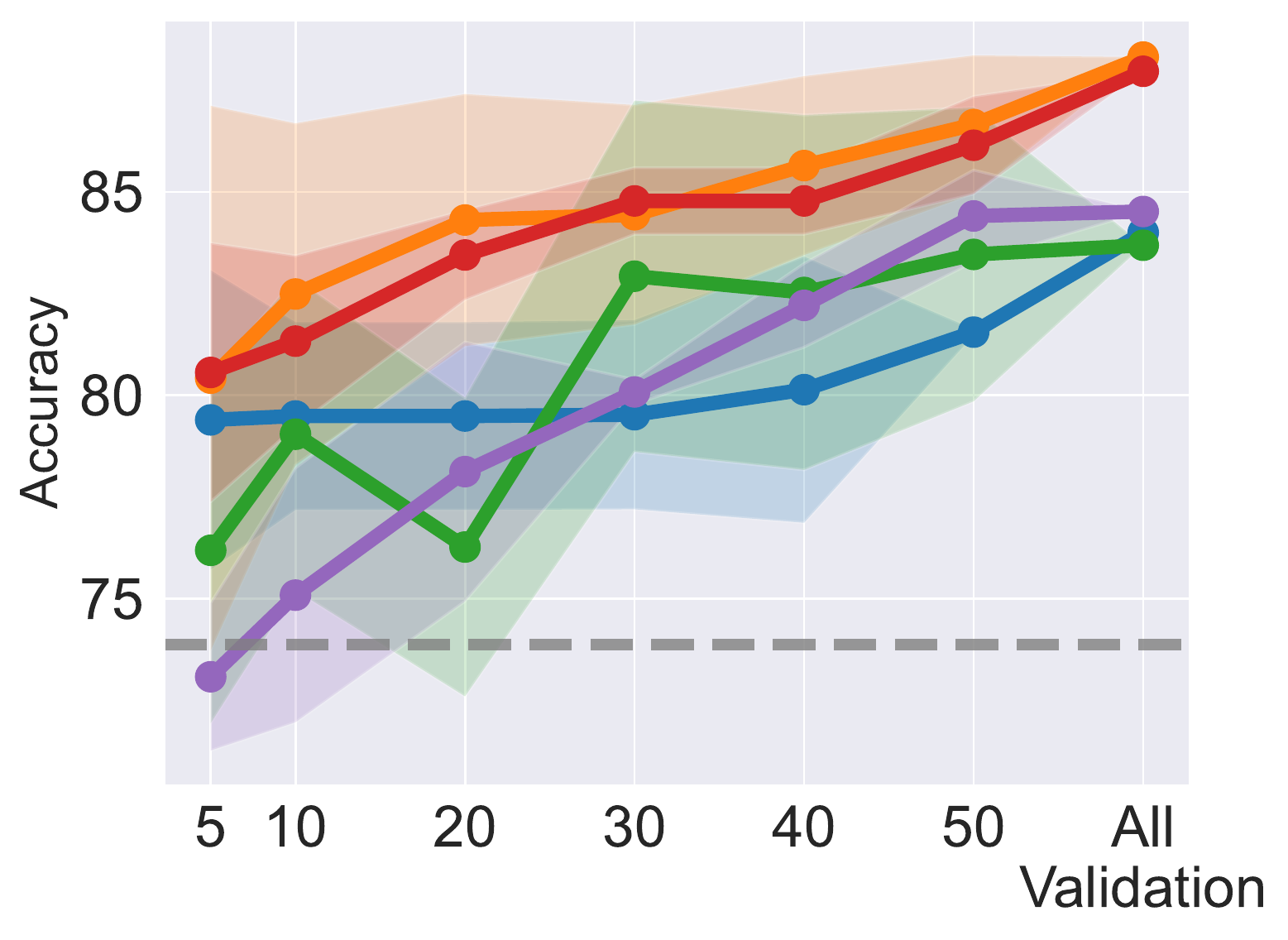}
         \caption{IMDb}

     \end{subfigure}\hfill
          \begin{subfigure}[b]{0.5\columnwidth}
         \centering
         \includegraphics[width=\columnwidth]{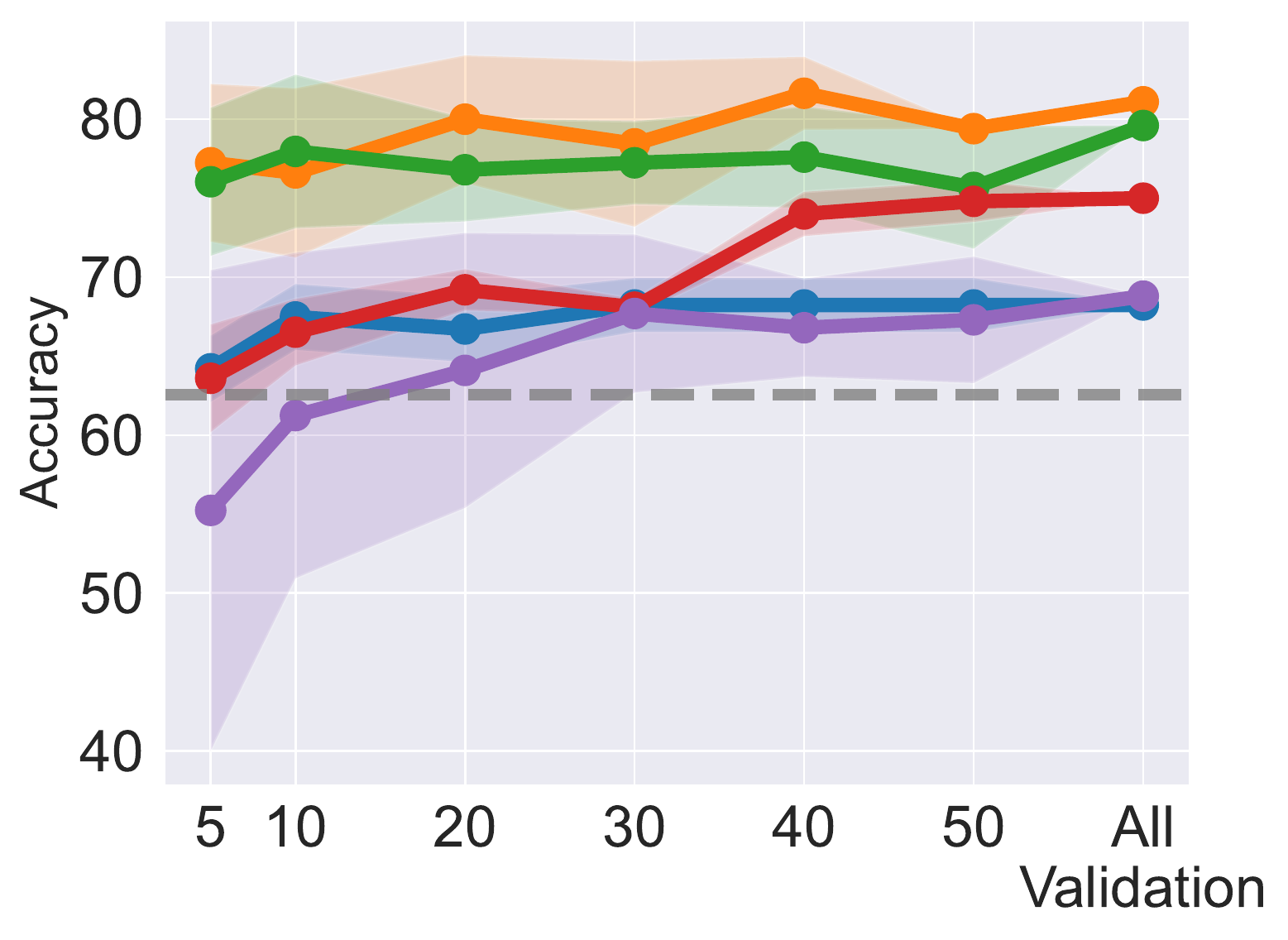}
         \caption{TREC}

     \end{subfigure}\hfill
     \begin{subfigure}[b]{0.5\columnwidth}
     \centering
     \includegraphics[width=\columnwidth]{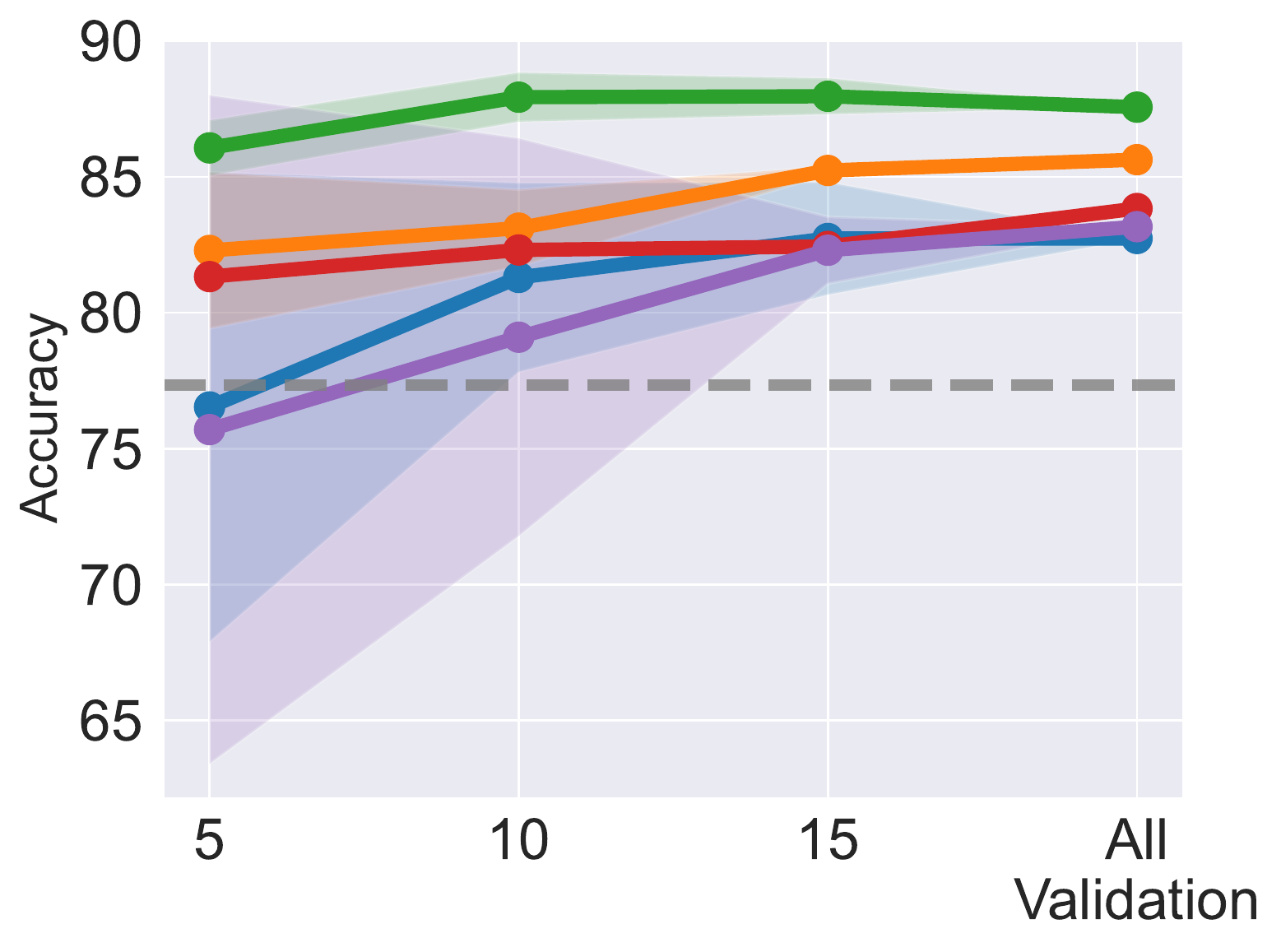}
     \caption{SemEval}

     \end{subfigure}\hfill
     \begin{subfigure}[b]{0.5\columnwidth}
     \centering
     \includegraphics[width=\columnwidth]{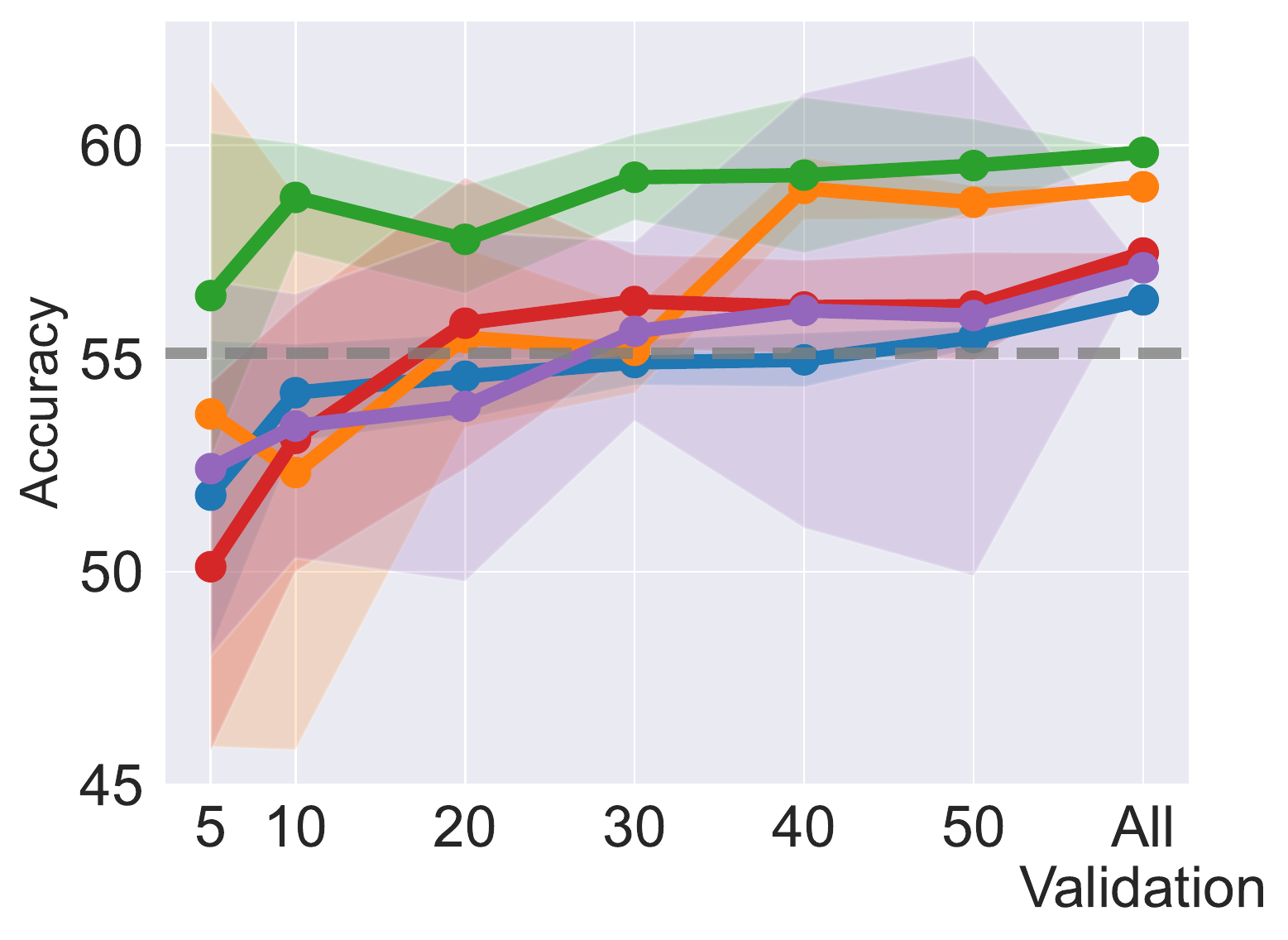}
     \caption{ChemProt}

     \end{subfigure}\hfill
     \begin{subfigure}[b]{0.5\columnwidth}
     \centering
     \includegraphics[width=\columnwidth]{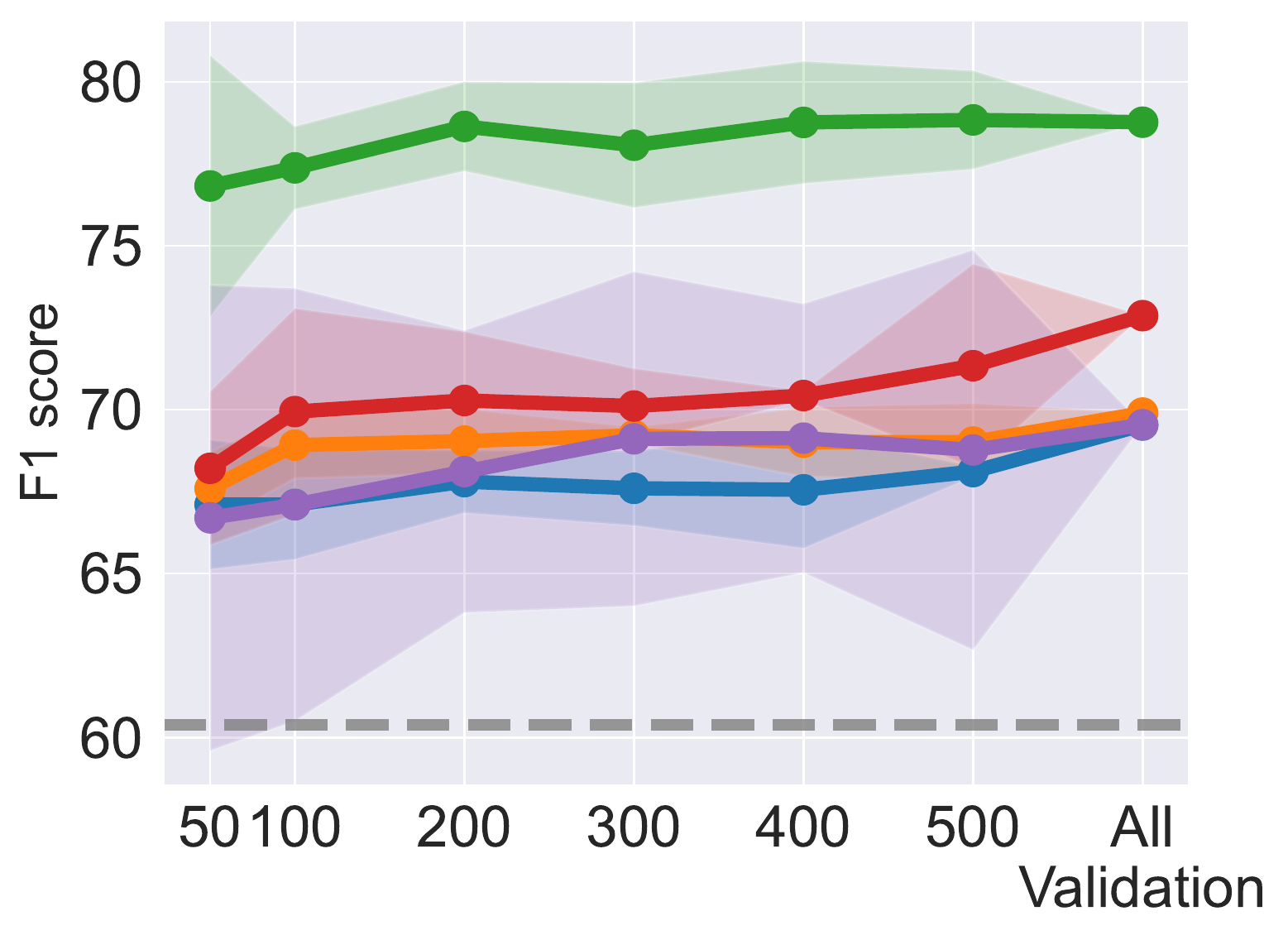}
     \caption{CoNLL-03}

     \end{subfigure}\hfill
      \begin{subfigure}[b]{0.5\columnwidth}
     \centering
     \includegraphics[width=\columnwidth]{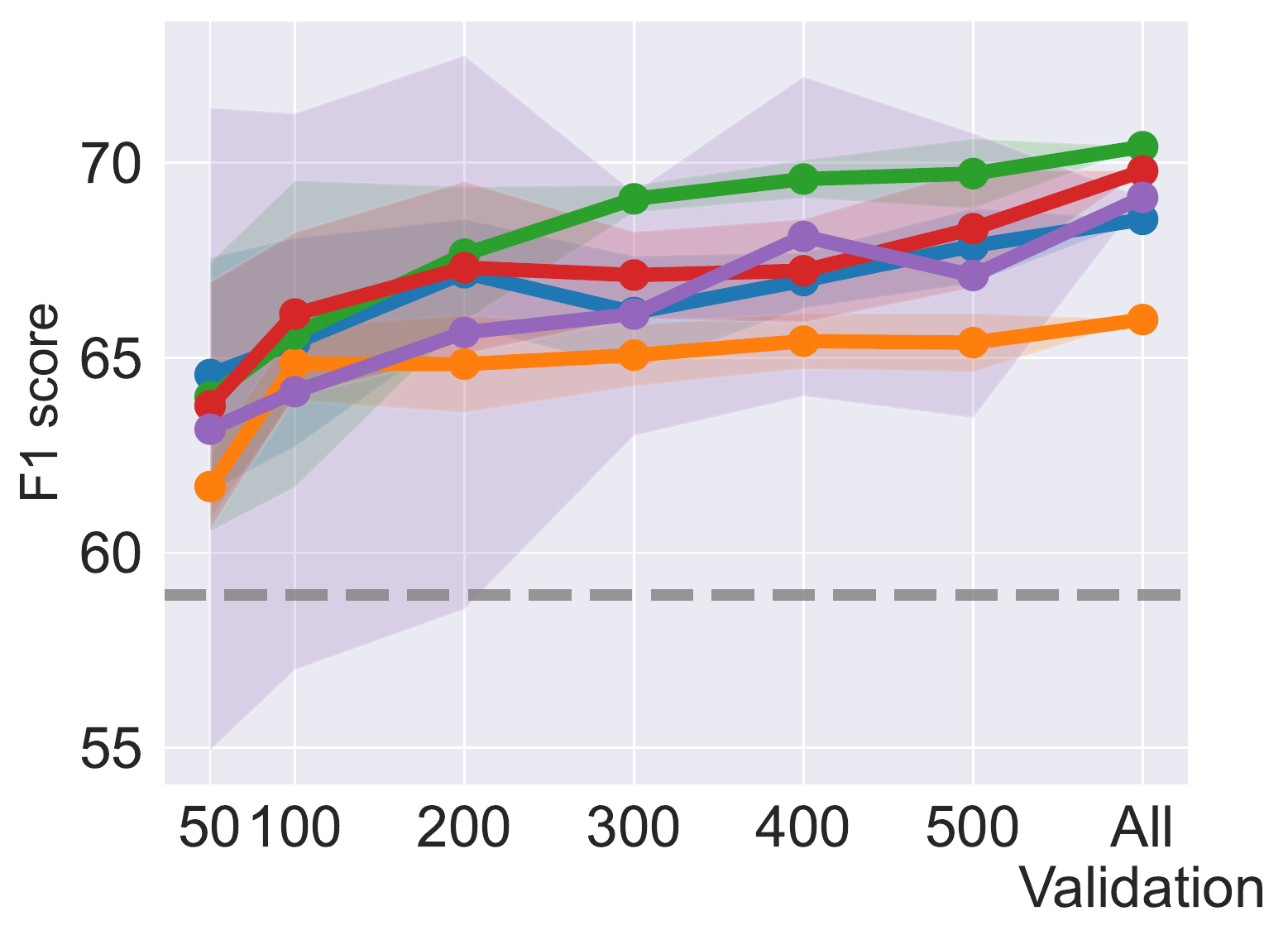}
     \caption{OntoNotes 5.0}

     \end{subfigure}\hfill
        \caption{\textbf{The impact of the number of clean validation samples on performance.}
        We plot average performance and standard deviation over 5 runs varying the size of the clean validation data. Whenever a small proportion of validation data is provided, most WSL techniques generalize better than the weak label baseline (grey dashed line). Performance improves with additional validation samples, but this tendency usually levels out with a moderate number of validation samples.
        }
        \label{fig:wls_small_validation}
\end{figure*}

\paragraph{Results.} Figure \ref{fig:zcd_fine_tuning} shows the relative performance gain for all considered WSL approaches. 
When model selection is performed on a clean validation set (green curve), all weak supervision baselines generalize better than the weak labels. Sophisticated methods like COSINE and L2R push the performance even further. This observation is consistent with previous findings \citep{zhang2021_wrench,Zheng2022_walnut}. However, when using a weakly labeled validation set (yellow curve), all WSL baselines become \textit{ineffective} and barely outperform the weak labels. More interestingly, models selected through the weakly labeled validation sets do not outperform models configured with random hyperparameters (purple curve). These results demonstrate that model selection on clean validation samples plays a vital role in the effectiveness of WSL methods. \textbf{Without clean validation samples, existing WSL approaches do not work.}

\section{How much clean data does WSL need?}
\label{sec:how_much_clean_data_does_wsl_need}
Now that we know clean samples are necessary for WSL approaches to work, a follow-up question would be: ``How many clean samples do we need?” Intuitively, we expect an improvement in performance as we increase the amount of clean data, but it is unclear how quickly this improvement starts to level off, i.e., we may find that a few dozen clean samples are enough for WSL approaches to perform model selection. The following section seeks to answer this question.

\paragraph{Setup.} We apply individual WSL approaches (see Section \ref{sec:overall_setp}) and vary the size of clean data sub-sampled from the original validation split. For text and relation classification tasks, we draw an increasing number of clean samples $N \in \{5, 10, 15, 20, 30, 40, 50\}$ per class when applicable.\footnote{The validation set of SemEval is too small to support $N>20$. Also, if a dataset is unbalanced, we randomly select $N \times C$ samples, where $C$ denotes the number of classes. This is a realistic sampling procedure when performing data annotation.} In the case of NER, as a sentence may contain multiple labels from different classes, selecting exactly $N$ samples per class at random is impractical. Hence, for NER we sample $N \in \{50, 100, 200, 300, 400, 500\}$ sentences for validation. For each $N$, we run the same experiment 5 times. Note that the clean data is \emph{used solely for model selection} in this set of experiments.

\begin{figure*}[t]
        \centering
        
     \begin{subfigure}[b]{0.5\columnwidth}
         \centering
         \includegraphics[width=\columnwidth]{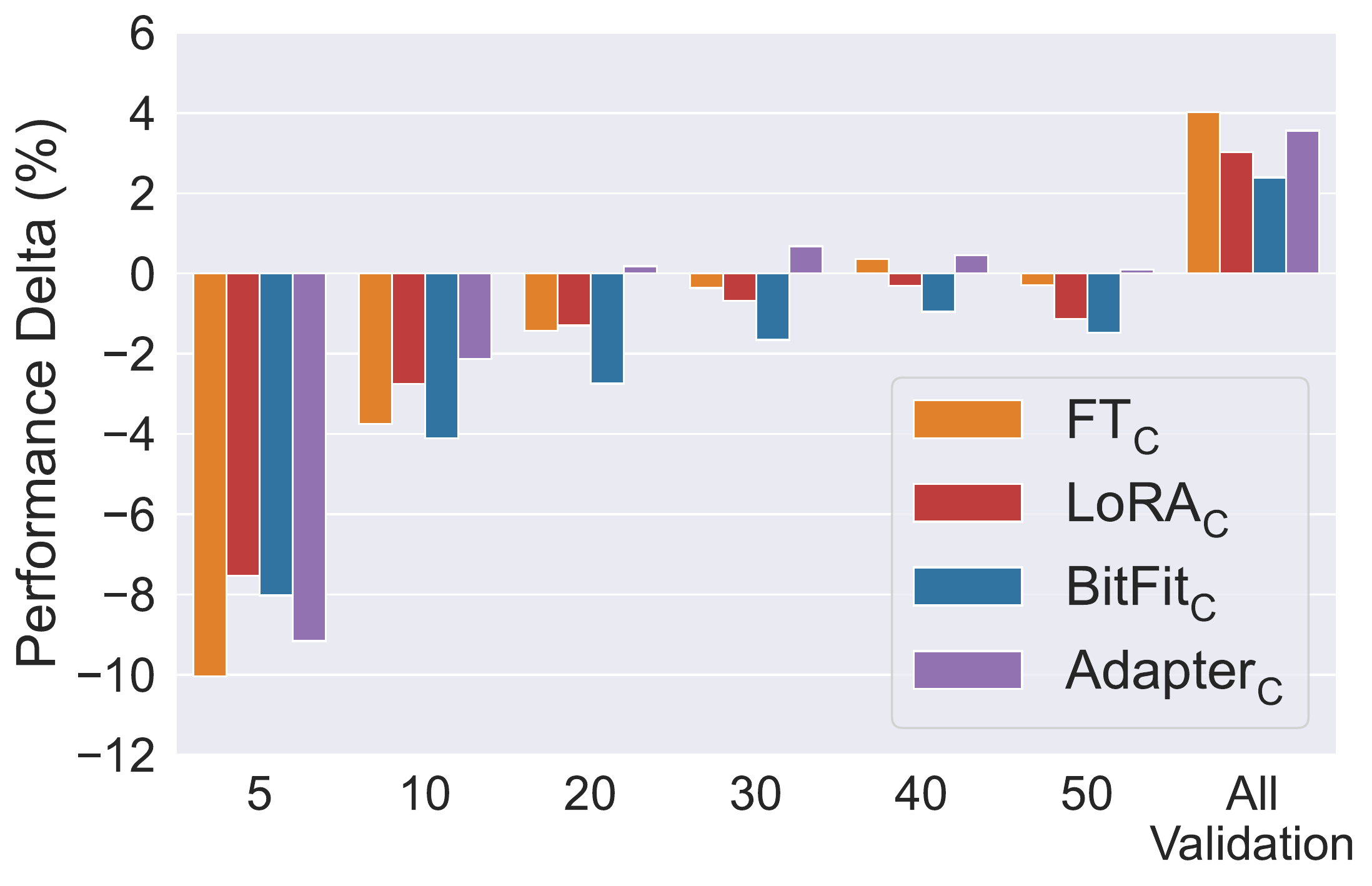}
         \caption{AGNews}
     \end{subfigure}\hfill
          \begin{subfigure}[b]{0.5\columnwidth}
         \centering
         \includegraphics[width=\columnwidth]{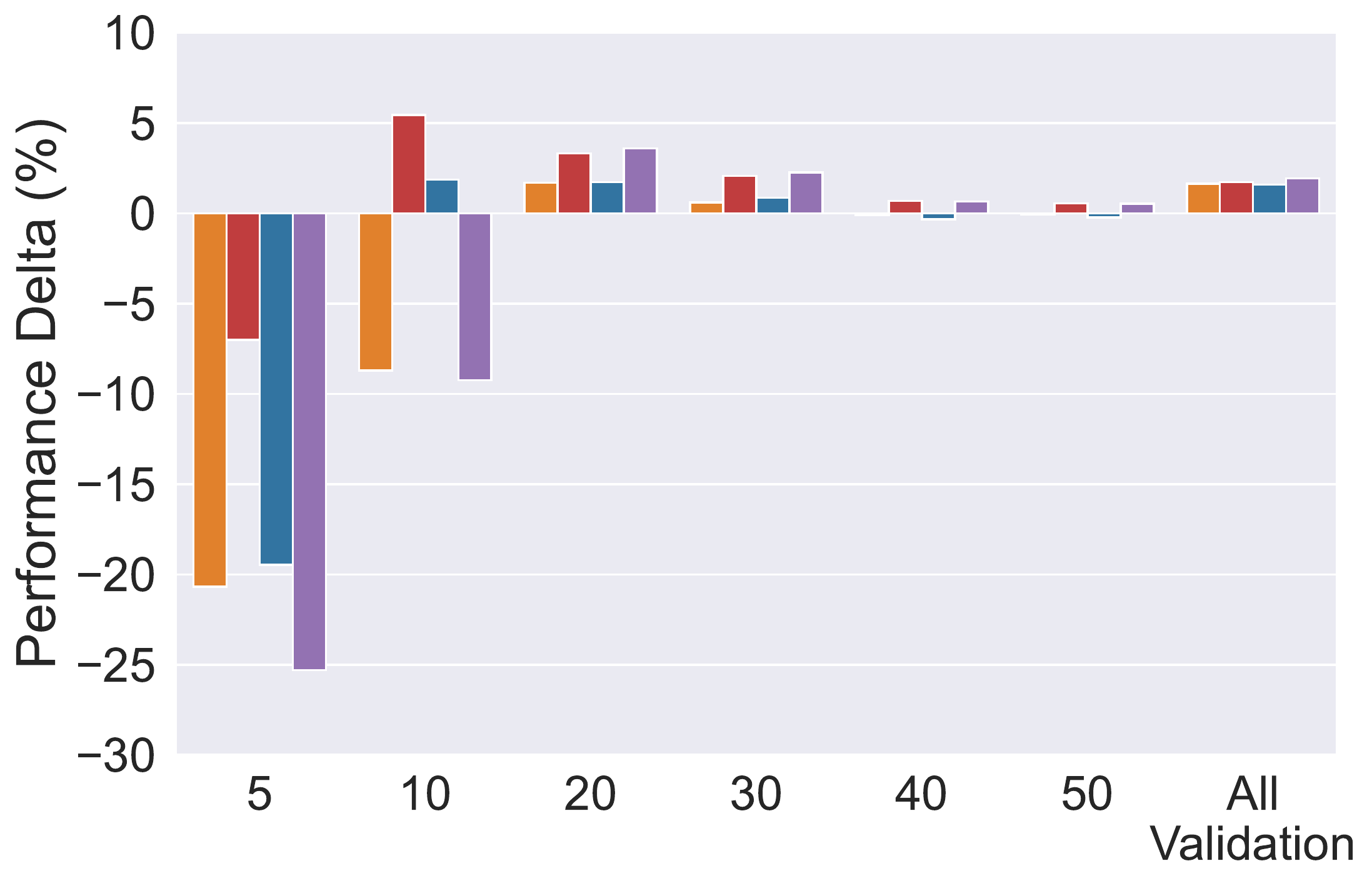}
         \caption{Yelp}

     \end{subfigure}\hfill
          \begin{subfigure}[b]{0.5\columnwidth}
         \centering
         \includegraphics[width=\columnwidth]{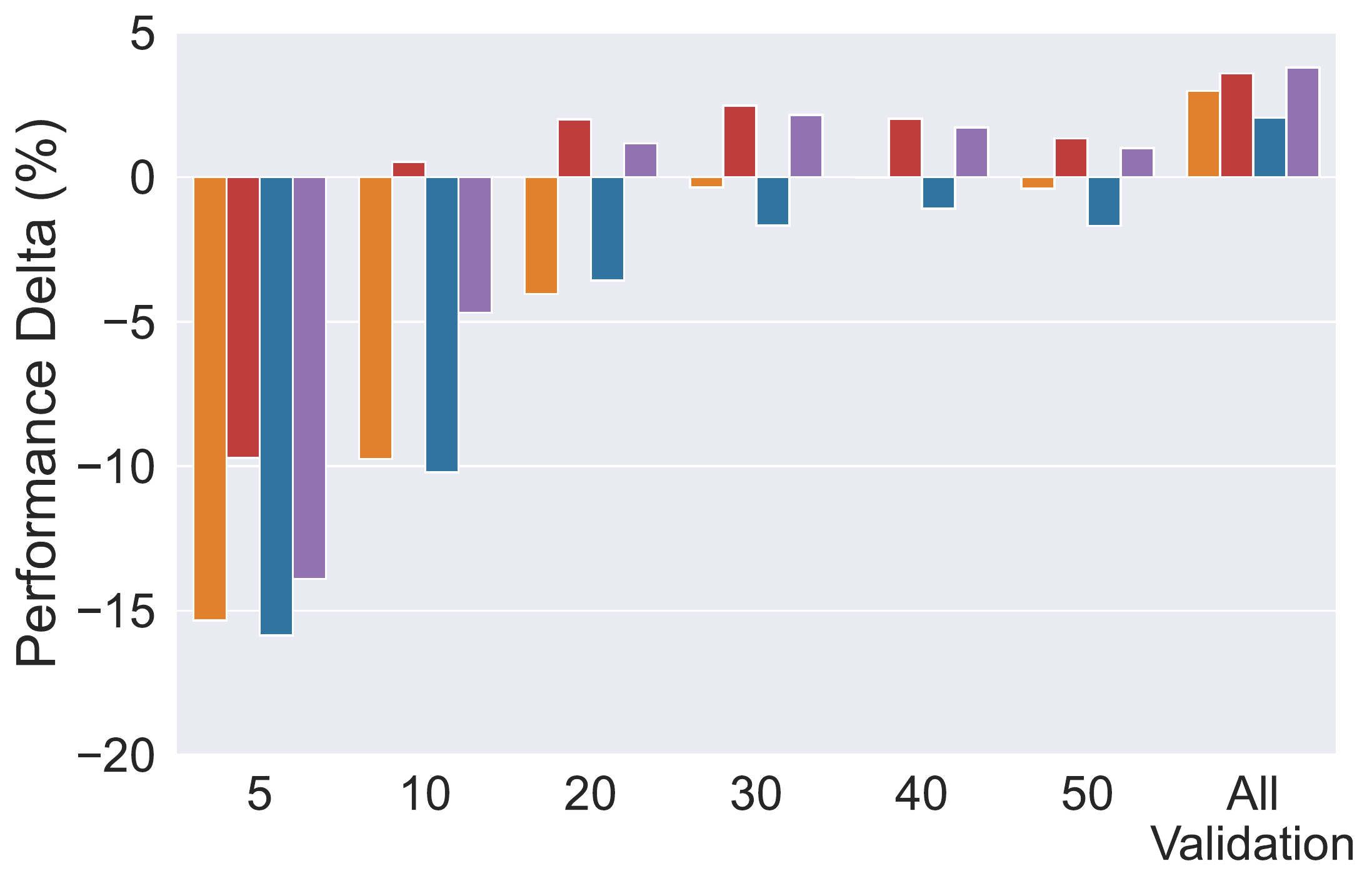}
         \caption{IMDb}

     \end{subfigure}\hfill
          \begin{subfigure}[b]{0.5\columnwidth}
         \centering
         \includegraphics[width=\columnwidth]{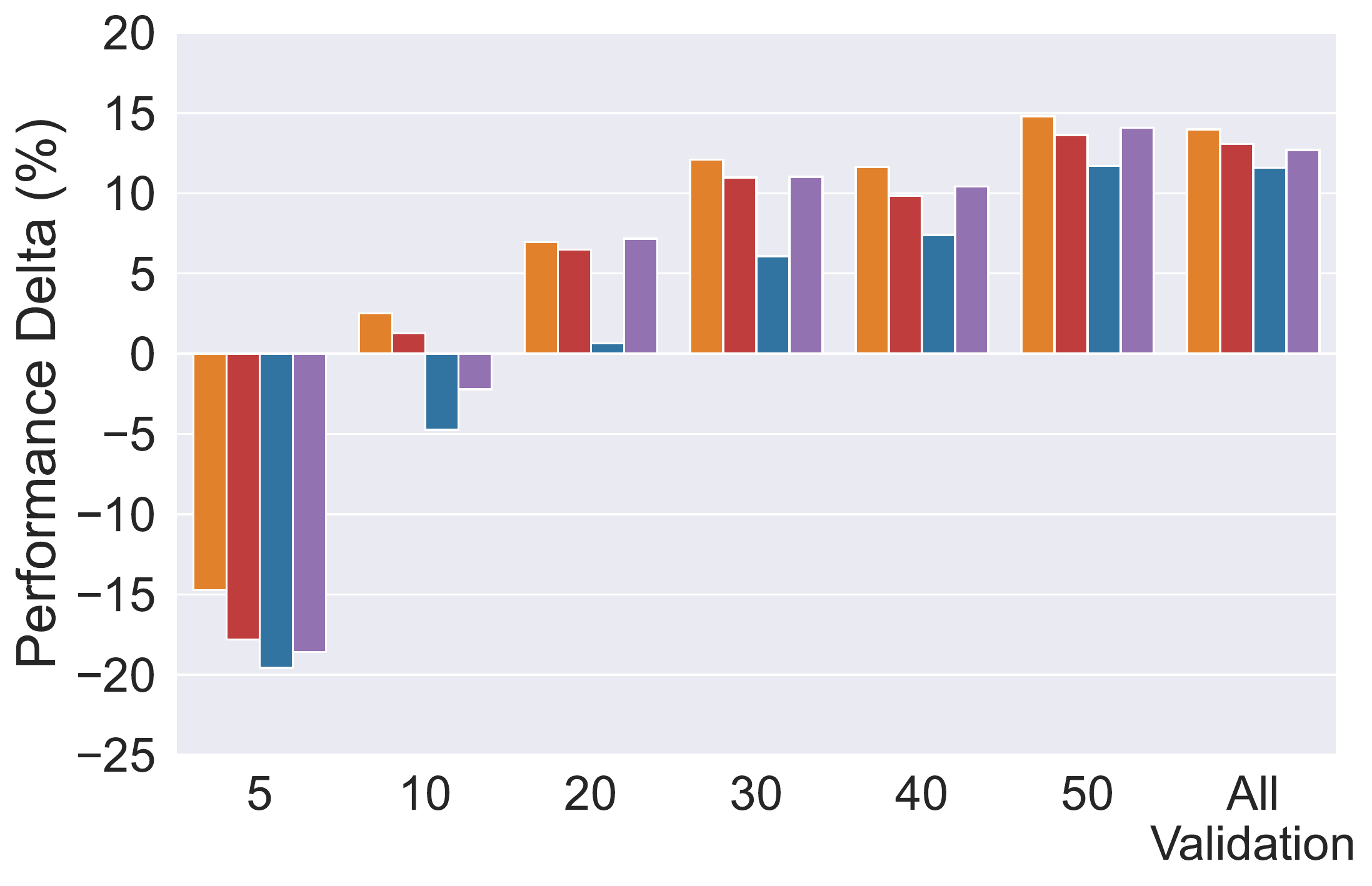}
         \caption{TREC}

     \end{subfigure}\hfill
     \begin{subfigure}[b]{0.5\columnwidth}
     \centering
     \includegraphics[width=\columnwidth]{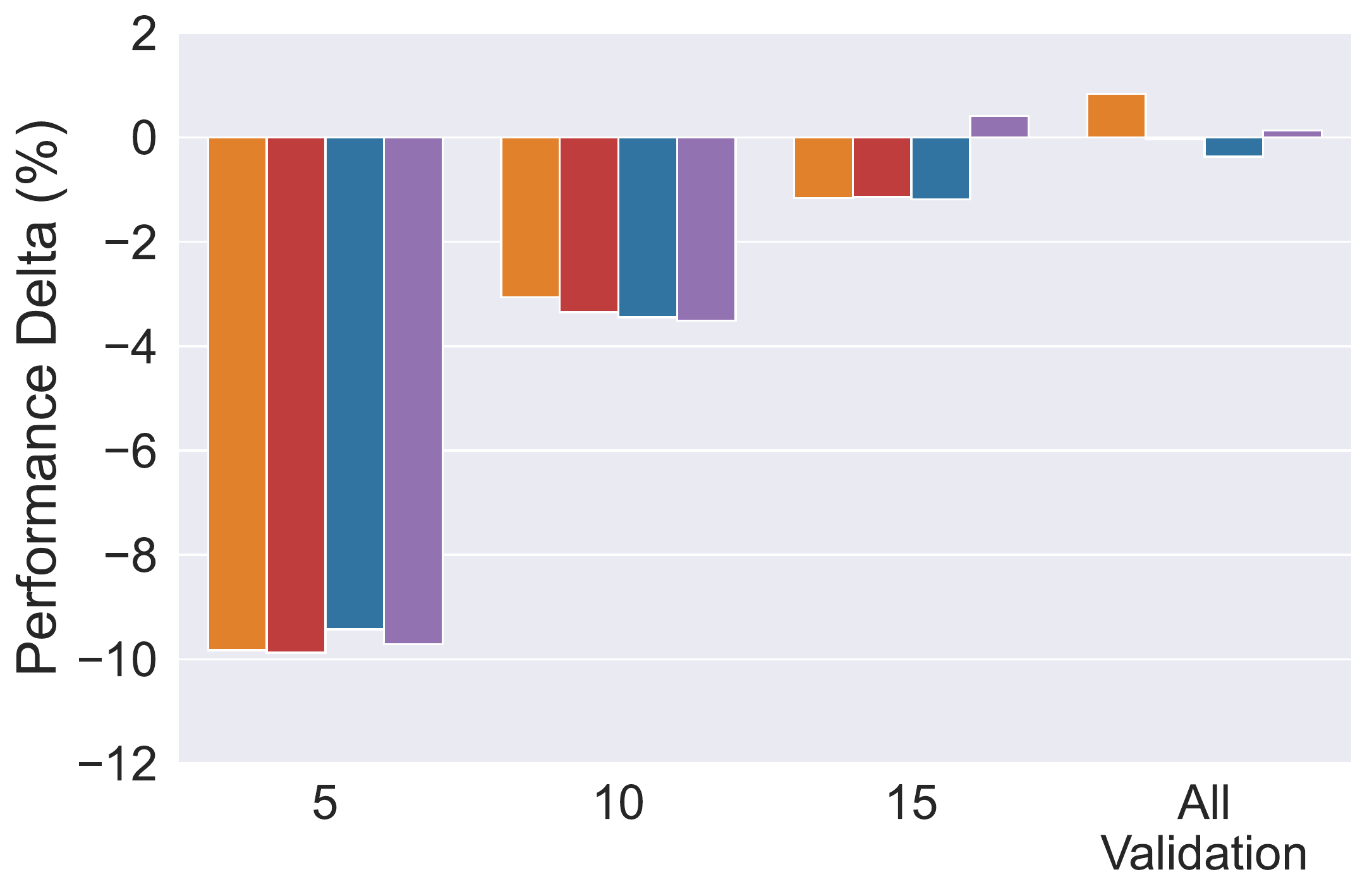}
     \caption{SemEval}

     \end{subfigure}\hfill
     \begin{subfigure}[b]{0.5\columnwidth}
     \centering
     \includegraphics[width=\columnwidth]{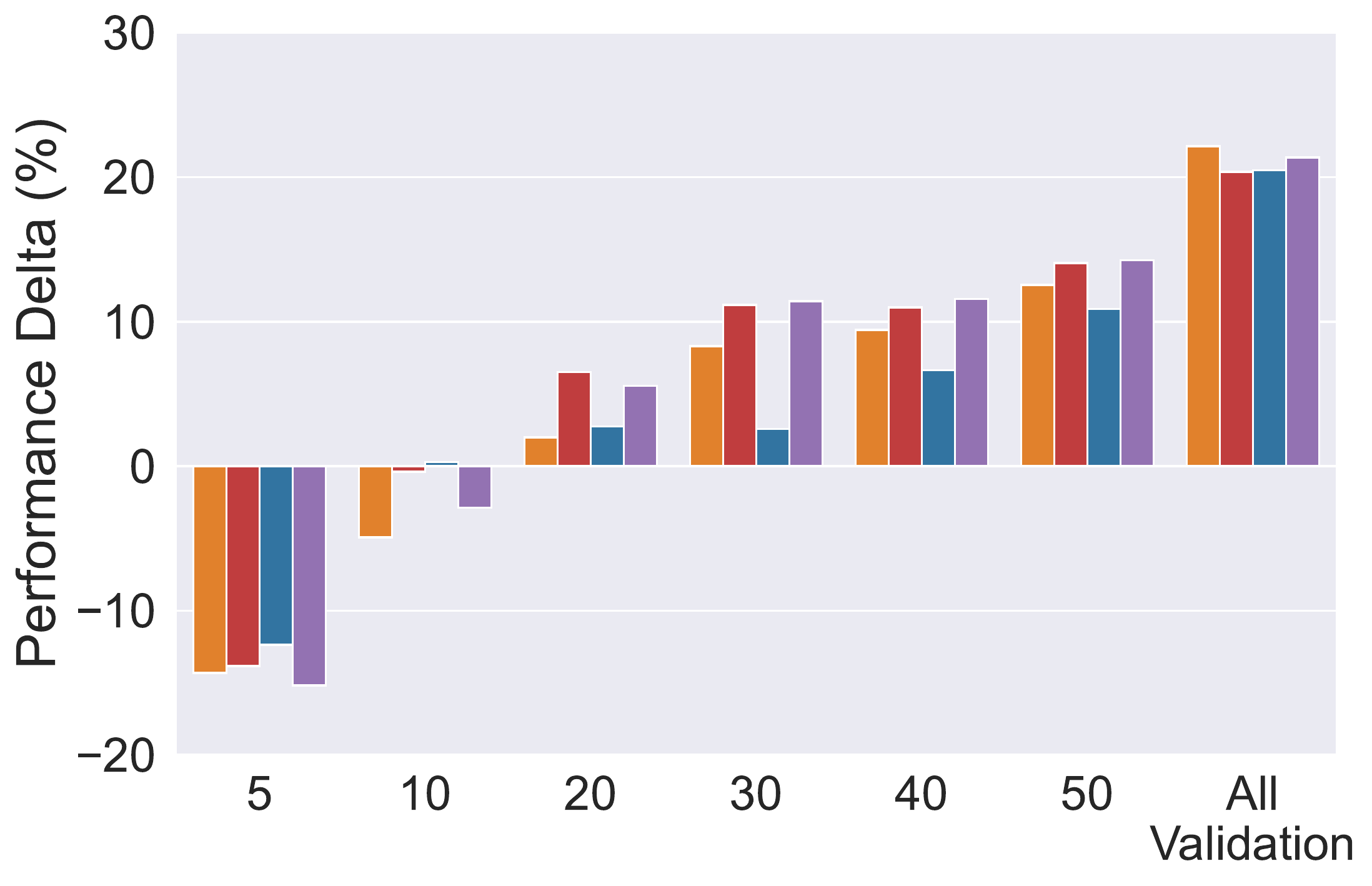}
     \caption{ChemProt}

     \end{subfigure}\hfill
     \begin{subfigure}[b]{0.5\columnwidth}
     \centering
     \includegraphics[width=\columnwidth]{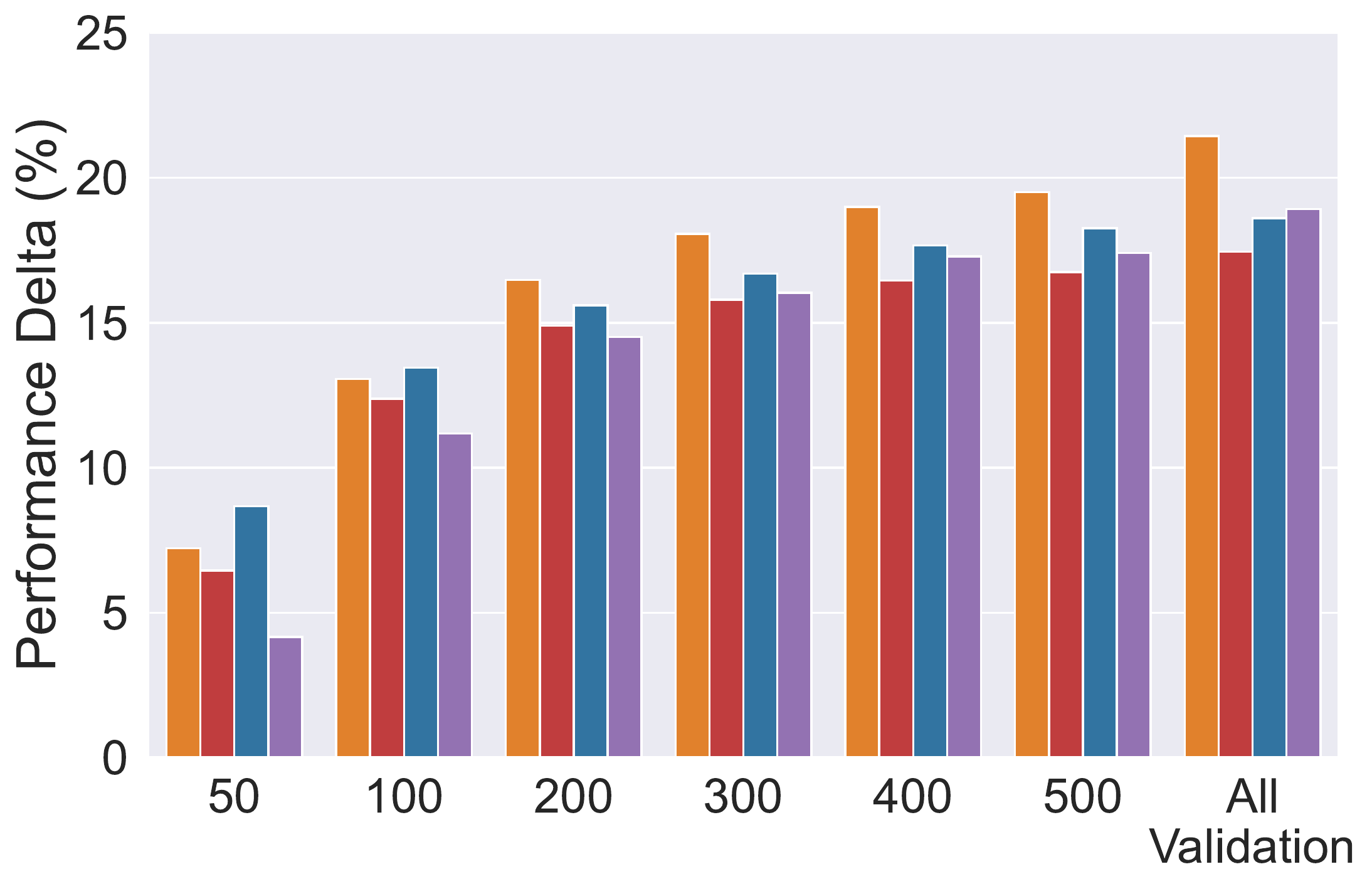}
     \caption{CoNLL-03}

     \end{subfigure}\hfill
      \begin{subfigure}[b]{0.5\columnwidth}
     \centering
     \includegraphics[width=\columnwidth]{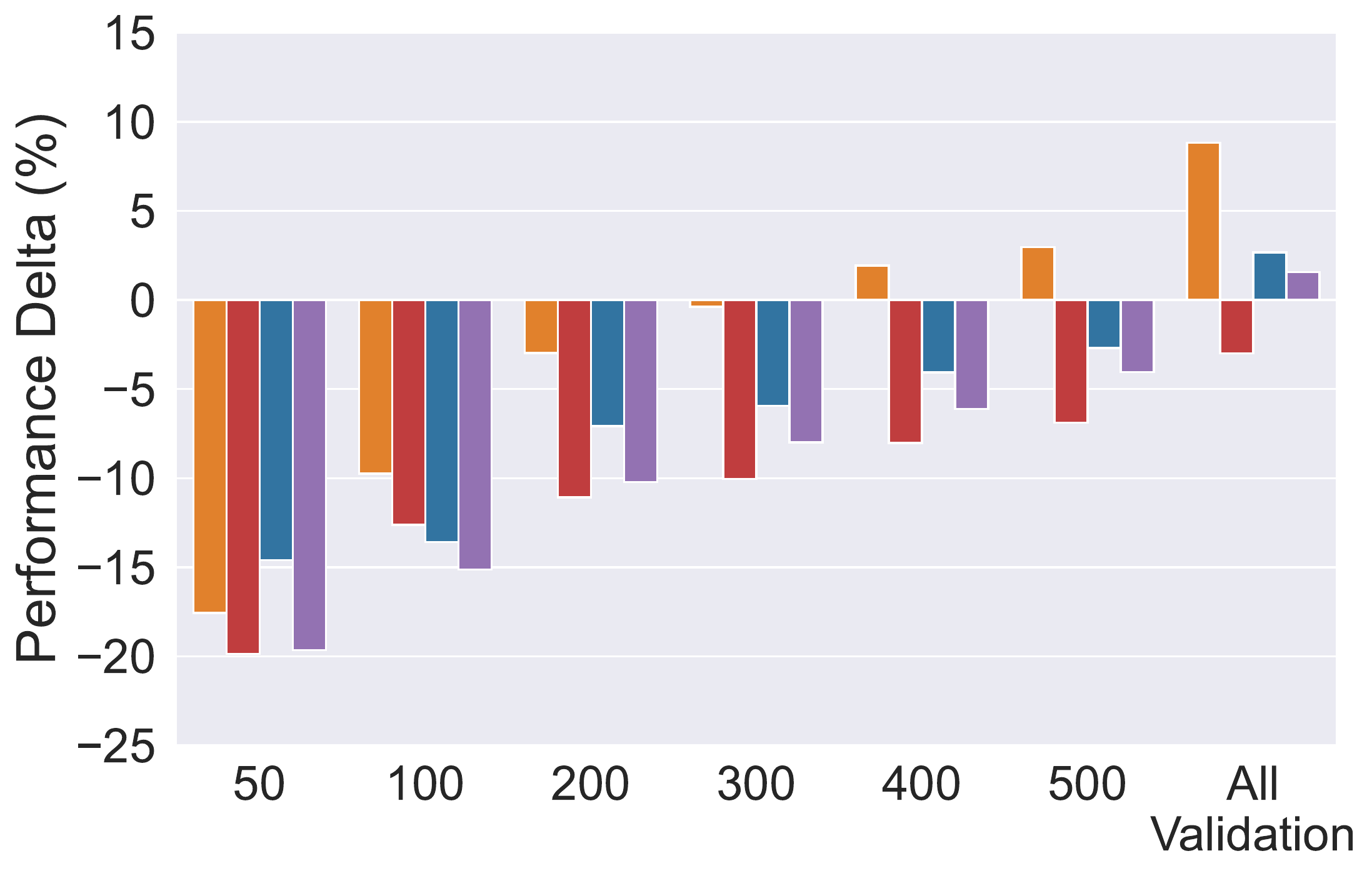}
     \caption{OntoNotes 5.0}

     \end{subfigure}\hfill
     
        \caption{\textbf{Using clean data for validation vs. training.}
        We show the average performance (Acc. and F1-score in \%) difference between (parameter-efficient) fine-tuning approaches and COSINE when varying amounts of clean samples. COSINE uses the clean samples for validation, whereas fine-tuning approaches directly train on them (indicated in the legend with the subscript `C'). For most sequence classification tasks, fine-tuning approaches work better once 10 clean samples are available for training. For NER, several hundreds of clean sentences may be required to attain better results via fine-tuning. Refer to Appendix \ref{sec:appendix:ft_on_clean_samples} for a comparison with other WSL approaches.}
        \label{fig:peft_vs_weak_cosine}
\end{figure*}

\paragraph{Results.} As shown in Figure~\ref{fig:wls_small_validation}, in most cases, a handful of validation samples already make WSL work better than the weak labels. We observe an increasing trend in performance with more validation samples, but typically this trend weakens with a moderate size of samples (\textasciitilde 30 samples per class or \textasciitilde 200 sentences) and adding more samples provides little benefit. There are a few exceptions. For example, on IMDb all methods except L2R consistently perform better with more validation data. On CoNLL-03, on the other hand, most methods seem to be less sensitive to the number of samples. Overall, the results suggest that \textbf{a small amount of clean validation samples may be sufficient for current WSL methods to achieve good performance.} Using thousands of validation samples, like in the established benchmarks \cite{zhang2021_wrench,Zheng2022_walnut}, is neither realistic nor necessary.

\section{Is WSL useful with less clean data?}
\label{sec:little_clean_data_for_peft}

The previous sections have shown that current WSL approaches (1) do not improve over direct fine-tuning on the existing validation splits (Figure~\ref{fig:train_on_full_clean_validation}) and (2) require only a small amount of validation samples to be effective (Figure~\ref{fig:wls_small_validation}). This section investigates whether the conclusion from Figure~\ref{fig:train_on_full_clean_validation} would change with less clean data, i.e., can WSL approaches outperform direct fine-tuning when less clean data is available?

\paragraph{Setup.} We follow the same procedure as in Section \ref{sec:how_much_clean_data_does_wsl_need} to subsample the \textit{cleanly annotated} validation sets and fine-tune models directly on the sampled data.
In addition to the standard fine-tuning approach \citep{devlin2019bert}, we also experiment with three parameter-efficient fine-tuning (PEFT) approaches as -- in the few-shot setting -- they have been shown to achieve comparable or even better performance than fine-tuning all parameters \cite{peters-etal-2019-tune, logan2021cutting, liu2022_few}. In particular, we include adapters \cite{houlsby2019parameter}, LoRA \cite{hu2021lora}, and BitFit \cite{zaken2022bitfit}.
 
We use one fixed set of hyperparameter configurations and train models for 6000 steps on each dataset.\footnote{The hyperparameters are randomly picked from the ranges mentioned in the original papers of corresponding methods and fixed across all experiments. \emph{We did not cherry-pick them based on the test performances}. In most cases the training loss converges within 300 steps. We intentionally extend training to show that we do not rely on extra data for early-stopping. We find that overfitting to the clean data does not hurt generalization. A similar observation is made in \citet{mosbach2021stability}. Detailed configurations are presented in Appendix \ref{sec:appendix:ft_on_clean_samples}.} We report performance at the last step and compare it with WSL approaches which use the same amount of clean data for validation.

\paragraph{Results.}
Figure \ref{fig:peft_vs_weak_cosine} shows the performance difference between the fine-tuning baselines and COSINE, one of the best-performing WSL approaches, when varying the number of clean samples. It can be seen that in extremely low-resource cases (less than 5 clean samples per class), COSINE outperforms fine-tuning. However, fine-tuning approaches quickly take over when more clean samples are available. LoRA performs better than COSINE on three out of four text classification tasks with just 10 samples per class. AGNews is the only exception, where COSINE outperforms LoRA by about 1\% when 20 samples per class are available, but adapters outperform COSINE in this case. Relation extraction has the same trend where 10--20 samples per class are often enough for fine-tuning approaches to catch up. For NER tasks, all fine-tuning approaches outperform COSINE with as few as 50 sentences on CoNLL-03. OntoNotes seems to be more challenging for fine-tuning and 400 sentences are required to overtake COSINE. Still, 400 sentences only account for $0.3\%$ of the weakly labeled samples used for training COSINE. This indicates that models can benefit much more from training on a small set of clean data rather than on vast amounts of weakly labeled data.
Note that the fine-tuning approaches we experiment with work out-of-the-box across NLP tasks. If one specific task is targeted, few-shot learning methods with manually designed prompts might perform even better.\footnote{For example, \citet{zhao2021calibrate} achieve an accuracy of 85.9\% on AGNews using just 4 labeled samples in total. For comparison, COSINE needs 20 labeled samples for validation to reach 84.21\%.} Hence, the performance shown here should be understood as a lower bound of what one can achieve by fine-tuning. Nevertheless, we can see that even considering the lower bound of fine-tuning-based methods, \textbf{the advantage of using WSL approaches vanishes when we have as few as 10 clean samples per class}. For many real-world applications, this annotation workload may be acceptable, limiting the applicability of WSL approaches.

\section{Can WSL benefit from fine-tuning?}
\label{sec:cft}

\begin{figure*}[ht!]
    \centering
     \begin{subfigure}[b]{1.99\columnwidth}
         \centering
         \includegraphics[width=\columnwidth]{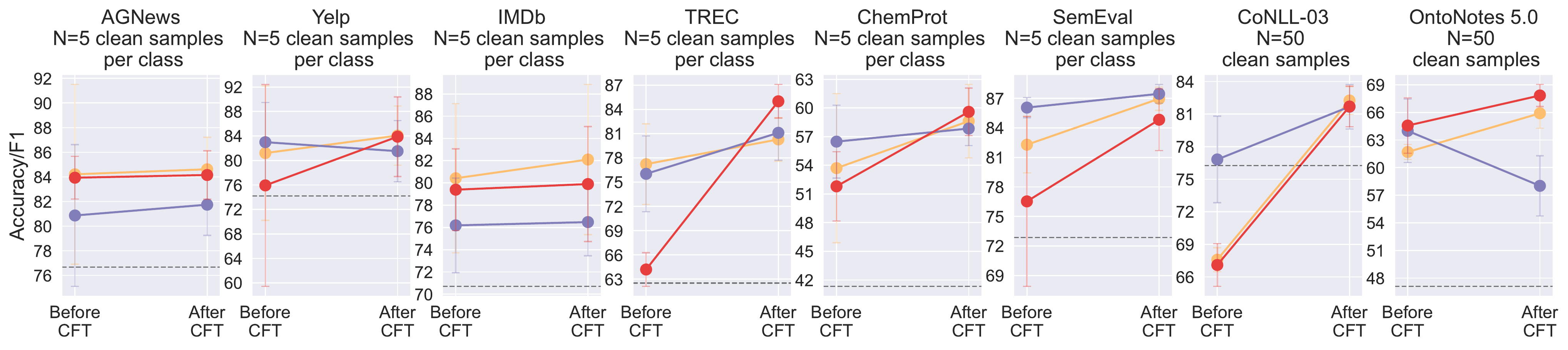}
         \caption{$N=5$ clean samples per class for classification tasks. $N=50$ clean samples for NER tasks.}
         \label{fig:wc_slope_plots_a}
     \end{subfigure}
     \vspace{0.5em}
          \begin{subfigure}[b]{1.99\columnwidth}
         \centering
         \includegraphics[width=\columnwidth]{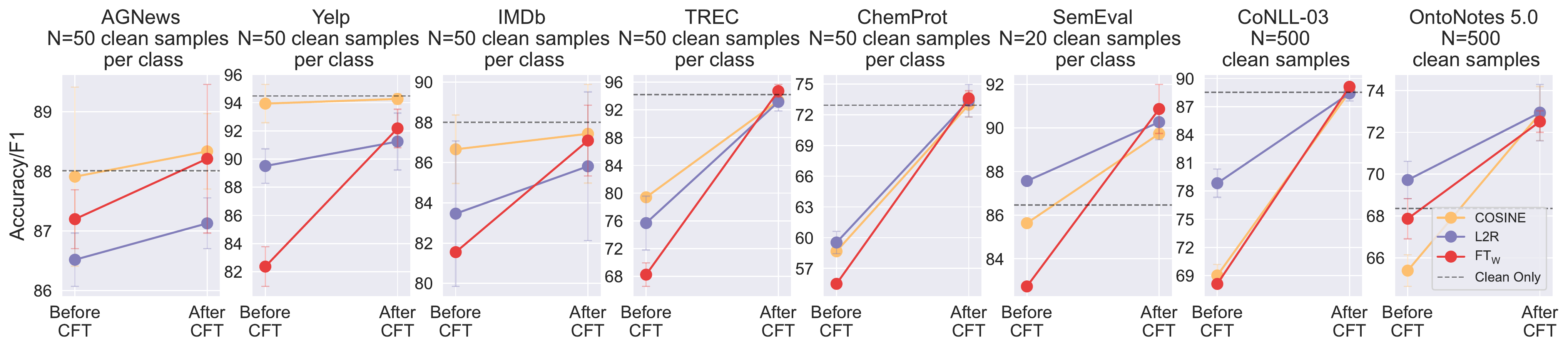}
         \caption{$N=50$ clean samples per class for classification tasks except for SemEval due to its limited validation size. $N=500$ clean samples for NER tasks}
         \label{fig:wc_slope_plots_b}

     \end{subfigure}\hfill
     
        \caption{\textbf{Performance before and after continuous fine-tuning (CFT) on the clean data}. The average performance and standard deviation over 5 runs are reported. Though CFT improves the performance of WSL approaches in general, the simplest baseline FT\textsubscript{W} gains the most from it. After applying CFT, FT\textsubscript{W} performs on par with or better than more sophisticated WSL approaches, suggesting these sophisticated approaches might have overestimated their actual value. Further plots are included in Appendix \ref{sec:appendix:cft}.}
        \label{fig:wc_slope_plots}
\end{figure*}

The WSL approaches have only used clean samples for validation so far, which is shown to be inefficient compared to training directly on them. We question whether enabling WSL methods to further fine-tune on these clean samples would improve their performance. In this section, we study a straightforward training approach that makes use of both clean and weak labels.\footnote{In Appendix \ref{sec:appendix:further_combination_baselines} we also explored other baselines that combine clean and weak data, but they perform considerably worse than the approach we consider in this section.}

\paragraph{Setup.} Given both the weakly labeled training data and a small amount of clean data, we consider a simple two-phase training baseline. In the first phase, we apply WSL approaches on the weakly labeled training set, using the clean data for validation. In the second phase, we take the model trained on the weakly labeled data as a starting point and continue to train it on the clean data. We call this approach continuous fine-tuning (\textbf{CFT}). In our experiment, we apply CFT to the two best-performing WSL approaches, COSINE and L2R, along with the most basic WSL baseline, FT\textsubscript{W}. We sample clean data in the same way as in Section \ref{sec:how_much_clean_data_does_wsl_need}. The training steps of the second phase are fixed at 6000. Each experiment is repeated 5 times with different seeds.

\paragraph{Results.}
Figure \ref{fig:wc_slope_plots} shows the model performance before and after applying CFT. It can be seen that CFT does indeed benefit WSL approaches in most cases even with very little clean data (Figure \ref{fig:wc_slope_plots_a}). For L2R, however, the improvement is less obvious, and there is even a decrease on Yelp and OntoNotes. This could be because L2R uses the validation loss to reweight training samples, meaning that the value of the validation samples beyond that may only be minimal. When more clean samples are provided, CFT exhibits a greater performance gain (Figure \ref{fig:wc_slope_plots_b}).
It is also noticeable that CFT reduces the performance gap among all three WSL methods substantially. Even the simplest approach, FT\textsubscript{W}, is comparable to or beats L2R and COSINE in all tasks after applying CFT. Considering that COSINE and L2R consume far more computing resources, our findings suggest that 
 \textbf{the net benefit of using sophisticated WSL approaches may be significantly overestimated and impractical for real-world use cases.}

Finally, we find the advantage of performing WSL diminishes with the increase of clean samples even after considering the boost from CFT. When 50 clean samples per class (500 sentences for NER) are available, applying WSL+CFT only results in a performance boost of less than 1\% on 6 out of 8 datasets, compared with the baseline which only fine-tunes on clean samples. Note that weak labels are no free lunch. Managing weak annotation resources necessitates experts who not only have linguistic expertise for annotation but also the ability to transform that knowledge into programs to automate annotations. This additional requirement naturally reduces the pool of eligible candidates and raises the cost. In this situation, annotating a certain amount of clean samples may be significantly faster and cheaper. Thus, we believe WSL has a long way to go before being truly helpful in realistic low-resource scenarios.

\begin{figure}[t!]
        \centering
        
     \begin{subfigure}[b]{1\columnwidth}
         \centering
         \includegraphics[width=\columnwidth]{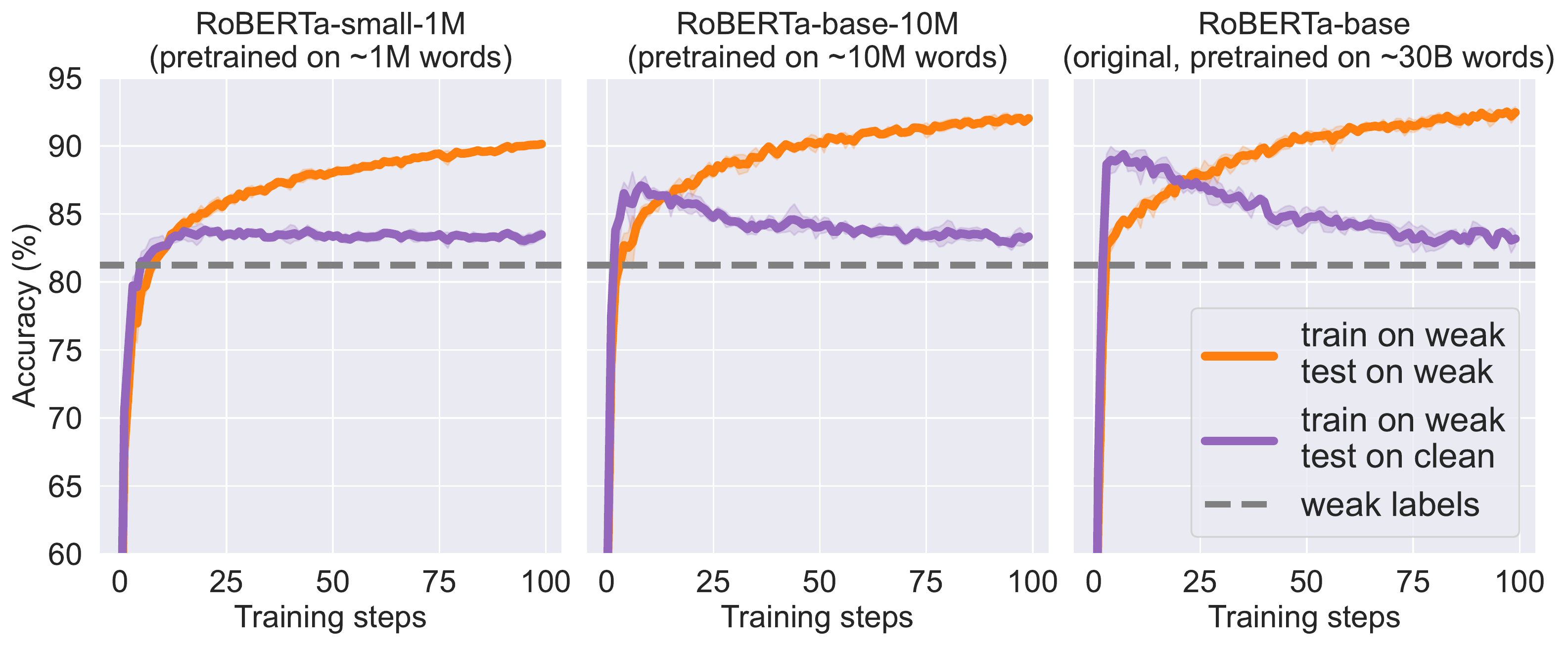}
         \caption{AGNews}
         \label{fig:plm_dff_a}
     \end{subfigure}\hfill
          \begin{subfigure}[b]{1\columnwidth}
         \centering
         \includegraphics[width=\columnwidth]{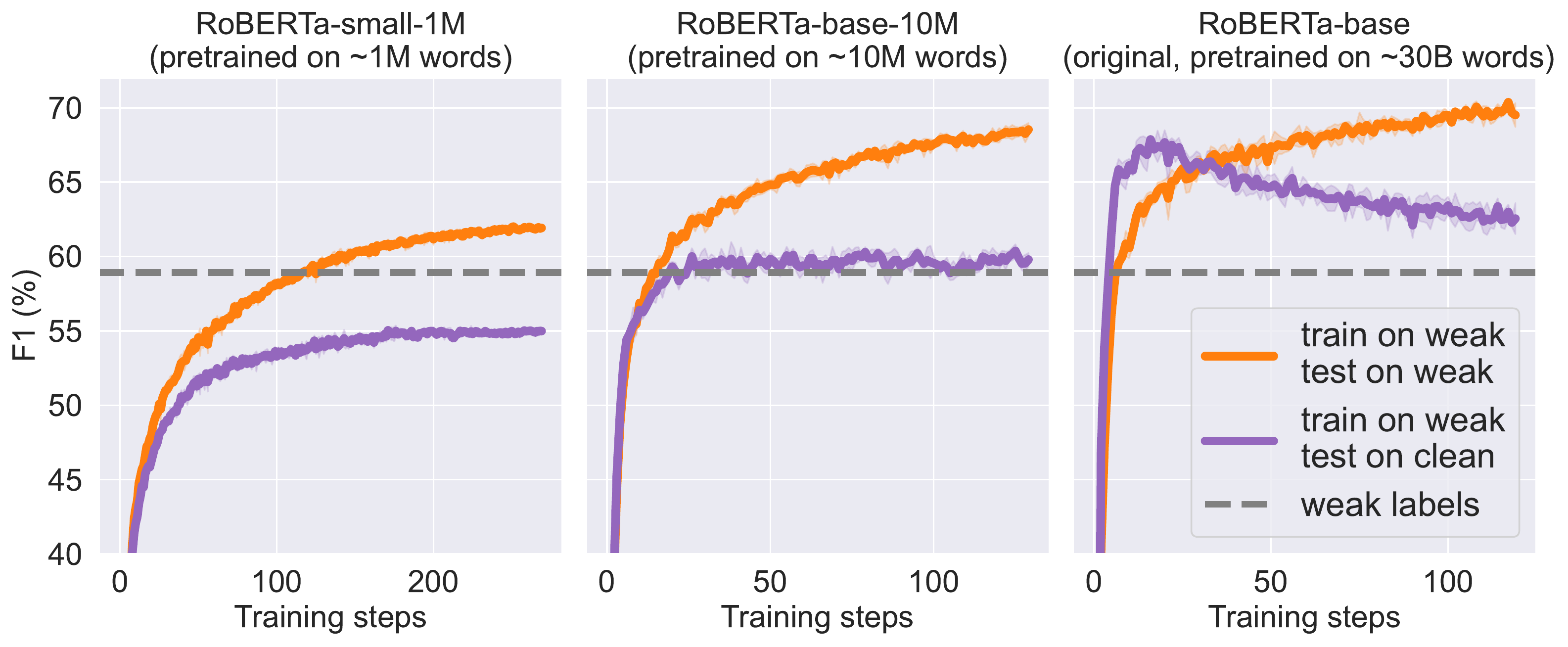}
         \caption{OntoNotes 5.0}
         \label{fig:plm_dff_b}

     \end{subfigure}\hfill
        \caption{\textbf{Performance curves of different PLMs during training}. PLMs are trained on weak labels and evaluated on both clean and weakly labeled test sets. Pre-training on larger corpora improves performance on the clean distribution. Further plots are in Appendix \ref{sec:appendix:cft_abalation}.}
        \label{fig:plm_dff}
\end{figure}

\begin{figure}[ht!]
        \centering
        
     \begin{subfigure}[b]{0.5\columnwidth}
         \centering
         \includegraphics[width=\columnwidth]{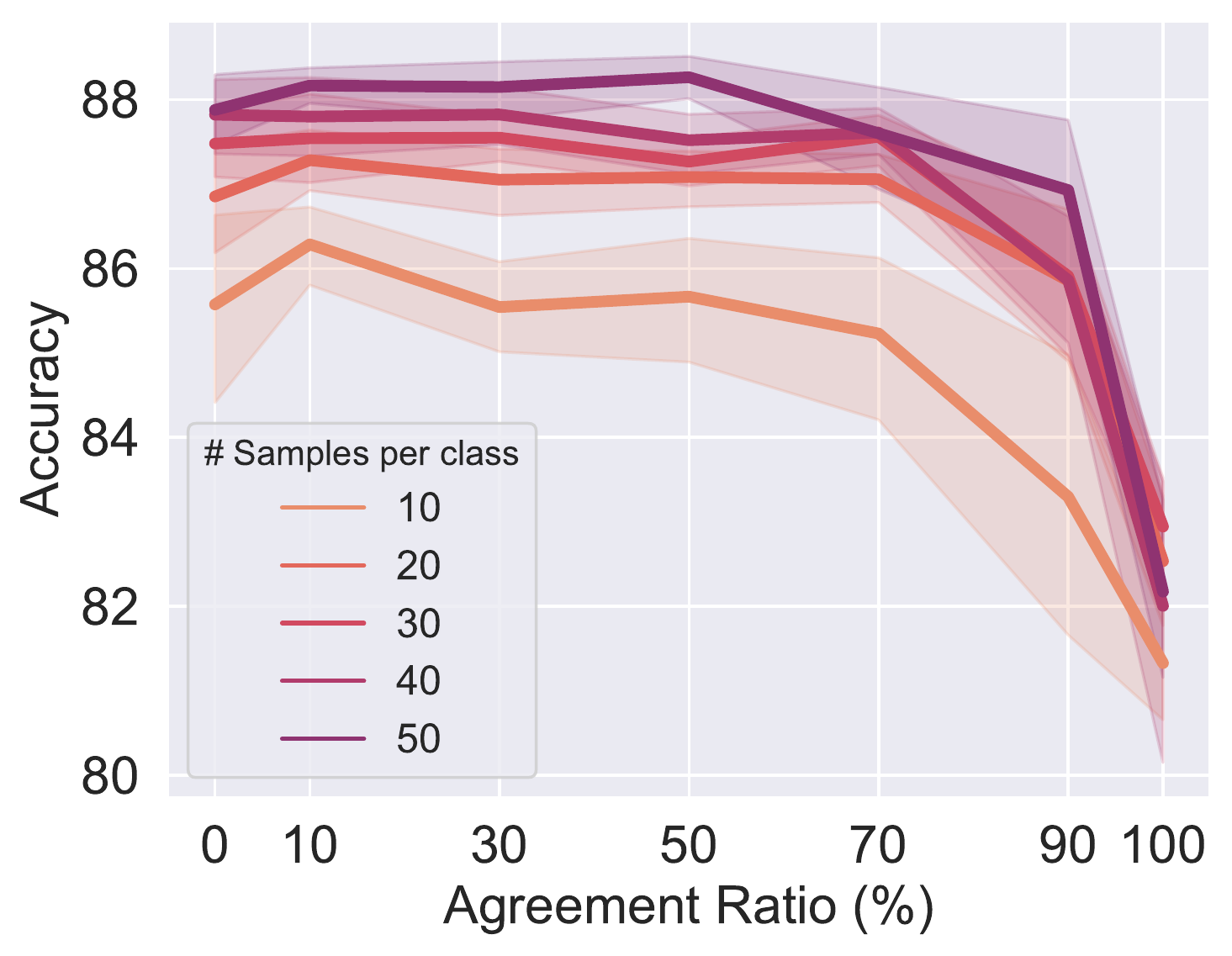}
         \caption{AGNews}
    \vspace{0.3cm}
     \end{subfigure}\hfill
          \begin{subfigure}[b]{0.5\columnwidth}
         \centering
         \includegraphics[width=\columnwidth]{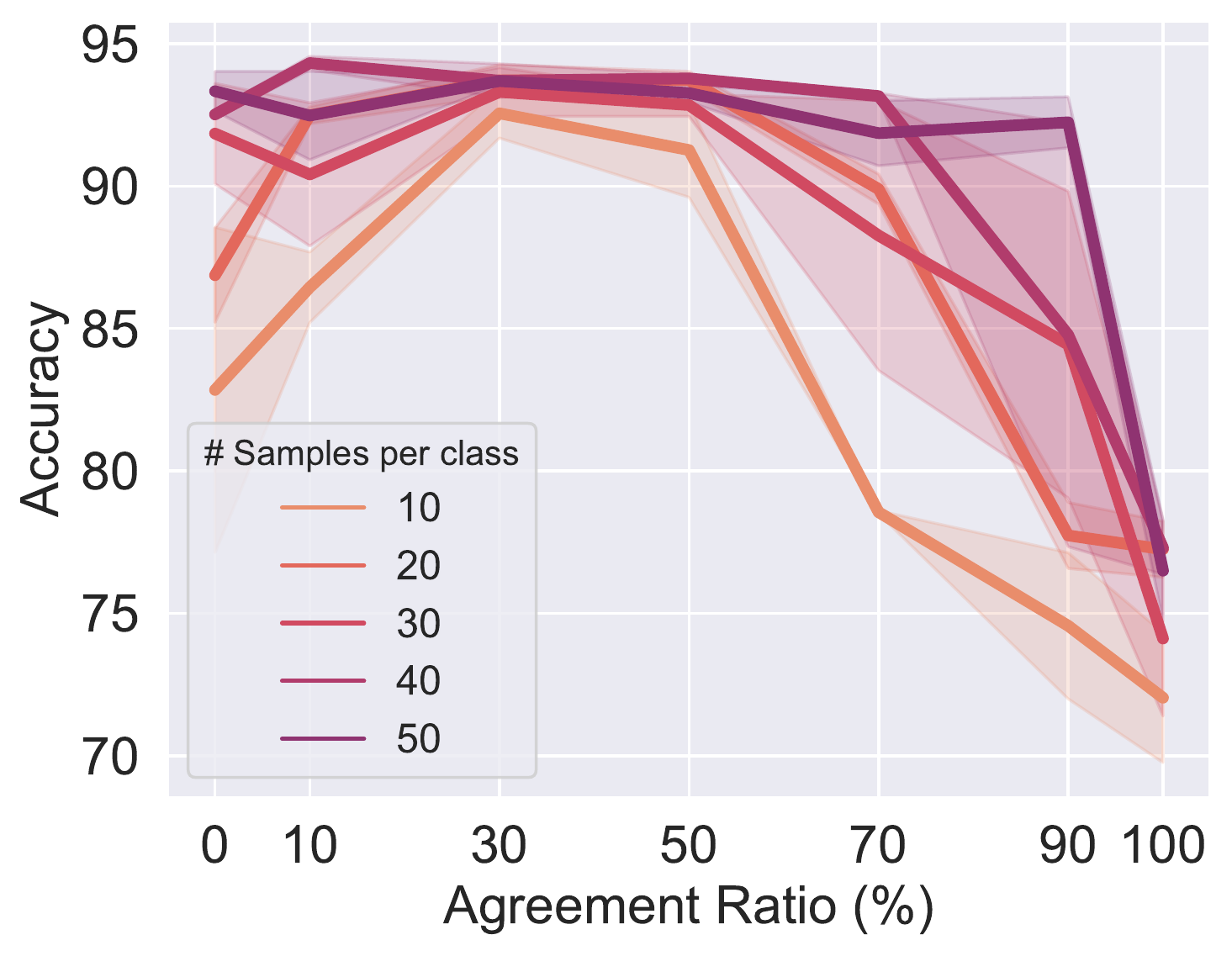}
         \caption{Yelp}
    \vspace{0.3cm}
     \end{subfigure}\hfill
          \begin{subfigure}[b]{0.5\columnwidth}
         \centering
         \includegraphics[width=\columnwidth]{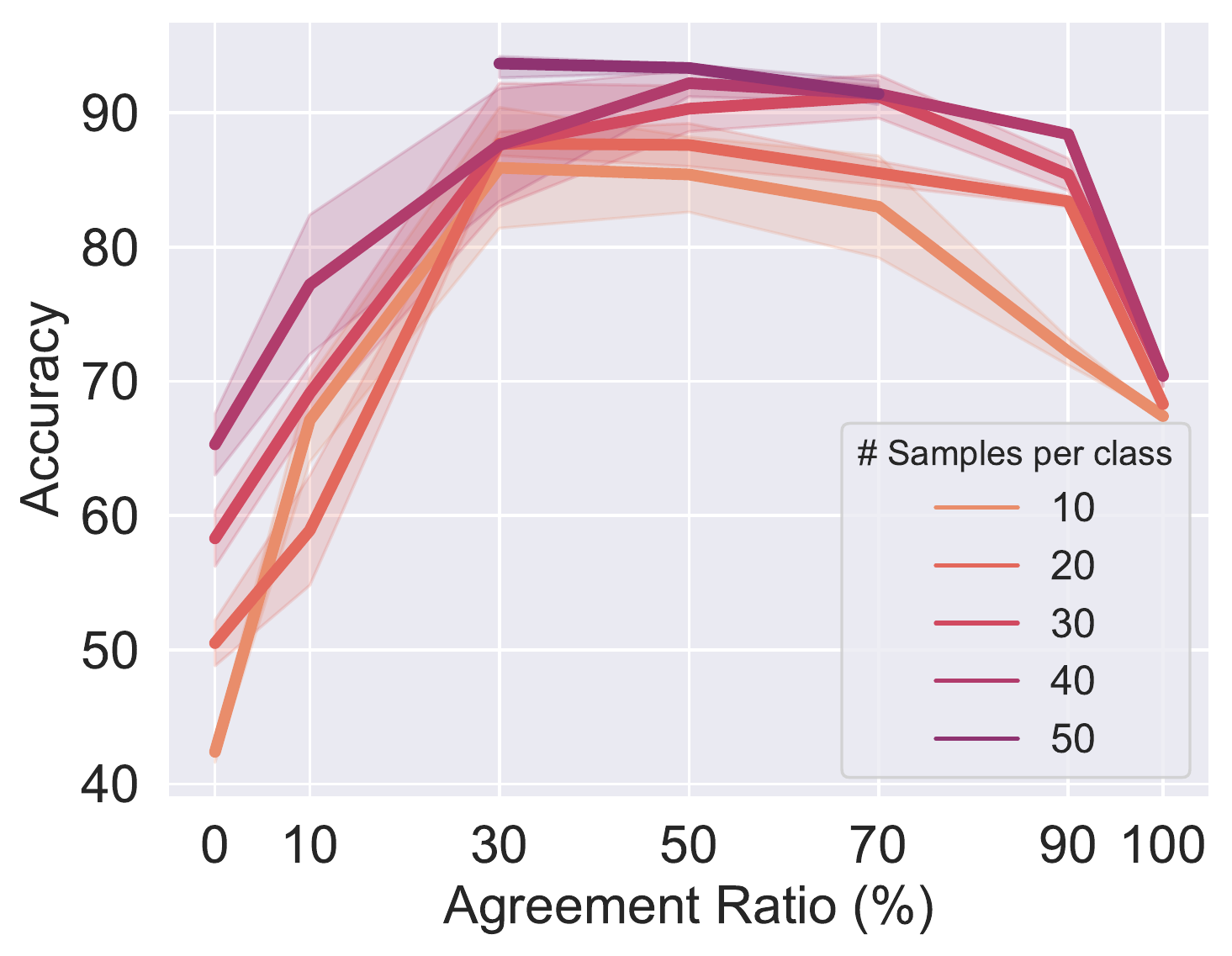}
         \caption{TREC$^*$}

     \end{subfigure}\hfill
          \begin{subfigure}[b]{0.5\columnwidth}
         \centering
         \includegraphics[width=\columnwidth]{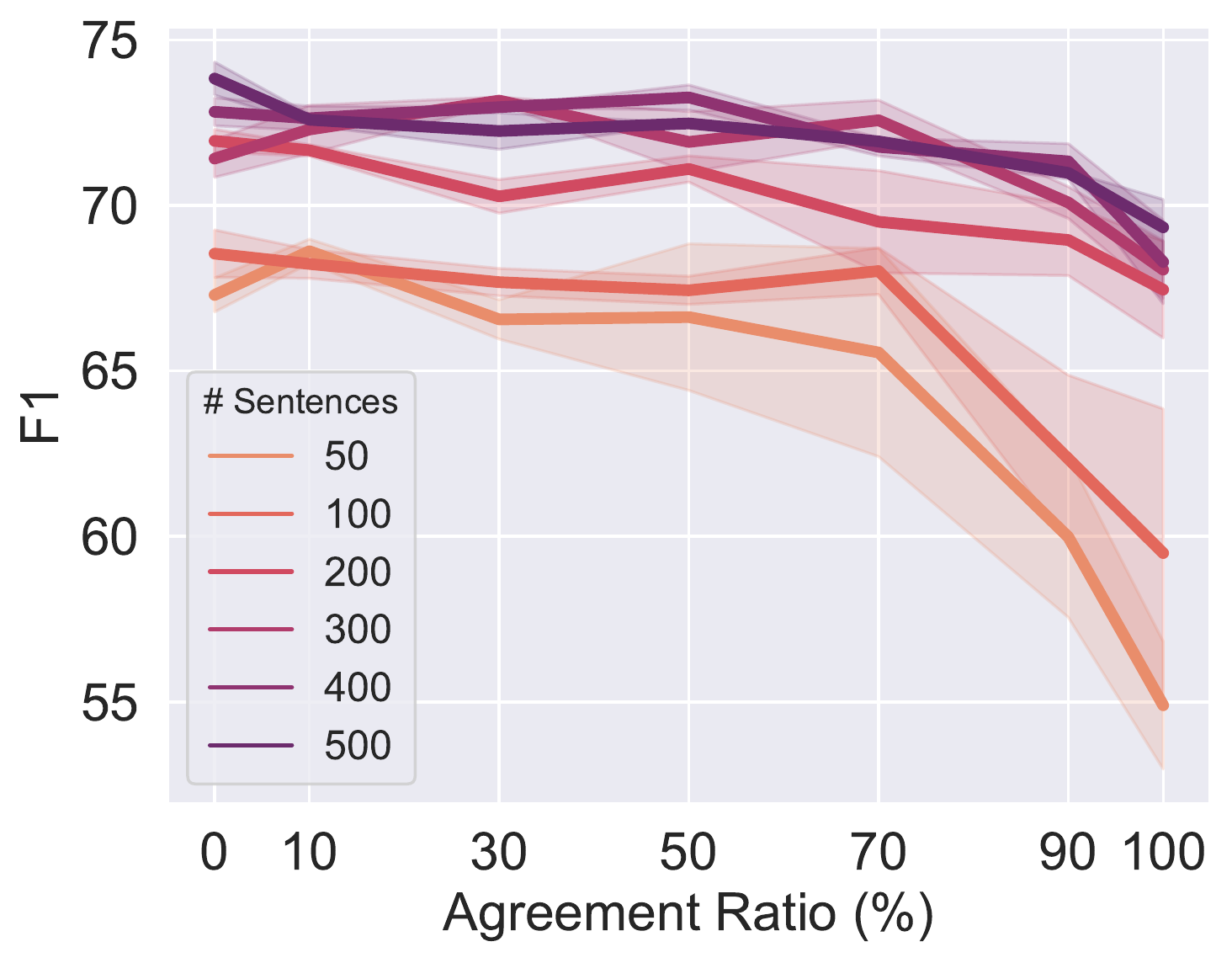}
         \caption{OntoNotes 5.0}

     \end{subfigure}

        \caption{\textbf{Model performance varying the number of clean samples $N$ and agreement ratio $\alpha$}. Large $\alpha$ generally causes a substantial drop in performance. $^*$: Certain combinations of $\alpha$ and $N$ are not feasible because the validation set lacks samples with clean and weak labels that coincide or differ. Further plots are in Appendix \ref{sec:appendix:cft_abalation}.}
        \label{fig:cft_ablation}
\end{figure}

\section{What makes FT\textsubscript{W}+CFT effective?}
\label{sec:cft_ablation}
As seen in the previous section, combining FT\textsubscript{W} with CFT yields a strong baseline that more sophisticated WSL approaches can hardly surpass. This section examines factors that contribute to the effectiveness of this method. Specifically, we aim to answer two questions: (1) ``How does FT\textsubscript{W} resist biases despite being trained only on weak labels?” and (2) ``How does CFT further reduce bias introduced by weak labels?”.
\paragraph{Setup.} To answer question (1), we modify the backbone PLM to see if its encoded knowledge plays an important role. In addition to RoBERTa-base, we explore two other PLMs that are pre-trained on less data: RoBERTa-small-1M and RoBERTa-base-10M, which are pre-trained on 1M and 10M words, respectively.\footnote{The original RoBERTa-base model is pre-trained on 100B words. The two less pre-trained models are obtained from \cite{warstadt2020learning}. RoBERTa-base-10M retains the same architecture as RoBERTa-base, while RoBERTa-small-1M contains fewer parameters.} We report model performance on both clean labels and weak labels to see which labels the model tends to fit. To answer question (2), we vary the agreement ratio in the clean samples to see how these clean labels help combat biases from weak labels. The agreement ratio is defined as the percentage of samples whose clean labels match the corresponding weak labels. Intuitively, if the clean label for a specific training example matches its weak label, then this example may not contribute additional information to help combat bias. A higher agreement ratio should therefore indicate fewer informative samples.

\paragraph{Results.} Figure \ref{fig:plm_dff} shows the performances for different PLMs. Pre-training on more data clearly helps to overcome biases from weak labels. When the pre-training corpus is small, the model tends to fit the noisy weak labels more quickly than the clean labels and struggles to outperform weak labels throughout the entire training process (\Cref{fig:plm_dff_a,fig:plm_dff_b}, left). With a large pre-training corpus, however, the model can make better predictions on clean labels than weak labels in the early stages of training, even when it is only trained on weak labels (\Cref{fig:plm_dff_a,fig:plm_dff_b}, right). If we apply proper early-stopping before the model is eventually dragged toward weak labels, we can attain a model that generalizes significantly better than the weak labels. This indicates that \emph{pre-training provides the model with an inductive bias to seek more general linguistic correlations instead of superficial correlations from the weak labels}, which aligns with previous findings in \citet{warstadt2020learning}. This turns out to be the key to why simple FT\textsubscript{W} works here. Figure~\ref{fig:cft_ablation} shows how the agreement ratio $\alpha$ in clean samples affects the performance. Performance declines substantially for $\alpha > 70\%$, showing that it is necessary to have contradictory samples in order to reap the full advantage of CFT. This is reasonable, given that having examples with clean labels that coincide with their weak labels may reinforce the unintended bias learned from the weakly labeled training set. The optimal agreement ratio lies around 50\%. However, having $\alpha =0$ also yields decent performance for most datasets except TREC, suggesting contradictory samples play a more important role here and at least a minimum set of contradictory samples are required for CFT to be beneficial.

\section{Conclusions and recommendations}
Our extensive experiments provide strong evidence that recent WSL approaches heavily overestimate their performance and practicality. We demonstrated that they hinge on clean samples for model selection to reach the claimed performance, yet models that are simply trained on these clean samples are already better. When both clean and weak labels are available, a simple baseline (FT\textsubscript{W}+CFT) performs on par with or better than more sophisticated methods while requiring much less computation and effort for model selection.

Inspired by prior work \cite{oliver2018realistic, perez2021true}, our recommendations for future WSL approaches are the following: 

\begin{itemize}
  \item Report the model selection criteria for proposed methods and, especially, how much they rely on the presence of clean data.
  \item Report how many cleanly annotated samples are required for a few-shot learning approach to reach the performance of a proposed WSL approach. If thousands of weakly annotated samples are comparable to a handful of clean samples -- as we have seen in Section \ref{sec:little_clean_data_for_peft} -- then WSL may not be the best choice for the given low-resource setting.
  \item If a proposed WSL method requires extra clean data, such as for validation, then the simple FT\textsubscript{W}+CFT baseline should be included in evaluation to claim the real benefits gained by applying the method.
\end{itemize}

We hope our findings and recommendations will spur more robust future work in WSL such that new methods are truly beneficial in realistic low-resource scenarios.

\section*{Limitations}
We facilitate fair comparisons and realistic evaluations of recent WSL approaches. However, our study is not exhaustive and has the following limitations. 

First, it may be possible to perform model selection by utilizing prior knowledge about the dataset. For example, if the noise ratio (the proportion of incorrect labels in the training set) is known in advance, it can be used to determine (a subset of) hyperparameters \citep{han2018co, li2020dividemix}. In this case, certain WSL approaches may still work without access to extra clean data.

Second, in this paper we concentrate on tasks in English where strong PLMs are available. As we have shown in Section \ref{sec:little_clean_data_for_peft}, training them on a small amount of data is sufficient for generalization. For low-resource languages where no PLMs are available, training may not be that effective, and WSL methods may achieve higher performance.

Third, we experiment with datasets from the established WRENCH benchmark, where the weak labels are frequently assigned by simple rules like as regular expressions (see Appendix \ref{sec:appendix:rules} for examples). However, in a broader context, weak supervision can have different forms. For example, \citet{smith2022language} generates weak labels through large language models. \citet{zhou2022hyperlink} use hyper-link information as weak labels for passage retrieval. We have not extended our research to more diverse types of weak labels.

Despite the above limitations, however, we identify the pitfalls in the existing evaluation of current WSL methods and demonstrate simple yet strong baselines through comprehensive experiments on a wide range of tasks.

\section*{Acknowledgments}
We thank Vagrant Gautam for thoughtful suggestions and insightful discussions. We would also like to thank our anonymous reviewers for their constructive feedback.

This work has been partially funded by the Deutsche Forschungsgemeinschaft (DFG, German Research Foundation) – Project-ID 232722074 – SFB 1102 and the EU Horizon 2020 projects ROXANNE under grant number 833635.

\bibliography{tws_ref}

\begin{thebibliography}{58}
\expandafter\ifx\csname natexlab\endcsname\relax\def\natexlab#1{#1}\fi

\bibitem[{Alex et~al.(2021)Alex, Lifland, Tunstall, Thakur, Maham, Riedel,
  Hine, Ashurst, Sedille, Carlier, Noetel, and
  Stuhlm{\"{u}}ller}]{alex2021raft}
Neel Alex, Eli Lifland, Lewis Tunstall, Abhishek Thakur, Pegah Maham, C.~Jess
  Riedel, Emmie Hine, Carolyn Ashurst, Paul Sedille, Alexis Carlier, Michael
  Noetel, and Andreas Stuhlm{\"{u}}ller. 2021.
\newblock \href
  {https://datasets-benchmarks-proceedings.neurips.cc/paper/2021/hash/ca46c1b9512a7a8315fa3c5a946e8265-Abstract-round2.html}
  {{RAFT:} {A} real-world few-shot text classification benchmark}.
\newblock In \emph{Proceedings of the Neural Information Processing Systems
  Track on Datasets and Benchmarks 1, NeurIPS Datasets and Benchmarks 2021,
  December 2021, virtual}.

\bibitem[{Arpit et~al.(2017)Arpit, Jastrzebski, Ballas, Krueger, Bengio,
  Kanwal, Maharaj, Fischer, Courville, Bengio, and
  Lacoste{-}Julien}]{arpit2017closer}
Devansh Arpit, Stanislaw Jastrzebski, Nicolas Ballas, David Krueger, Emmanuel
  Bengio, Maxinder~S. Kanwal, Tegan Maharaj, Asja Fischer, Aaron~C. Courville,
  Yoshua Bengio, and Simon Lacoste{-}Julien. 2017.
\newblock \href {http://proceedings.mlr.press/v70/arpit17a.html} {A closer look
  at memorization in deep networks}.
\newblock In \emph{Proceedings of the 34th International Conference on Machine
  Learning, {ICML} 2017, Sydney, NSW, Australia, 6-11 August 2017}, volume~70
  of \emph{Proceedings of Machine Learning Research}, pages 233--242. {PMLR}.

\bibitem[{Bragg et~al.(2021)Bragg, Cohan, Lo, and Beltagy}]{bragg2021flex}
Jonathan Bragg, Arman Cohan, Kyle Lo, and Iz~Beltagy. 2021.
\newblock \href
  {https://proceedings.neurips.cc/paper/2021/hash/8493eeaccb772c0878f99d60a0bd2bb3-Abstract.html}
  {{FLEX:} unifying evaluation for few-shot {NLP}}.
\newblock In \emph{Advances in Neural Information Processing Systems 34: Annual
  Conference on Neural Information Processing Systems 2021, NeurIPS 2021,
  December 6-14, 2021, virtual}, pages 15787--15800.

\bibitem[{Devlin et~al.(2019)Devlin, Chang, Lee, and
  Toutanova}]{devlin2019bert}
Jacob Devlin, Ming{-}Wei Chang, Kenton Lee, and Kristina Toutanova. 2019.
\newblock \href {https://doi.org/10.18653/v1/n19-1423} {{BERT:} pre-training of
  deep bidirectional transformers for language understanding}.
\newblock In \emph{Proceedings of the 2019 Conference of the North American
  Chapter of the Association for Computational Linguistics: Human Language
  Technologies, {NAACL-HLT} 2019, Minneapolis, MN, USA, June 2-7, 2019, Volume
  1 (Long and Short Papers)}, pages 4171--4186. Association for Computational
  Linguistics.

\bibitem[{Gao et~al.(2021)Gao, Fisch, and Chen}]{gao2021making}
Tianyu Gao, Adam Fisch, and Danqi Chen. 2021.
\newblock \href {https://doi.org/10.18653/v1/2021.acl-long.295} {Making
  pre-trained language models better few-shot learners}.
\newblock In \emph{Proceedings of the 59th Annual Meeting of the Association
  for Computational Linguistics and the 11th International Joint Conference on
  Natural Language Processing, {ACL/IJCNLP} 2021, (Volume 1: Long Papers),
  Virtual Event, August 1-6, 2021}, pages 3816--3830. Association for
  Computational Linguistics.

\bibitem[{Han et~al.(2018)Han, Yao, Yu, Niu, Xu, Hu, Tsang, and
  Sugiyama}]{han2018co}
Bo~Han, Quanming Yao, Xingrui Yu, Gang Niu, Miao Xu, Weihua Hu, Ivor~W. Tsang,
  and Masashi Sugiyama. 2018.
\newblock \href
  {https://proceedings.neurips.cc/paper/2018/hash/a19744e268754fb0148b017647355b7b-Abstract.html}
  {Co-teaching: Robust training of deep neural networks with extremely noisy
  labels}.
\newblock In \emph{Advances in Neural Information Processing Systems 31: Annual
  Conference on Neural Information Processing Systems 2018, NeurIPS 2018,
  December 3-8, 2018, Montr{\'{e}}al, Canada}, pages 8536--8546.

\bibitem[{Hendrickx et~al.(2010)Hendrickx, Kim, Kozareva, Nakov,
  {\'O}~S{\'e}aghdha, Pad{\'o}, Pennacchiotti, Romano, and
  Szpakowicz}]{hendrickx2010_semeval}
Iris Hendrickx, Su~Nam Kim, Zornitsa Kozareva, Preslav Nakov, Diarmuid
  {\'O}~S{\'e}aghdha, Sebastian Pad{\'o}, Marco Pennacchiotti, Lorenza Romano,
  and Stan Szpakowicz. 2010.
\newblock \href {https://aclanthology.org/S10-1006} {{S}em{E}val-2010 task 8:
  Multi-way classification of semantic relations between pairs of nominals}.
\newblock In \emph{Proceedings of the 5th International Workshop on Semantic
  Evaluation}, pages 33--38, Uppsala, Sweden. Association for Computational
  Linguistics.

\bibitem[{Houlsby et~al.(2019)Houlsby, Giurgiu, Jastrzebski, Morrone,
  de~Laroussilhe, Gesmundo, Attariyan, and Gelly}]{houlsby2019parameter}
Neil Houlsby, Andrei Giurgiu, Stanislaw Jastrzebski, Bruna Morrone, Quentin
  de~Laroussilhe, Andrea Gesmundo, Mona Attariyan, and Sylvain Gelly. 2019.
\newblock \href {http://proceedings.mlr.press/v97/houlsby19a.html}
  {Parameter-efficient transfer learning for {NLP}}.
\newblock In \emph{Proceedings of the 36th International Conference on Machine
  Learning, {ICML} 2019, 9-15 June 2019, Long Beach, California, {USA}},
  volume~97 of \emph{Proceedings of Machine Learning Research}, pages
  2790--2799. {PMLR}.

\bibitem[{Howard and Ruder(2018)}]{Howard2018}
Jeremy Howard and Sebastian Ruder. 2018.
\newblock \href {https://doi.org/10.18653/v1/P18-1031} {Universal language
  model fine-tuning for text classification}.
\newblock In \emph{Proceedings of the 56th Annual Meeting of the Association
  for Computational Linguistics, {ACL} 2018, Melbourne, Australia, July 15-20,
  2018, Volume 1: Long Papers}, pages 328--339. Association for Computational
  Linguistics.

\bibitem[{Hu et~al.(2022)Hu, Shen, Wallis, Allen{-}Zhu, Li, Wang, Wang, and
  Chen}]{hu2021lora}
Edward~J. Hu, Yelong Shen, Phillip Wallis, Zeyuan Allen{-}Zhu, Yuanzhi Li,
  Shean Wang, Lu~Wang, and Weizhu Chen. 2022.
\newblock \href {https://openreview.net/forum?id=nZeVKeeFYf9} {{LoRA}: Low-rank
  adaptation of large language models}.
\newblock In \emph{The Tenth International Conference on Learning
  Representations, {ICLR} 2022, Virtual Event, April 25-29, 2022}.
  OpenReview.net.

\bibitem[{Jiang et~al.(2021)Jiang, Zhang, Cao, Yin, and
  Zhao}]{Jiang2021_needle}
Haoming Jiang, Danqing Zhang, Tianyu Cao, Bing Yin, and Tuo Zhao. 2021.
\newblock \href {https://doi.org/10.18653/v1/2021.acl-long.140} {Named entity
  recognition with small strongly labeled and large weakly labeled data}.
\newblock In \emph{Proceedings of the 59th Annual Meeting of the Association
  for Computational Linguistics and the 11th International Joint Conference on
  Natural Language Processing, {ACL/IJCNLP} 2021, (Volume 1: Long Papers),
  Virtual Event, August 1-6, 2021}, pages 1775--1789. Association for
  Computational Linguistics.

\bibitem[{Karamanolakis et~al.(2021)Karamanolakis, Mukherjee, Zheng, and
  Awadallah}]{karamanolakis2021astra}
Giannis Karamanolakis, Subhabrata Mukherjee, Guoqing Zheng, and Ahmed~Hassan
  Awadallah. 2021.
\newblock \href {https://doi.org/10.18653/v1/2021.naacl-main.66} {Self-training
  with weak supervision}.
\newblock In \emph{Proceedings of the 2021 Conference of the North American
  Chapter of the Association for Computational Linguistics: Human Language
  Technologies, {NAACL-HLT} 2021, Online, June 6-11, 2021}, pages 845--863.
  Association for Computational Linguistics.

\bibitem[{Krallinger et~al.(2017)Krallinger, Rabal, Akhondi, P{\'e}rez,
  Santamar{\'\i}a, Rodr{\'\i}guez, Tsatsaronis, Intxaurrondo, L{\'o}pez, Nandal
  et~al.}]{krallinger2017_chemprot}
Martin Krallinger, Obdulia Rabal, Saber~A Akhondi, Mart{\i}n~P{\'e}rez
  P{\'e}rez, Jes{\'u}s Santamar{\'\i}a, Gael~P{\'e}rez Rodr{\'\i}guez, Georgios
  Tsatsaronis, Ander Intxaurrondo, Jos{\'e}~Antonio L{\'o}pez, Umesh Nandal,
  et~al. 2017.
\newblock \href
  {https://biocreative.bioinformatics.udel.edu/tasks/biocreative-vi/track-5/}
  {Overview of the {BioCreative VI} chemical-protein interaction track}.
\newblock In \emph{Proceedings of the sixth BioCreative challenge evaluation
  workshop}, volume~1, pages 141--146.

\bibitem[{Li et~al.(2020)Li, Socher, and Hoi}]{li2020dividemix}
Junnan Li, Richard Socher, and Steven C.~H. Hoi. 2020.
\newblock \href {https://openreview.net/forum?id=HJgExaVtwr} {{DivideMix}:
  Learning with noisy labels as semi-supervised learning}.
\newblock In \emph{8th International Conference on Learning Representations,
  {ICLR} 2020, Addis Ababa, Ethiopia, April 26-30, 2020}. OpenReview.net.

\bibitem[{Li et~al.(2021)Li, Xiong, and Hoi}]{li2021comatch}
Junnan Li, Caiming Xiong, and Steven C.~H. Hoi. 2021.
\newblock \href {https://doi.org/10.1109/ICCV48922.2021.00934} {{CoMatch}:
  Semi-supervised learning with contrastive graph regularization}.
\newblock In \emph{2021 {IEEE/CVF} International Conference on Computer Vision,
  {ICCV} 2021, Montreal, QC, Canada, October 10-17, 2021}, pages 9455--9464.
  {IEEE}.

\bibitem[{Li and Roth(2002)}]{li2002_trec}
Xin Li and Dan Roth. 2002.
\newblock \href {https://aclanthology.org/C02-1150} {Learning question
  classifiers}.
\newblock In \emph{{COLING} 2002: The 19th International Conference on
  Computational Linguistics}.

\bibitem[{Liang et~al.(2020)Liang, Yu, Jiang, Er, Wang, Zhao, and
  Zhang}]{Liang2020_bond}
Chen Liang, Yue Yu, Haoming Jiang, Siawpeng Er, Ruijia Wang, Tuo Zhao, and Chao
  Zhang. 2020.
\newblock \href {https://doi.org/10.1145/3394486.3403149} {{BOND:}
  {BERT-}assisted open-domain named entity recognition with distant
  supervision}.
\newblock In \emph{{KDD} '20: The 26th {ACM} {SIGKDD} Conference on Knowledge
  Discovery and Data Mining, Virtual Event, CA, USA, August 23-27, 2020}, pages
  1054--1064. {ACM}.

\bibitem[{Lison et~al.(2021)Lison, Barnes, and Hubin}]{lison-etal-2021-skweak}
Pierre Lison, Jeremy Barnes, and Aliaksandr Hubin. 2021.
\newblock \href {https://doi.org/10.18653/v1/2021.acl-demo.40} {skweak: Weak
  supervision made easy for {NLP}}.
\newblock In \emph{Proceedings of the 59th Annual Meeting of the Association
  for Computational Linguistics and the 11th International Joint Conference on
  Natural Language Processing: System Demonstrations}, pages 337--346, Online.
  Association for Computational Linguistics.

\bibitem[{Liu et~al.(2022)Liu, Tam, Muqeeth, Mohta, Huang, Bansal, and
  Raffel}]{liu2022_few}
Haokun Liu, Derek Tam, Mohammed Muqeeth, Jay Mohta, Tenghao Huang, Mohit
  Bansal, and Colin Raffel. 2022.
\newblock \href
  {http://papers.nips.cc/paper\_files/paper/2022/hash/0cde695b83bd186c1fd456302888454c-Abstract-Conference.html}
  {Few-shot parameter-efficient fine-tuning is better and cheaper than
  in-context learning}.
\newblock In \emph{NeurIPS}.

\bibitem[{Liu et~al.(2019)Liu, Ott, Goyal, Du, Joshi, Chen, Levy, Lewis,
  Zettlemoyer, and Stoyanov}]{liu2019roberta}
Yinhan Liu, Myle Ott, Naman Goyal, Jingfei Du, Mandar Joshi, Danqi Chen, Omer
  Levy, Mike Lewis, Luke Zettlemoyer, and Veselin Stoyanov. 2019.
\newblock \href {http://arxiv.org/abs/1907.11692} {{RoBERTa}: {A} robustly
  optimized {BERT} pretraining approach}.
\newblock \emph{CoRR}, abs/1907.11692.

\bibitem[{Logan~IV et~al.(2022)Logan~IV, Balazevic, Wallace, Petroni, Singh,
  and Riedel}]{logan2021cutting}
Robert Logan~IV, Ivana Balazevic, Eric Wallace, Fabio Petroni, Sameer Singh,
  and Sebastian Riedel. 2022.
\newblock \href {https://doi.org/10.18653/v1/2022.findings-acl.222} {Cutting
  down on prompts and parameters: Simple few-shot learning with language
  models}.
\newblock In \emph{Findings of the Association for Computational Linguistics:
  ACL 2022}, pages 2824--2835, Dublin, Ireland. Association for Computational
  Linguistics.

\bibitem[{Loshchilov and Hutter(2019)}]{loshchilov2017decoupled}
Ilya Loshchilov and Frank Hutter. 2019.
\newblock \href {https://openreview.net/forum?id=Bkg6RiCqY7} {Decoupled weight
  decay regularization}.
\newblock In \emph{7th International Conference on Learning Representations,
  {ICLR} 2019, New Orleans, LA, USA, May 6-9, 2019}. OpenReview.net.

\bibitem[{Lu et~al.(2022)Lu, Bartolo, Moore, Riedel, and
  Stenetorp}]{lu2022fantastically}
Yao Lu, Max Bartolo, Alastair Moore, Sebastian Riedel, and Pontus Stenetorp.
  2022.
\newblock \href {https://doi.org/10.18653/v1/2022.acl-long.556} {Fantastically
  ordered prompts and where to find them: Overcoming few-shot prompt order
  sensitivity}.
\newblock In \emph{Proceedings of the 60th Annual Meeting of the Association
  for Computational Linguistics (Volume 1: Long Papers), {ACL} 2022, Dublin,
  Ireland, May 22-27, 2022}, pages 8086--8098. Association for Computational
  Linguistics.

\bibitem[{Maas et~al.(2011)Maas, Daly, Pham, Huang, Ng, and
  Potts}]{maas2011_imdb}
Andrew~L. Maas, Raymond~E. Daly, Peter~T. Pham, Dan Huang, Andrew~Y. Ng, and
  Christopher Potts. 2011.
\newblock \href {https://aclanthology.org/P11-1015/} {Learning word vectors for
  sentiment analysis}.
\newblock In \emph{Proceedings of the 49th Annual Meeting of the Association
  for Computational Linguistics: Human Language Technologies - Volume 1}, HLT
  '11, page 142–150, USA. Association for Computational Linguistics.

\bibitem[{Miyato et~al.(2018)Miyato, Maeda, Koyama, and Ishii}]{Miyato2018}
Takeru Miyato, Shin-ichi Maeda, Masanori Koyama, and Shin Ishii. 2018.
\newblock \href {https://arxiv.org/abs/1704.03976} {Virtual adversarial
  training: a regularization method for supervised and semi-supervised
  learning}.
\newblock \emph{IEEE transactions on pattern analysis and machine
  intelligence}, 41(8):1979--1993.

\bibitem[{Mosbach et~al.(2021)Mosbach, Andriushchenko, and
  Klakow}]{mosbach2021stability}
Marius Mosbach, Maksym Andriushchenko, and Dietrich Klakow. 2021.
\newblock \href {https://openreview.net/forum?id=nzpLWnVAyah} {On the stability
  of fine-tuning {BERT:} {M}isconceptions, explanations, and strong baselines}.
\newblock In \emph{9th International Conference on Learning Representations,
  {ICLR} 2021, Virtual Event, Austria, May 3-7, 2021}. OpenReview.net.

\bibitem[{Oliver et~al.(2018)Oliver, Odena, Raffel, Cubuk, and
  Goodfellow}]{oliver2018realistic}
Avital Oliver, Augustus Odena, Colin Raffel, Ekin~Dogus Cubuk, and Ian~J.
  Goodfellow. 2018.
\newblock \href
  {https://proceedings.neurips.cc/paper/2018/hash/c1fea270c48e8079d8ddf7d06d26ab52-Abstract.html}
  {Realistic evaluation of deep semi-supervised learning algorithms}.
\newblock In \emph{Advances in Neural Information Processing Systems 31: Annual
  Conference on Neural Information Processing Systems 2018, NeurIPS 2018,
  December 3-8, 2018, Montr{\'{e}}al, Canada}, pages 3239--3250.

\bibitem[{Perez et~al.(2021)Perez, Kiela, and Cho}]{perez2021true}
Ethan Perez, Douwe Kiela, and Kyunghyun Cho. 2021.
\newblock \href
  {https://proceedings.neurips.cc/paper/2021/hash/5c04925674920eb58467fb52ce4ef728-Abstract.html}
  {True few-shot learning with language models}.
\newblock In \emph{Advances in Neural Information Processing Systems 34: Annual
  Conference on Neural Information Processing Systems 2021, NeurIPS 2021,
  December 6-14, 2021, virtual}, pages 11054--11070.

\bibitem[{Peters et~al.(2019)Peters, Ruder, and Smith}]{peters-etal-2019-tune}
Matthew~E. Peters, Sebastian Ruder, and Noah~A. Smith. 2019.
\newblock \href {https://doi.org/10.18653/v1/W19-4302} {To tune or not to tune?
  {A}dapting pretrained representations to diverse tasks}.
\newblock In \emph{Proceedings of the 4th Workshop on Representation Learning
  for NLP (RepL4NLP-2019)}, pages 7--14, Florence, Italy. Association for
  Computational Linguistics.

\bibitem[{Pfeiffer et~al.(2020)Pfeiffer, R{\"u}ckl{\'e}, Poth, Kamath,
  Vuli{\'c}, Ruder, Cho, and Gurevych}]{Pfeiffer2020}
Jonas Pfeiffer, Andreas R{\"u}ckl{\'e}, Clifton Poth, Aishwarya Kamath, Ivan
  Vuli{\'c}, Sebastian Ruder, Kyunghyun Cho, and Iryna Gurevych. 2020.
\newblock \href {https://aclanthology.org/2020.emnlp-demos.7/} {{AdapterHub}: A
  framework for adapting transformers}.
\newblock In \emph{Proceedings of the 2020 Conference on Empirical Methods in
  Natural Language Processing: System Demonstrations}, pages 46--54.

\bibitem[{Pradhan et~al.(2013)Pradhan, Moschitti, Xue, Ng, Bj{\"o}rkelund,
  Uryupina, Zhang, and Zhong}]{pradhan2013_ontonotes}
Sameer Pradhan, Alessandro Moschitti, Nianwen Xue, Hwee~Tou Ng, Anders
  Bj{\"o}rkelund, Olga Uryupina, Yuchen Zhang, and Zhi Zhong. 2013.
\newblock \href {https://aclanthology.org/W13-3516} {Towards robust linguistic
  analysis using {O}nto{N}otes}.
\newblock In \emph{Proceedings of the Seventeenth Conference on Computational
  Natural Language Learning}, pages 143--152, Sofia, Bulgaria. Association for
  Computational Linguistics.

\bibitem[{Ratner et~al.(2017)Ratner, Bach, Ehrenberg, Fries, Wu, and
  R\'{e}}]{ratner2017_snorkel}
Alexander Ratner, Stephen~H. Bach, Henry Ehrenberg, Jason Fries, Sen Wu, and
  Christopher R\'{e}. 2017.
\newblock \href {https://doi.org/10.14778/3157794.3157797} {Snorkel: Rapid
  training data creation with weak supervision}.
\newblock \emph{Proc. VLDB Endow.}, 11(3):269–282.

\bibitem[{Ren et~al.(2018)Ren, Zeng, Yang, and Urtasun}]{ren2018learning}
Mengye Ren, Wenyuan Zeng, Bin Yang, and Raquel Urtasun. 2018.
\newblock \href {http://proceedings.mlr.press/v80/ren18a.html} {Learning to
  reweight examples for robust deep learning}.
\newblock In \emph{Proceedings of the 35th International Conference on Machine
  Learning, {ICML} 2018, Stockholmsm{\"{a}}ssan, Stockholm, Sweden, July 10-15,
  2018}, volume~80 of \emph{Proceedings of Machine Learning Research}, pages
  4331--4340. {PMLR}.

\bibitem[{Ren et~al.(2020)Ren, Li, Su, Kartchner, Mitchell, and
  Zhang}]{ren2020_denoising}
Wendi Ren, Yinghao Li, Hanting Su, David Kartchner, Cassie Mitchell, and Chao
  Zhang. 2020.
\newblock \href {https://doi.org/10.18653/v1/2020.findings-emnlp.334}
  {Denoising multi-source weak supervision for neural text classification}.
\newblock In \emph{Findings of the Association for Computational Linguistics:
  EMNLP 2020}, pages 3739--3754, Online. Association for Computational
  Linguistics.

\bibitem[{Schick and Sch{\"u}tze(2022)}]{schick2022true}
Timo Schick and Hinrich Sch{\"u}tze. 2022.
\newblock \href {https://doi.org/10.1162/tacl_a_00485} {True few-shot learning
  with {P}rompts{---}{A} real-world perspective}.
\newblock \emph{Transactions of the Association for Computational Linguistics},
  10:716--731.

\bibitem[{Shu et~al.(2019)Shu, Xie, Yi, Zhao, Zhou, Xu, and Meng}]{shu2019meta}
Jun Shu, Qi~Xie, Lixuan Yi, Qian Zhao, Sanping Zhou, Zongben Xu, and Deyu Meng.
  2019.
\newblock \href
  {https://proceedings.neurips.cc/paper/2019/hash/e58cc5ca94270acaceed13bc82dfedf7-Abstract.html}
  {Meta-weight-net: Learning an explicit mapping for sample weighting}.
\newblock In \emph{Advances in Neural Information Processing Systems 32: Annual
  Conference on Neural Information Processing Systems 2019, NeurIPS 2019,
  December 8-14, 2019, Vancouver, BC, Canada}, pages 1917--1928.

\bibitem[{Smith et~al.(2022)Smith, Fries, Hancock, and
  Bach}]{smith2022language}
Ryan Smith, Jason~A. Fries, Braden Hancock, and Stephen~H. Bach. 2022.
\newblock \href {https://doi.org/10.48550/arXiv.2205.02318} {Language models in
  the loop: Incorporating prompting into weak supervision}.
\newblock \emph{CoRR}, abs/2205.02318.

\bibitem[{Stephan et~al.(2022)Stephan, Kougia, and Roth}]{Stephan2022}
Andreas Stephan, Vasiliki Kougia, and Benjamin Roth. 2022.
\newblock \href {https://aclanthology.org/2022.findings-emnlp.288} {{S}ep{LL}:
  Separating latent class labels from weak supervision noise}.
\newblock In \emph{Findings of the Association for Computational Linguistics:
  EMNLP 2022}, pages 3918--3929, Abu Dhabi, United Arab Emirates. Association
  for Computational Linguistics.

\bibitem[{T{\"{a}}nzer et~al.(2022)T{\"{a}}nzer, Ruder, and
  Rei}]{tanzer2022memorisation}
Michael T{\"{a}}nzer, Sebastian Ruder, and Marek Rei. 2022.
\newblock \href {https://doi.org/10.18653/v1/2022.acl-long.521} {Memorisation
  versus generalisation in pre-trained language models}.
\newblock In \emph{Proceedings of the 60th Annual Meeting of the Association
  for Computational Linguistics (Volume 1: Long Papers), {ACL} 2022, Dublin,
  Ireland, May 22-27, 2022}, pages 7564--7578. Association for Computational
  Linguistics.

\bibitem[{Tarvainen and Valpola(2017)}]{tarvainen2017mean}
Antti Tarvainen and Harri Valpola. 2017.
\newblock \href {https://openreview.net/forum?id=ry8u21rtl} {Mean teachers are
  better role models: Weight-averaged consistency targets improve
  semi-supervised deep learning results}.
\newblock In \emph{5th International Conference on Learning Representations,
  {ICLR} 2017, Toulon, France, April 24-26, 2017, Workshop Track Proceedings}.
  OpenReview.net.

\bibitem[{Tjong Kim~Sang and De~Meulder(2003)}]{tjong2003_conll}
Erik~F. Tjong Kim~Sang and Fien De~Meulder. 2003.
\newblock \href {https://aclanthology.org/W03-0419} {Introduction to the
  {C}o{NLL}-2003 shared task: Language-independent named entity recognition}.
\newblock In \emph{Proceedings of the Seventh Conference on Natural Language
  Learning at {HLT}-{NAACL} 2003}, pages 142--147.

\bibitem[{Warstadt et~al.(2020)Warstadt, Zhang, Li, Liu, and
  Bowman}]{warstadt2020learning}
Alex Warstadt, Yian Zhang, Xiaocheng Li, Haokun Liu, and Samuel~R. Bowman.
  2020.
\newblock \href {https://doi.org/10.18653/v1/2020.emnlp-main.16} {Learning
  which features matter: {R}o{BERT}a acquires a preference for linguistic
  generalizations (eventually)}.
\newblock In \emph{Proceedings of the 2020 Conference on Empirical Methods in
  Natural Language Processing (EMNLP)}, pages 217--235, Online. Association for
  Computational Linguistics.

\bibitem[{Wolf et~al.(2020)Wolf, Debut, Sanh, Chaumond, Delangue, Moi, Cistac,
  Rault, Louf, Funtowicz, Davison, Shleifer, von Platen, Ma, Jernite, Plu, Xu,
  Scao, Gugger, Drame, Lhoest, and Rush}]{Wolf2020}
Thomas Wolf, Lysandre Debut, Victor Sanh, Julien Chaumond, Clement Delangue,
  Anthony Moi, Pierric Cistac, Tim Rault, Rémi Louf, Morgan Funtowicz, Joe
  Davison, Sam Shleifer, Patrick von Platen, Clara Ma, Yacine Jernite, Julien
  Plu, Canwen Xu, Teven~Le Scao, Sylvain Gugger, Mariama Drame, Quentin Lhoest,
  and Alexander~M. Rush. 2020.
\newblock \href {https://www.aclweb.org/anthology/2020.emnlp-demos.6}
  {Transformers: State-of-the-art natural language processing}.
\newblock In \emph{Proceedings of the 2020 Conference on Empirical Methods in
  Natural Language Processing: System Demonstrations}, pages 38--45, Online.
  Association for Computational Linguistics.

\bibitem[{Xie et~al.(2020)Xie, Dai, Hovy, Luong, and Le}]{xie2020unsupervised}
Qizhe Xie, Zihang Dai, Eduard~H. Hovy, Thang Luong, and Quoc Le. 2020.
\newblock \href
  {https://proceedings.neurips.cc/paper/2020/hash/44feb0096faa8326192570788b38c1d1-Abstract.html}
  {Unsupervised data augmentation for consistency training}.
\newblock In \emph{Advances in Neural Information Processing Systems 33: Annual
  Conference on Neural Information Processing Systems 2020, NeurIPS 2020,
  December 6-12, 2020, virtual}.

\bibitem[{Yu et~al.(2021)Yu, Zuo, Jiang, Ren, Zhao, and Zhang}]{Yu2021_cosine}
Yue Yu, Simiao Zuo, Haoming Jiang, Wendi Ren, Tuo Zhao, and Chao Zhang. 2021.
\newblock \href {https://doi.org/10.18653/v1/2021.naacl-main.84} {Fine-tuning
  pre-trained language model with weak supervision: {A} contrastive-regularized
  self-training approach}.
\newblock In \emph{Proceedings of the 2021 Conference of the North American
  Chapter of the Association for Computational Linguistics: Human Language
  Technologies, {NAACL-HLT} 2021, Online, June 6-11, 2021}, pages 1063--1077.
  Association for Computational Linguistics.

\bibitem[{Zaken et~al.(2022)Zaken, Goldberg, and Ravfogel}]{zaken2022bitfit}
Elad~Ben Zaken, Yoav Goldberg, and Shauli Ravfogel. 2022.
\newblock \href {https://doi.org/10.18653/v1/2022.acl-short.1} {{BitFit}:
  Simple parameter-efficient fine-tuning for transformer-based masked
  language-models}.
\newblock In \emph{Proceedings of the 60th Annual Meeting of the Association
  for Computational Linguistics (Volume 2: Short Papers), {ACL} 2022, Dublin,
  Ireland, May 22-27, 2022}, pages 1--9. Association for Computational
  Linguistics.

\bibitem[{Zhang et~al.(2021{\natexlab{a}})Zhang, Wang, Hou, Wu, Wang, Okumura,
  and Shinozaki}]{zhang2021flexmatch}
Bowen Zhang, Yidong Wang, Wenxin Hou, Hao Wu, Jindong Wang, Manabu Okumura, and
  Takahiro Shinozaki. 2021{\natexlab{a}}.
\newblock \href
  {https://proceedings.neurips.cc/paper/2021/hash/995693c15f439e3d189b06e89d145dd5-Abstract.html}
  {{FlexMatch}: Boosting semi-supervised learning with curriculum pseudo
  labeling}.
\newblock In \emph{Advances in Neural Information Processing Systems 34: Annual
  Conference on Neural Information Processing Systems 2021, NeurIPS 2021,
  December 6-14, 2021, virtual}, pages 18408--18419.

\bibitem[{Zhang et~al.(2017)Zhang, Bengio, Hardt, Recht, and
  Vinyals}]{zhang2017_rethinking}
Chiyuan Zhang, Samy Bengio, Moritz Hardt, Benjamin Recht, and Oriol Vinyals.
  2017.
\newblock \href {https://openreview.net/forum?id=Sy8gdB9xx} {Understanding deep
  learning requires rethinking generalization}.
\newblock In \emph{5th International Conference on Learning Representations,
  {ICLR} 2017, Toulon, France, April 24-26, 2017, Conference Track
  Proceedings}. OpenReview.net.

\bibitem[{Zhang et~al.(2022)Zhang, Hsieh, Yu, Zhang, and
  Ratner}]{zhang2022_weak_survey}
Jieyu Zhang, Cheng{-}Yu Hsieh, Yue Yu, Chao Zhang, and Alexander Ratner. 2022.
\newblock \href {http://arxiv.org/abs/2202.05433} {A survey on programmatic
  weak supervision}.
\newblock \emph{CoRR}, abs/2202.05433.

\bibitem[{Zhang et~al.(2021{\natexlab{b}})Zhang, Yu, NameError, Wang, Yang,
  Yang, and Ratner}]{zhang2021_wrench}
Jieyu Zhang, Yue Yu, NameError, Yujing Wang, Yaming Yang, Mao Yang, and
  Alexander Ratner. 2021{\natexlab{b}}.
\newblock \href
  {https://datasets-benchmarks-proceedings.neurips.cc/paper/2021/hash/1c9ac0159c94d8d0cbedc973445af2da-Abstract-round2.html}
  {{WRENCH:} {A} comprehensive benchmark for weak supervision}.
\newblock In \emph{Proceedings of the Neural Information Processing Systems
  Track on Datasets and Benchmarks 1, NeurIPS Datasets and Benchmarks 2021,
  December 2021, virtual}.

\bibitem[{Zhang et~al.(2015)Zhang, Zhao, and LeCun}]{zhang2015character}
Xiang Zhang, Junbo Zhao, and Yann LeCun. 2015.
\newblock \href
  {https://papers.nips.cc/paper_files/paper/2015/hash/250cf8b51c773f3f8dc8b4be867a9a02-Abstract.html}
  {Character-level convolutional networks for text classification}.
\newblock \emph{Advances in neural information processing systems}, 28.

\bibitem[{Zhao et~al.(2021)Zhao, Wallace, Feng, Klein, and
  Singh}]{zhao2021calibrate}
Zihao Zhao, Eric Wallace, Shi Feng, Dan Klein, and Sameer Singh. 2021.
\newblock \href {http://proceedings.mlr.press/v139/zhao21c.html} {Calibrate
  before use: Improving few-shot performance of language models}.
\newblock In \emph{Proceedings of the 38th International Conference on Machine
  Learning, {ICML} 2021, 18-24 July 2021, Virtual Event}, volume 139 of
  \emph{Proceedings of Machine Learning Research}, pages 12697--12706. {PMLR}.

\bibitem[{Zheng et~al.(2021)Zheng, Awadallah, and Dumais}]{zheng2021meta}
Guoqing Zheng, Ahmed~Hassan Awadallah, and Susan~T. Dumais. 2021.
\newblock \href {https://ojs.aaai.org/index.php/AAAI/article/view/17319} {Meta
  label correction for noisy label learning}.
\newblock In \emph{Thirty-Fifth {AAAI} Conference on Artificial Intelligence,
  {AAAI} 2021, Thirty-Third Conference on Innovative Applications of Artificial
  Intelligence, {IAAI} 2021, The Eleventh Symposium on Educational Advances in
  Artificial Intelligence, {EAAI} 2021, Virtual Event, February 2-9, 2021},
  pages 11053--11061. {AAAI} Press.

\bibitem[{Zheng et~al.(2022{\natexlab{a}})Zheng, Karamanolakis, Shu, and
  Awadallah}]{Zheng2022_walnut}
Guoqing Zheng, Giannis Karamanolakis, Kai Shu, and Ahmed Awadallah.
  2022{\natexlab{a}}.
\newblock \href {https://doi.org/10.18653/v1/2022.naacl-main.64} {{WALNUT}: A
  benchmark on semi-weakly supervised learning for natural language
  understanding}.
\newblock In \emph{Proceedings of the 2022 Conference of the North American
  Chapter of the Association for Computational Linguistics: Human Language
  Technologies}, pages 873--899, Seattle, United States. Association for
  Computational Linguistics.

\bibitem[{Zheng et~al.(2022{\natexlab{b}})Zheng, Zhou, Qian, Ding, Liao, Jian,
  Salakhutdinov, Tang, Ruder, and Yang}]{zheng2021fewnlu}
Yanan Zheng, Jing Zhou, Yujie Qian, Ming Ding, Chonghua Liao, Li~Jian, Ruslan
  Salakhutdinov, Jie Tang, Sebastian Ruder, and Zhilin Yang.
  2022{\natexlab{b}}.
\newblock \href {https://doi.org/10.18653/v1/2022.acl-long.38} {{FewNLU}:
  Benchmarking state-of-the-art methods for few-shot natural language
  understanding}.
\newblock In \emph{Proceedings of the 60th Annual Meeting of the Association
  for Computational Linguistics (Volume 1: Long Papers), {ACL} 2022, Dublin,
  Ireland, May 22-27, 2022}, pages 501--516. Association for Computational
  Linguistics.

\bibitem[{Zhou et~al.(2022)Zhou, Li, Shang, Luo, Zhan, Hu, Zhang, Jiang, Cao,
  Yu, Jiang, Liu, and Chen}]{zhou2022hyperlink}
Jiawei Zhou, Xiaoguang Li, Lifeng Shang, Lan Luo, Ke~Zhan, Enrui Hu, Xinyu
  Zhang, Hao Jiang, Zhao Cao, Fan Yu, Xin Jiang, Qun Liu, and Lei Chen. 2022.
\newblock \href {https://doi.org/10.18653/v1/2022.acl-long.493}
  {Hyperlink-induced pre-training for passage retrieval in open-domain question
  answering}.
\newblock In \emph{Proceedings of the 60th Annual Meeting of the Association
  for Computational Linguistics (Volume 1: Long Papers), {ACL} 2022, Dublin,
  Ireland, May 22-27, 2022}, pages 7135--7146. Association for Computational
  Linguistics.

\bibitem[{Zhu et~al.(2022)Zhu, Hedderich, Zhai, Adelani, and
  Klakow}]{zhu2022bert}
Dawei Zhu, Michael~A. Hedderich, Fangzhou Zhai, David~Ifeoluwa Adelani, and
  Dietrich Klakow. 2022.
\newblock \href {https://doi.org/10.18653/v1/2022.insights-1.8} {Is {BERT}
  robust to label noise? {A} study on learning with noisy labels in text
  classification}.
\newblock In \emph{Proceedings of the Third Workshop on Insights from Negative
  Results in NLP, Insights@ACL 2022, Dublin, Ireland, May 26, 2022}, pages
  62--67. Association for Computational Linguistics.

\bibitem[{Zhu et~al.(2023)Zhu, Shen, Hedderich, and Klakow}]{Zhu2023}
Dawei Zhu, Xiaoyu Shen, Michael~A. Hedderich, and Dietrich Klakow. 2023.
\newblock \href {https://aclanthology.org/2023.eacl-main.74} {Meta
  self-refinement for robust learning with weak supervision}.
\newblock In \emph{Proceedings of the 17th Conference of the European Chapter
  of the Association for Computational Linguistics, {EACL} 2023, Dubrovnik,
  Croatia, May 2-6, 2023}, pages 1043--1058. Association for Computational
  Linguistics.

\end{thebibliography}

\clearpage
\appendix

\section{Datasets}
\label{sec:appendix:dataset_details}

\begin{table*}[ht]
    \tiny
    \centering
    \resizebox{1.\textwidth}{!}{%
    \begin{tabular}{llccccccccccc}
    \toprule 
        & & & & & & \multicolumn{3}{l}{\textbf{Avg. over labeling functions (LFs)}} & & \\\cline{6-9}\rule{0pt}{3ex} 
       \textbf{Dataset} & \textbf{Task} & \textbf{\#Classes} & \textbf{\#LFs} & \textbf{ \%Ovr. Coverage} & \textbf{\%Coverage} & \textbf{\%Overlap} & \textbf{\%Conflict} & \textbf{\%Prec.} & \textbf{MV} & \textbf{\#Train} & \textbf{\#Dev} & \textbf{\#Test} \\
       \midrule
        AGNews & News Class. & 4 & 9 & 69.08 & 10.34 & 5.05 & 2.43 & 81.66 & 81.23 & 96,000 & 12,000 & 12,000 \\\rule{0pt}{2ex} 
        IMDb & Movie Sentiment Class. &  2 & 5 & 87.58 & 23.60 & 11.60 & 4.50 & 69.88 & 73.86 & 20,000 & 2,500 & 2,500 \\\rule{0pt}{2ex} 
        Yelp & Business Sentiment  Class. & 2 & 8 & 82.78 & 18.34 & 13.58 & 4.94 & 73.05 & 73.31 & 30,400 & 3,800 & 3,800 \\\rule{0pt}{2ex} 
        TREC & Question Class. & 6 & 68 & 95.13 & 2.55 & 1.82 & 0.84 & 75.92 & 62.58 & 4,965 & 500 & 500 \\\rule{0pt}{2ex} 
        SemEval & Web Text Relation Class. & 9 & 164 & 100.00 & 0.77 & 0.32 & 0.14 & 97.69 & 77.33 & 1,749 & 200 & 692 \\\rule{0pt}{2ex} 
        ChemProt & Chemical Relation Class. & 10 & 26 & 85.62 & 5.93 & 4.40 & 3.95 & 46.65 & 55.12 & 12,861 & 1,607 & 1,607 \\\rule{0pt}{2ex} 
        CoNLL-03 & English News NER &  4 & 16 & 100 & 100 & 4.30 & 1.44 & 72.19 & 60.38 & 14,041 & 3250 & 3453 \\\rule{0pt}{2ex} 
        OntoNotes 5.0 & Multi-Domain NER & 18 & 17 & 100 & 100 & 1.55 & 0.54 & 54.84 & 58.92 & 115,812 & 5,000 & 22,897 \\
        \bottomrule
    \end{tabular}
    }%
    \caption{Detailed data statistics. Note that `Class.' is an abbreviation for classification. Coverage is the amount of samples a labeling function (LF) matches. For NER datasets, labeling functions return an entity or "O" thus coverage is always 100\%. Overlap asks how many samples have at least 2 matching labeling functions. MV (majority vote) performance is given as F1-score for the NER datasets and as accuracy on the test set otherwise.}
    \label{tab:app_full_data_stats}
\end{table*}

In the following, we give a more comprehensive description of the datasets used. A subset of the commonly used WRENCH \citep{zhang2021_wrench} benchmark is used, covering various aspects such as task type, coverage and dataset size.
There is a total of four classification, two relation extraction and two sequence labeling datasets. See Table \ref{tab:app_full_data_stats} for a detailed set of data statistics.

\paragraph{AGNews}\citep{zhang2015character} is a topic classification dataset. The task is to classify news articles into four topics, namely world, sports, business and Sci-Fi/technology. Each labeling function is composed of multiple keywords to search for. The number of keywords differs from a few up to dozens.

\paragraph{IMDb}\citep{maas2011_imdb} is a dataset of movie reviews sampled from the IMDb website. The task is binary sentiment analysis. The labeling functions are composed of keyword searches and regular expressions.

\paragraph{Yelp}\citep{zhang2015character} is another sentiment analysis dataset, containing crowd-sourced business reviews. The labeling functions are created using keywords and a lexicon-based sentiment analysis library. 

\paragraph{TREC}\citep{li2002_trec} is a question classification dataset, i.e., it asks what type of response is expected. The labels are abbreviation, description and abstract concepts, entities, human beings, locations or numeric values. The labeling functions are created using regular expressions and make a lot of use of question words such as "what", "where" or "who".

\paragraph{SemEval}\citep{hendrickx2010_semeval} is a relation classification dataset, using nine relation types. Examples for relation labels are cause-effect, entity-origin or message-topic. Labeling functions are created using entities within a regular expression.

\paragraph{ChemProt}\citep{krallinger2017_chemprot} is another relation classification dataset, focusing on chemical research literature. It contains ten different types of relations, for example chemical-protein relations such as ``biological properties upregulator''. The labeling functions are created using rules.

\paragraph{CoNLL-03}\citep{tjong2003_conll} is a named entity recognition (NER) dataset, with labels for the entities "person", "location", "organization", and "miscellaneous". Labeling functions are built using previously trained keywords, regular expressions and NER models.

\paragraph{OntoNotes 5.0}\citep{pradhan2013_ontonotes} is a another NER dataset, using more fine-grained entities as CoNLL-03. Here, a subset of the CoNLL weak labeling sources is combined with keyword and regular expression based weak labeling sources.

\section{Labeling functions}
\label{sec:appendix:rules}

\begin{table*}
\centering
\resizebox{1.\textwidth}{!}{%
    \begin{tabular}{ll}
\toprule 
\textbf{Label} & \textbf{Labeling Function} \\
\midrule
POS & beautiful, handsome, talented \\
NEG & than this, than the film, than the movie \\
POS & .*(highly|do|would|definitely|certainly|strongly|i|we).*(recommend|nominate).* \\
POS & .*(high|timeless|priceless|HAS|great|real|instructive).*(value|quality|meaning|significance).* \\
\bottomrule
\end{tabular}
}%
\caption{Examples of two keyword based and two regular expression based rules for the IMDb dataset.}
\label{tab:app_imdb_rules}
\end{table*}

\begin{table*}
\small
\centering
\begin{tabular}{ll}
\toprule
        Label &                                                                                                                                                                               Labeling Function \\
\midrule
 ABBREVIATION &                                                                                               ( |\textasciicircum )(what|what)[\textasciicircum \textbackslash w]* (\textbackslash w+ )\{0,1\}(does|does)[\textasciicircum \textbackslash w]* ([\textasciicircum \textbackslash s]+ )*(stand for)[\textasciicircum \textbackslash w]*( |\$) \\
 DESCRIPTION &                                                                                                              ( |\textasciicircum )(explain|describe|how|how)[\textasciicircum \textbackslash w]* (\textbackslash w+ )\{0,1\}(can|can)[\textasciicircum \textbackslash w]*( |\$) \\
       ENTITY &                                                                                             ( |\textasciicircum )(which|what|what)[\textasciicircum \textbackslash w]* ([\textasciicircum \textbackslash s]+ )*(organization|trust|company|company)[\textasciicircum \textbackslash w]*( |\$) \\
        HUMAN &                                                                                                                                                          ( |\textasciicircum )(who|who)[\textasciicircum \textbackslash w]*( |\$) \\
     LOCATION &                                                                                                ( |\textasciicircum )(which|what|where|where)[\textasciicircum \textbackslash w]* ([\textasciicircum \textbackslash s]+ )*(situated|located|located)[\textasciicircum \textbackslash w]*( |\$) \\
      NUMERIC &                                                                                                                 ( |\textasciicircum )(by how|how|how)[\textasciicircum \textbackslash w]* (\textbackslash w+ )\{0,1\}(much|many|many)[\textasciicircum \textbackslash w]*( |\$) \\
\bottomrule
\end{tabular}
\caption{Rules for the TREC dataset. For each label a representative labeling function is given.}
\label{tab:app_trec_rules}
\end{table*}

\begin{table*}
\centering
\begin{tabular}{ll}
\toprule
        Label &                                                                                                                                                                               Labeling Function \\
\midrule
Cause-Effect(e1,e2) &                                    SUBJ-O caused OBJ-O \\
Component-Whole(e1,e2) &                          SUBJ-O is a part of the OBJ-O \\
Content-Container(e1,e2) &                  SUBJ-O was contained in a large OBJ-O \\
Entity-Destination(e1,e2) &                                      SUBJ-O into OBJ-O \\
Entity-Origin(e1,e2) &                          SUBJ-O emerged from the OBJ-O \\
Instrument-Agency(e2,e1) &                                  SUBJ-O took the OBJ-O \\
Member-Collection(e2,e1) &                              SUBJ-O of different OBJ-O \\
Message-Topic(e1,e2) &                           SUBJ-O states that the OBJ-O \\
Product-Producer(e1,e2) &                        SUBJ-O created by the OBJ-TITLE \\
\bottomrule
\end{tabular}
\caption{One labeling function for each label of the SemEval dataset. Here e1 and e2 are entities which are already available in the dataset.}
\label{tab:app_semeval_rules}
\end{table*}

\begin{table*}
\small
\centering
\begin{tabular}{ll}
\toprule
        Label &                                                                                                                                                                               Labeling Function \\
\midrule
PERSON &     RegEx searching list one of 7559 first names, followed by an upper-cased word      \\
        LOCATION &  List of 15205 places \\
     ORGANIZATION &  WTO, Starbucks, mcdonald, google, Baidu, IBM, Sony, Nikon        \\
      MISCELLANEOUS &  List of countries, languages, events and facilities \\
\bottomrule
\end{tabular}
\caption{For each label, one labeling function of the CoNLL-03 dataset is displayed.}
\label{tab:app_conll_rules}
\end{table*}

Weak labeling sources are often abstracted as labeling functions and vary in aspects such as coverage, precision, or overlap \cite{ratner2017_snorkel, karamanolakis2021astra}. To showcase how the weak labeling process works, a selection of examples of labeling functions is presented. More specifically, we provide examples of rules for the two classification datasets IMDb (Table \ref{tab:app_imdb_rules}) and TREC (Table \ref{tab:app_trec_rules}), the relation classification dataset SemEval (Table \ref{tab:app_semeval_rules}) and the NER dataset CoNLL-03 (Table \ref{tab:app_conll_rules}).

\section{Overall implementation details}
\label{sec:appendix:implementation_details}
This section summarizes the overall implementation details of WSL approaches used in our paper. Refer to Appendix \ref{sec:appendix:ft_on_clean_samples} for hyperparameter configurations of PEFT approaches. We use the PyTorch framework\footnote{\url{https://pytorch.org/}} to implement all approaches discussed in the paper. Hugging Face \cite{Wolf2020} is used for downloading and training the RoBERTa-base model. AdapterHub \cite{Pfeiffer2020} is used for implementing parameter-efficient fine-tuning.

\paragraph{Hyperparameters}
In this paper, we implemented five WSL methods: FT \cite{devlin2019bert}, L2R \cite{ren2018learning}, MLC \cite{zheng2021meta}, BOND \cite{Liang2020_bond}, and COSINE \cite{Yu2021_cosine}. We report the search ranges of the hyperparameters in Table \ref{tab:appendix:hyperparameter_search_range}.

\begin{table*}
\centering
\begin{minipage}{0.48\linewidth}
        \centering
    \begin{tabular}{lc}
        \toprule 
        \textbf{Hyperparameter} & \textbf{Search Range} \\\midrule  
         Learning rate & 2e-5, 3e-5, 5e-5\\
         Warm-up steps & 50, 100, 200\\
         \bottomrule 
    \end{tabular}
        \subcaption{FT (for both training on clean or weak labels)}
\end{minipage}%
\begin{minipage}{0.48\linewidth}
        \centering
        \begin{tabular}{lc}
        \toprule 
        \textbf{Hyperparameter} & \textbf{Search Range} \\\midrule  
         Learning rate & 2e-5, 3e-5, 5e-5\\
         Meta-learning rate & 1e-4, 2e-5, 1e-5\\
         \bottomrule 
    \end{tabular}
        \subcaption{L2R}
\end{minipage}%

\begin{minipage}{0.48\linewidth}
        \centering
        \begin{tabular}{lc}
        \toprule 
        \textbf{Hyperparameter} & \textbf{Search Range} \\\midrule  
         Learning rate & 2e-5, 3e-5, 5e-5\\
         Meta-learning rate & 1e-4, 2e-5, 1e-5\\
         hdim & 512, 768\\
         \bottomrule 
    \end{tabular}
        \subcaption{MLC}
\end{minipage}%
\begin{minipage}{0.48\linewidth}
        \centering
        \begin{tabular}{cc}
        \toprule
        \textbf{Hyperparameter} & \textbf{Search Range} \\\midrule  
        Learning rate & 2e-5, 3e-5, 5e-5\\
        $T_{1}$ & 5000 \\
        $T_{2}$ & 5000 \\
        $T_{3}$ & 50, 100, 300, 500 \\
        Confidence threshold & 0.1, 0.3, 0.5, 0.7, 0.8, 0.9 \\ \bottomrule
        \end{tabular}
        \subcaption{BOND}
\end{minipage}%

\begin{minipage}{0.48\linewidth}
        \centering
        \begin{tabular}{cc}
        \toprule
        \textbf{Hyperparameter} & \textbf{Search Range} \\\midrule
        Learning rate & 2e-5, 3e-5, 5e-5\\
        $T_{1}$ & 5000 \\
        $T_{2}$ & 5000 \\
        $T_{3}$ & 50, 100, 300, 500 \\
        Distance measure & cosine \\
        Regularization factor & 0.05 0.1 0.2 \\
        Confidence threshold & 0.1, 0.3, 0.5, 0.7, 0.8, 0.9 \\ \bottomrule
        \end{tabular}
        \subcaption{COSINE}
\end{minipage}%
\caption{
The search range of the hyperparameters of the five WSL approaches considered in the paper. For BOND and COSINE, we set $T_{1}$ and $T_{2}$ to constant values, because we stop training once early-stopping is triggered.
}
\label{tab:appendix:hyperparameter_search_range}
\end{table*}

We do not search for batch size as we find it has minor effects on the final performance. Instead, a batch size of 32 is used across experiments. Also, RoBERTa-base \cite{liu2019roberta} is used as the backbone PLM and AdamW \cite{loshchilov2017decoupled} is the optimizer used across all methods.

\begin{table*}[]
\begin{tabular}{lcccccccc}
\hline
       & \multicolumn{1}{l}{AGNews} & \multicolumn{1}{l}{IMDb} & \multicolumn{1}{l}{Yelp} & \multicolumn{1}{l}{TREC} & \multicolumn{1}{l}{SemEval} & \multicolumn{1}{l}{ChemProt} & \multicolumn{1}{l}{CoNLL-03} & \multicolumn{1}{l}{OntoNotes 5.0} \\ \hline
FT     & 0.2                        & 0.2                      & 0.2                      & 0.1                      & 0.1                         & 0.2                          & 0.2                          & 0.5                               \\
L2R    & 2.0                          & 1.2                      & 1.5                      & 0.3                      & 0.3                         & 0.4                          & 0.9                          & 1.2                               \\
MLC    & 1.2                        & 0.8                      & 1.2                      & 0.3                      & 0.2                         & 0.5                          & 1.2                          & 1.0                               \\
BOND   & 0.5                        & 0.2                      & 0.5                      & 0.1                      & 0.1                         & 0.2                          & 0.4                          & 1.1                               \\
COSINE & 0.6                        & 0.2                      & 0.6                      & 0.2                      & 0.2                         & 0.3                          & 0.5                          & 1.5                               \\ \hline
\end{tabular}
\caption{Running time in hours of each WSL method when trained on a weakly labeled training set. Since we also track the validation and test performance during training, the training time reported here actually overestimates the training time required for each method.}
\label{tab:appendix:running_time}
\end{table*}

\paragraph{Computing infrastructure and training cost}
We use Nvidia V100-32 GPUs for training deep learning models. All WSL approaches studied in this paper can fit into one single GPU. We report the training time of the WSL methods in Table \ref{tab:appendix:running_time}.

\begin{figure*}[t]
        \centering
        \small
    \begin{tabular}[c]{ccccc}
     & L2R & MLC & BOND & COSINE \\ 
     \raisebox{30pt}{\rotatebox[origin=c]{90}{AGNews}}&
     \begin{subfigure}[b]{0.45\columnwidth}
         \centering
         \includegraphics[width=\columnwidth]{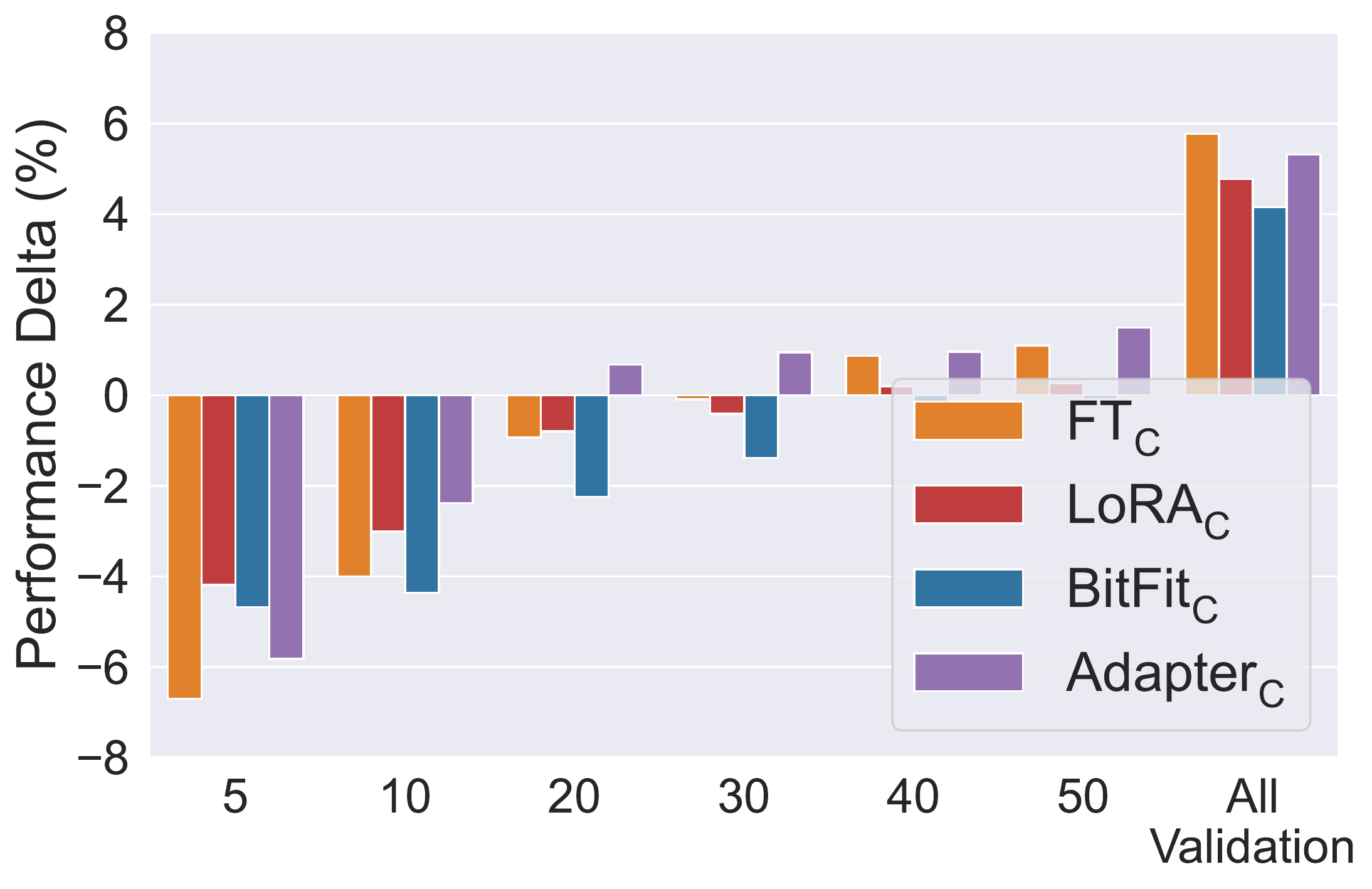}
     \end{subfigure} &%
          \begin{subfigure}[b]{0.45\columnwidth}
         \centering
         \includegraphics[width=\columnwidth]{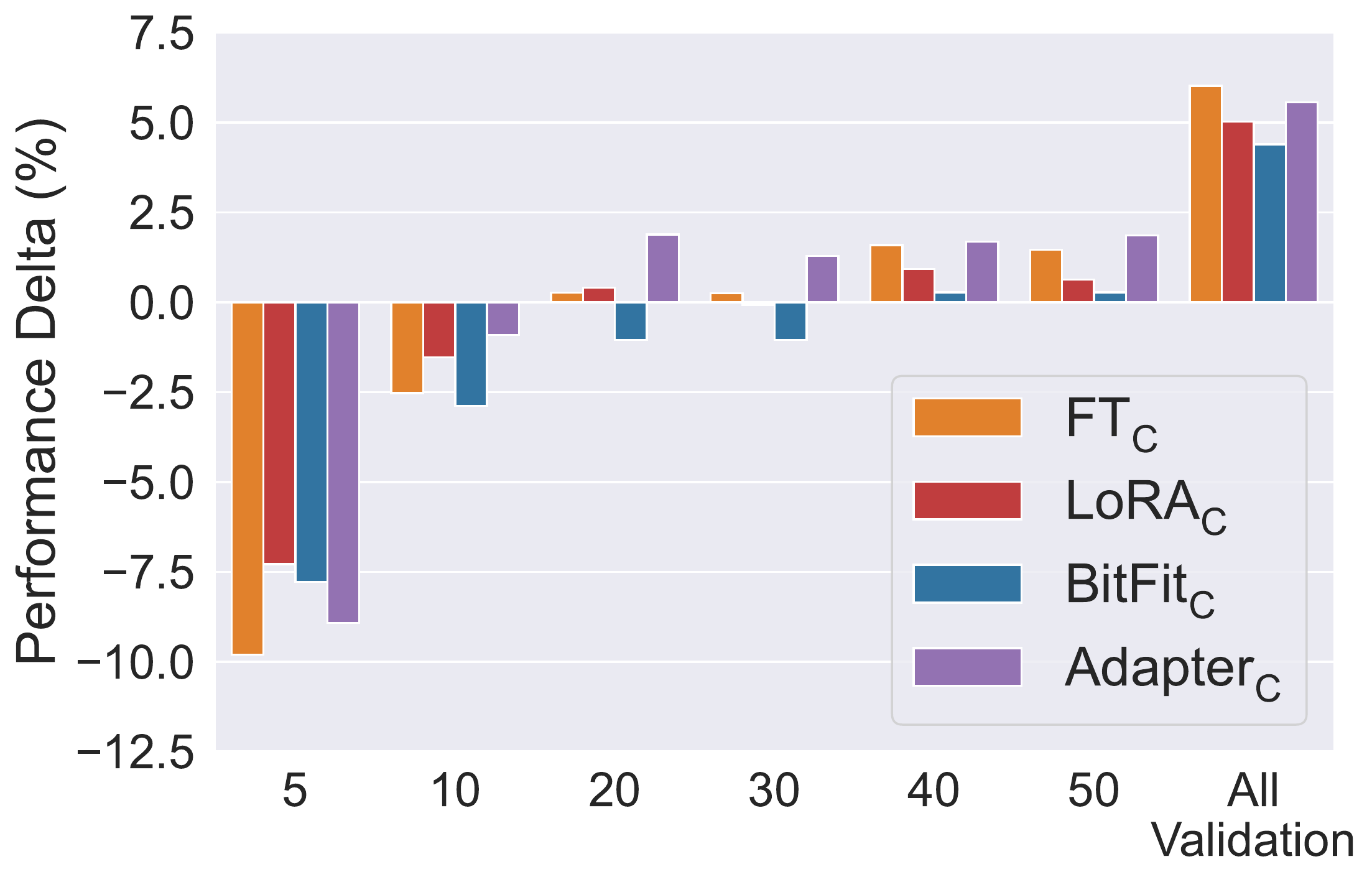}

     \end{subfigure} &%
          \begin{subfigure}[b]{0.45\columnwidth}
         \centering
         \includegraphics[width=\columnwidth]{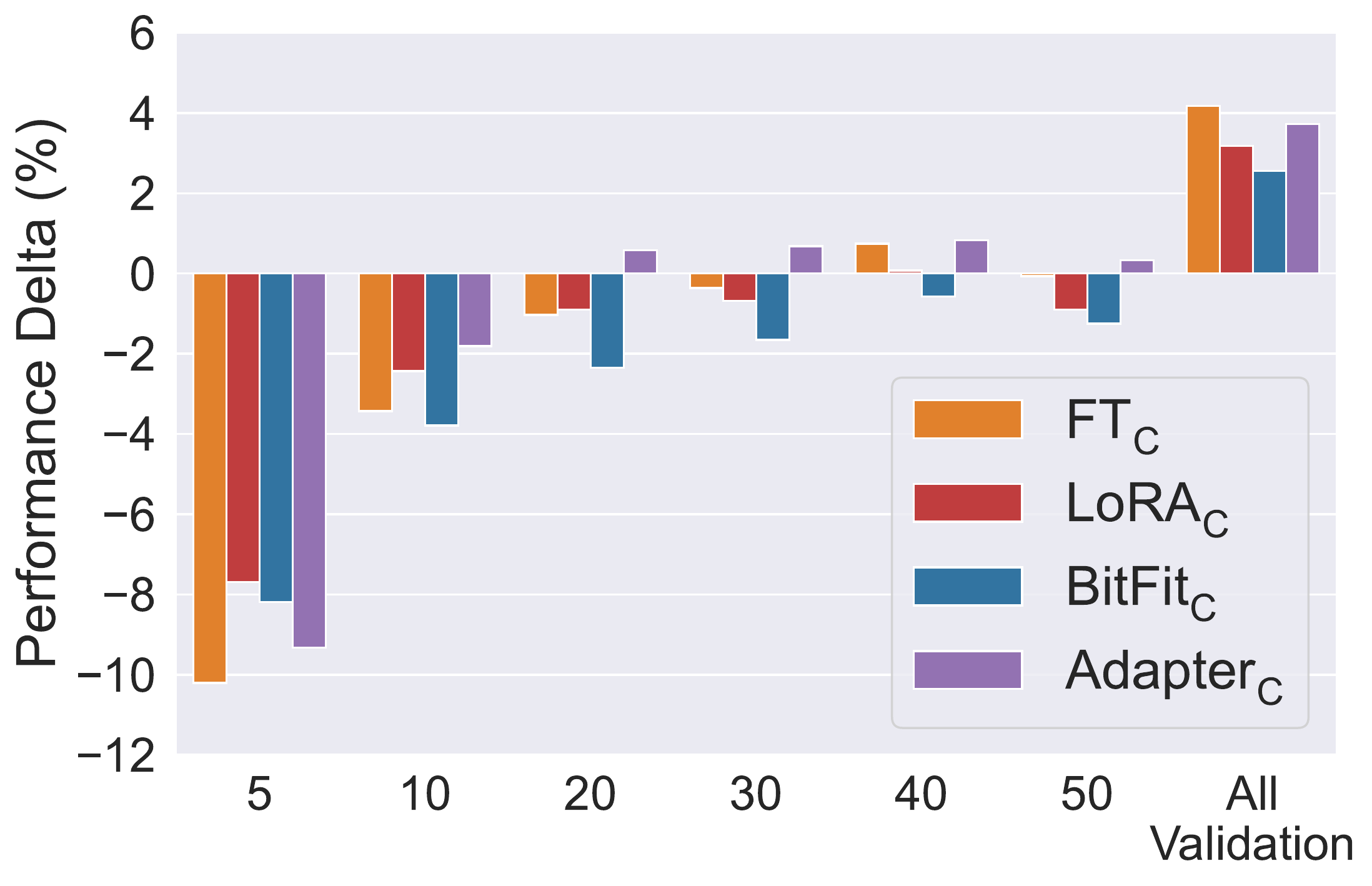}

     \end{subfigure} & %
          \begin{subfigure}[b]{0.45\columnwidth}
         \centering
         \includegraphics[width=\columnwidth]{figs/lcd/lcd_peft_on_small_validation_bars/agnews_compare_with_COSINE.pdf}

     \end{subfigure} \\

        \raisebox{30pt}{\rotatebox[origin=c]{90}{Yelp}}&
     \begin{subfigure}[b]{0.45\columnwidth}
         \centering
         \includegraphics[width=\columnwidth]{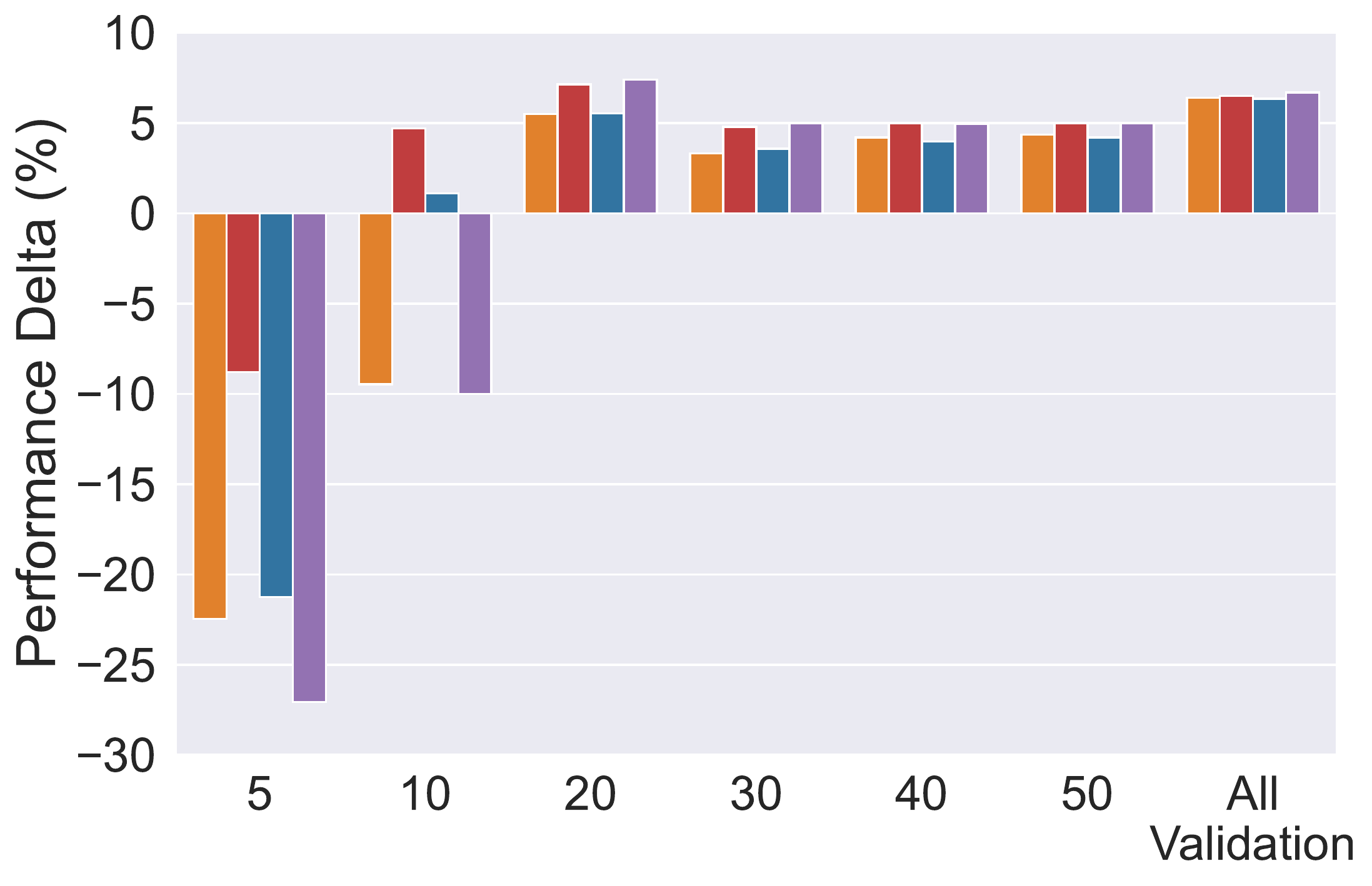}
     \end{subfigure}&%
          \begin{subfigure}[b]{0.45\columnwidth}
         \centering
         \includegraphics[width=\columnwidth]{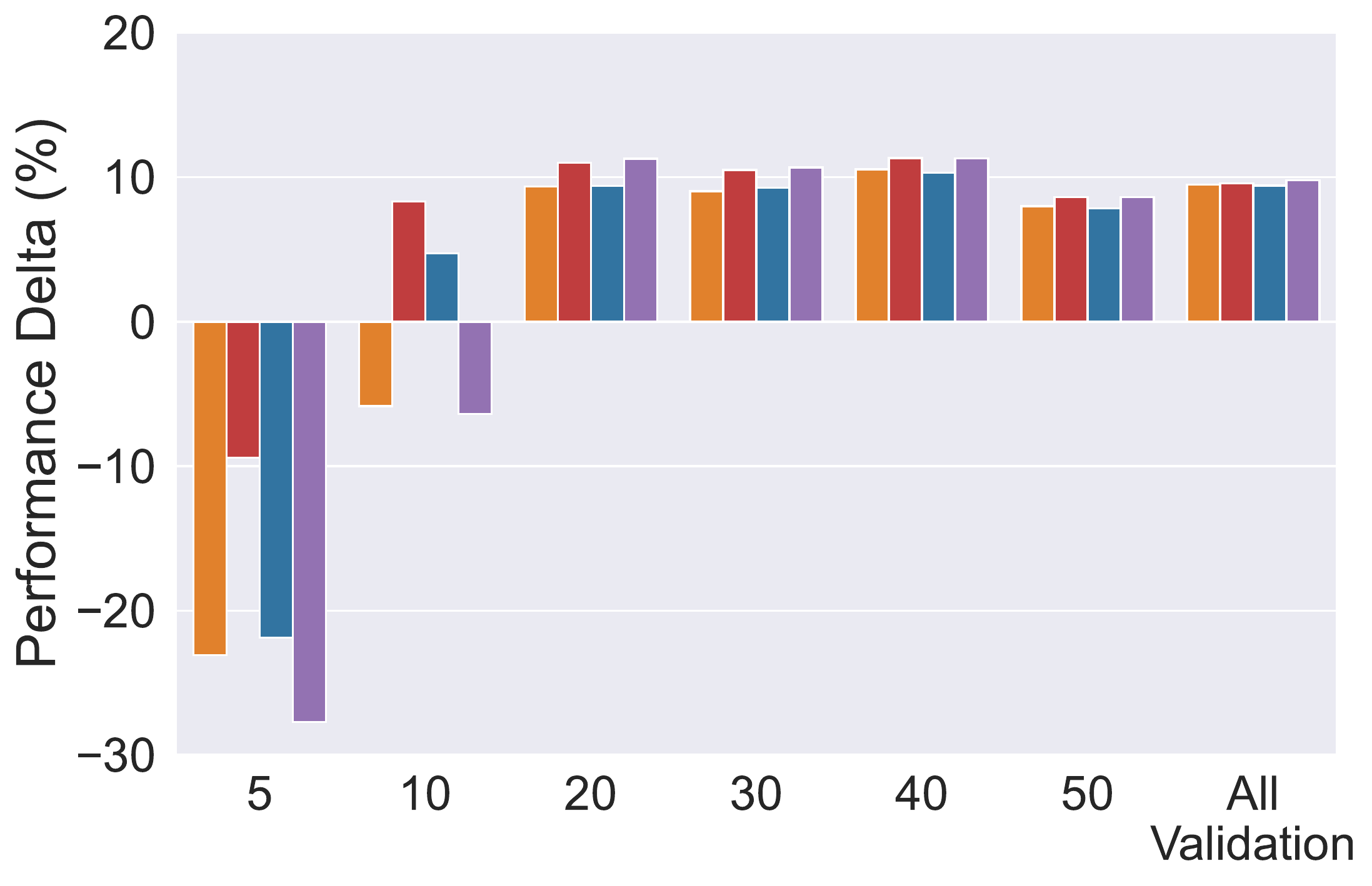}

     \end{subfigure}&%
          \begin{subfigure}[b]{0.45\columnwidth}
         \centering
         \includegraphics[width=\columnwidth]{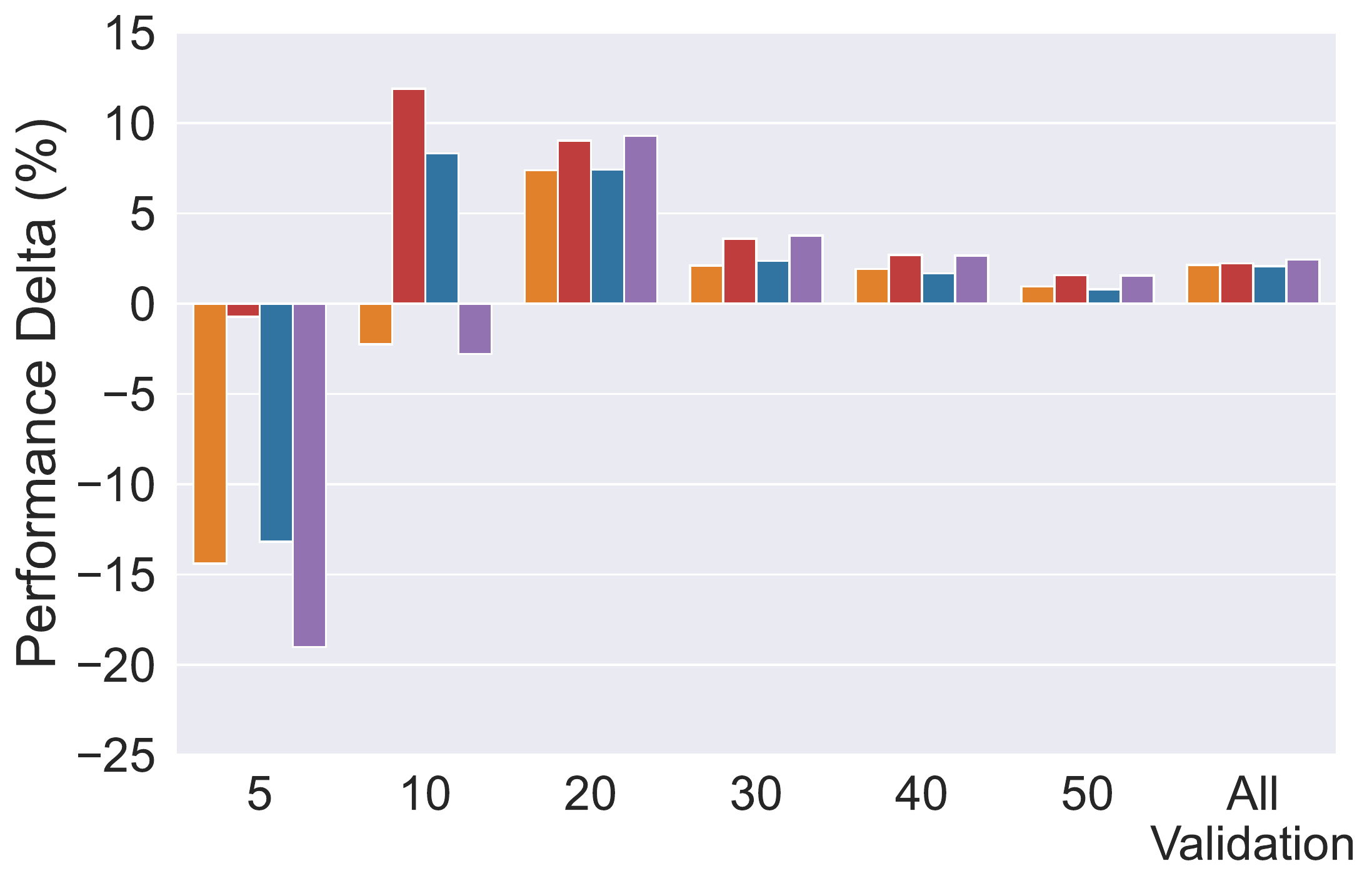}

     \end{subfigure}&%
          \begin{subfigure}[b]{0.45\columnwidth}
         \centering
         \includegraphics[width=\columnwidth]{figs/lcd/lcd_peft_on_small_validation_bars/yelp_compare_with_COSINE.pdf}

     \end{subfigure} \\

     \raisebox{30pt}{\rotatebox[origin=c]{90}{IMDb}}&
     \begin{subfigure}[b]{0.45\columnwidth}
         \centering
         \includegraphics[width=\columnwidth]{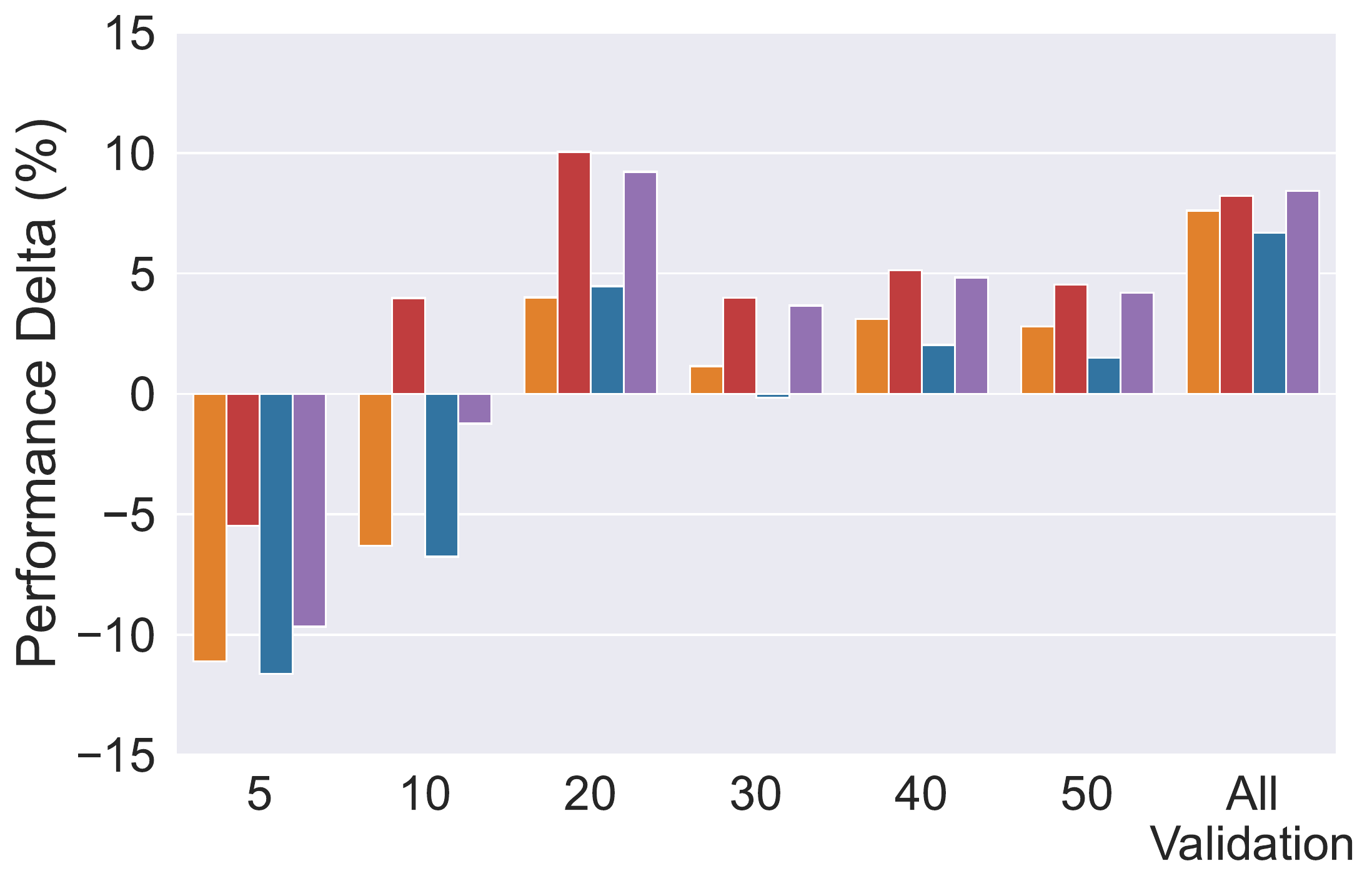}
     \end{subfigure} &%
          \begin{subfigure}[b]{0.45\columnwidth}
         \centering
         \includegraphics[width=\columnwidth]{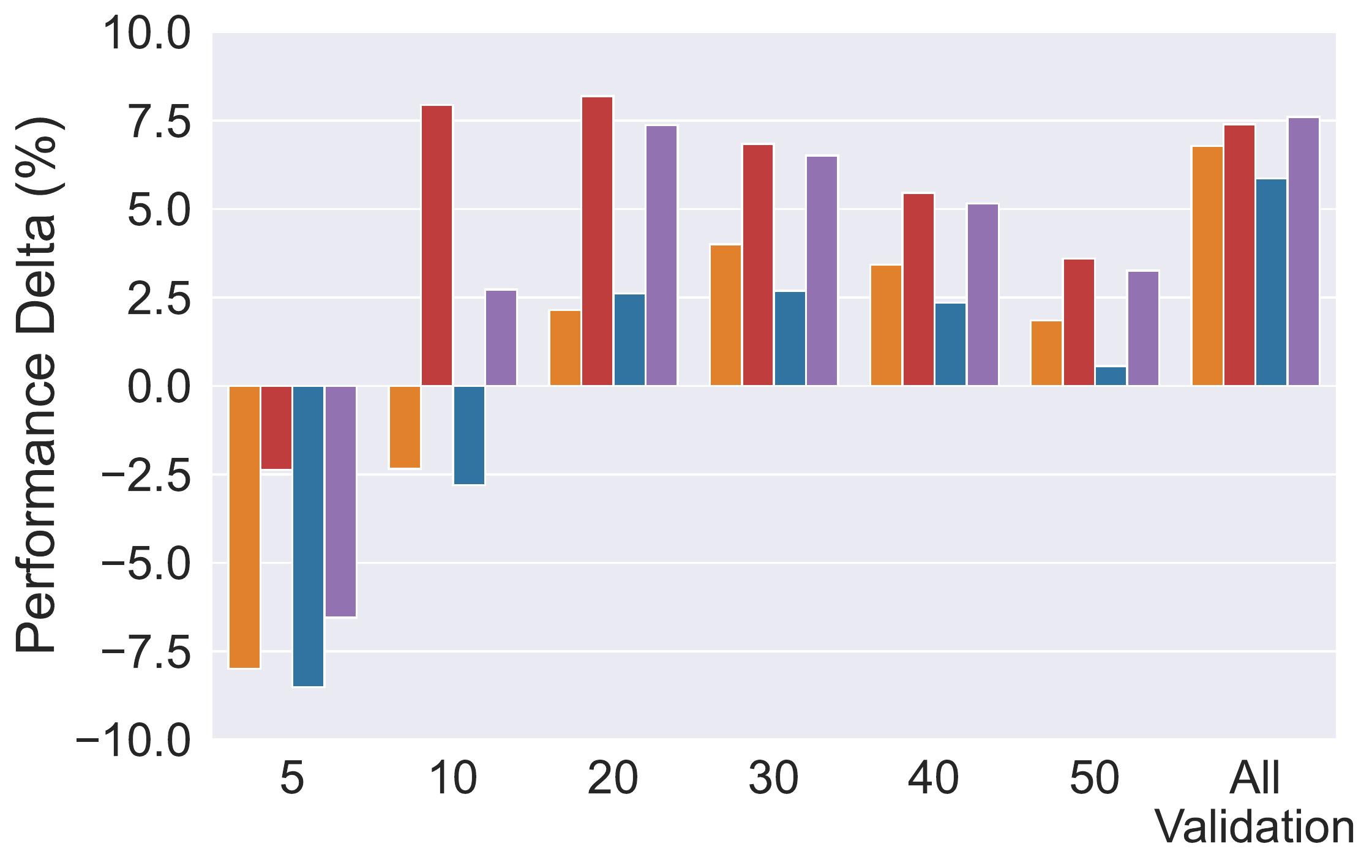}

     \end{subfigure} &%
          \begin{subfigure}[b]{0.45\columnwidth}
         \centering
         \includegraphics[width=\columnwidth]{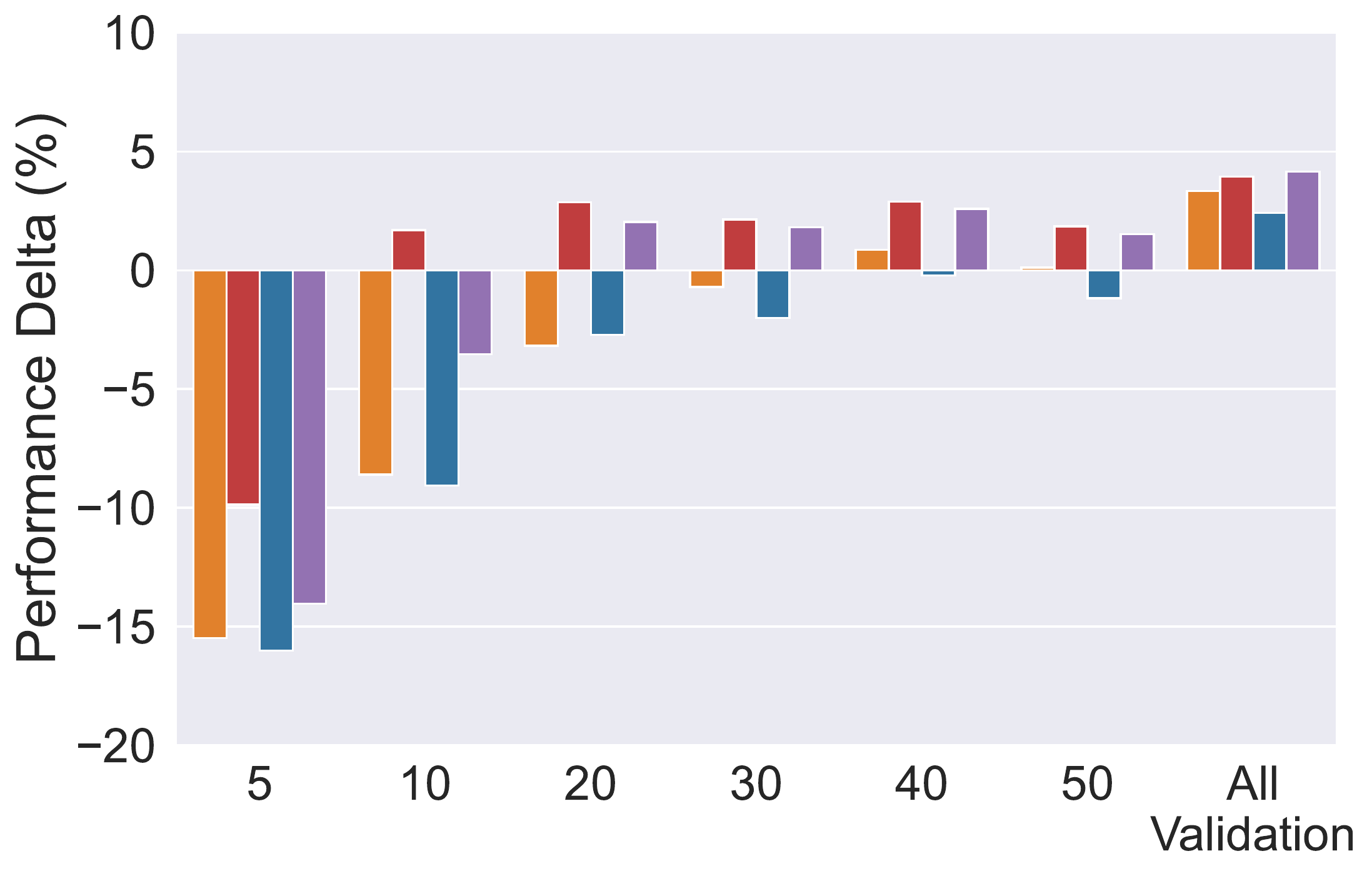}

     \end{subfigure} & %
          \begin{subfigure}[b]{0.45\columnwidth}
         \centering
         \includegraphics[width=\columnwidth]{figs/lcd/lcd_peft_on_small_validation_bars/imdb_compare_with_COSINE.pdf}

     \end{subfigure} \\

         \raisebox{30pt}{\rotatebox[origin=c]{90}{TREC}}&
     \begin{subfigure}[b]{0.45\columnwidth}
         \centering
         \includegraphics[width=\columnwidth]{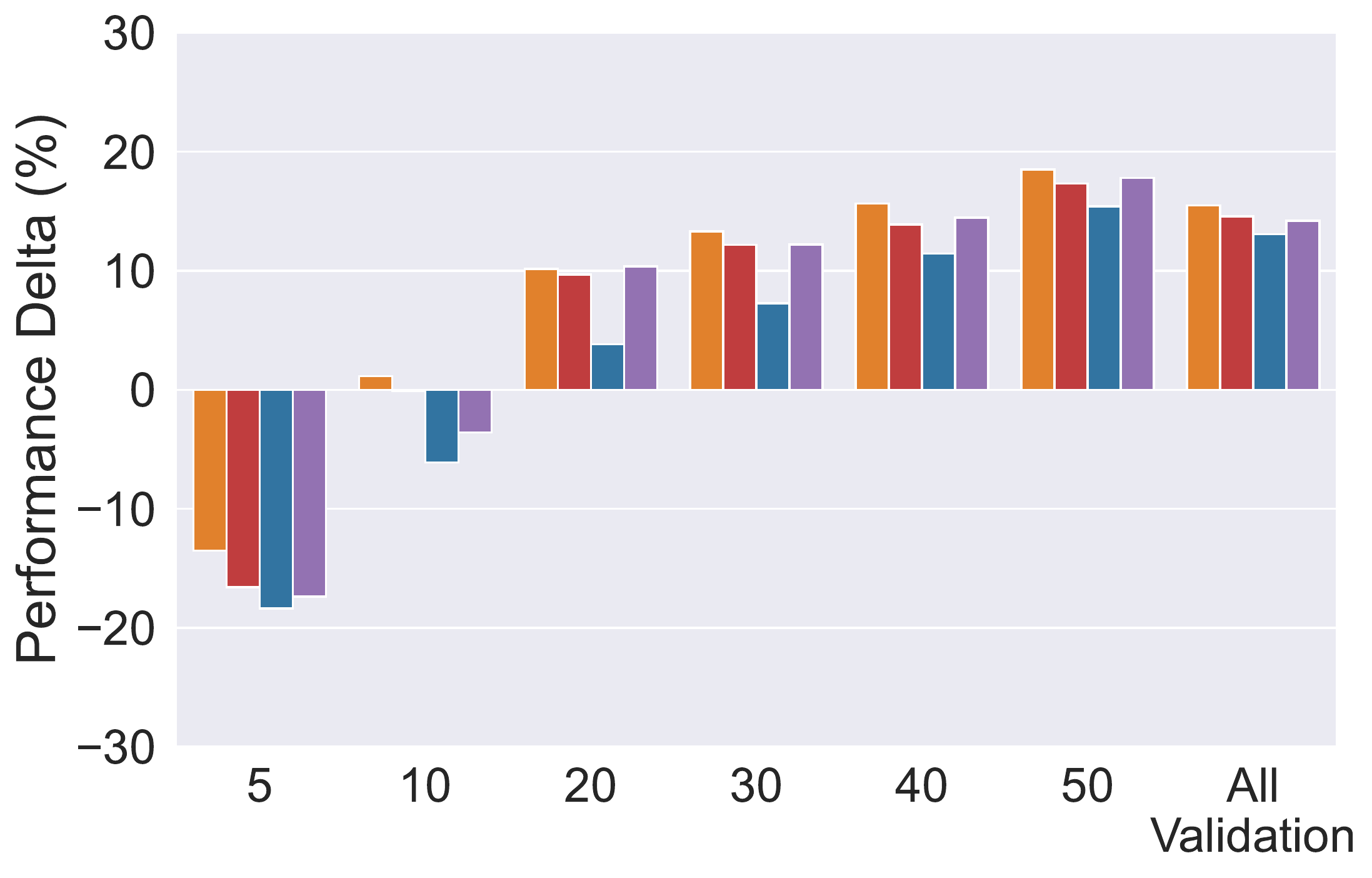}
     \end{subfigure}&%
          \begin{subfigure}[b]{0.45\columnwidth}
         \centering
         \includegraphics[width=\columnwidth]{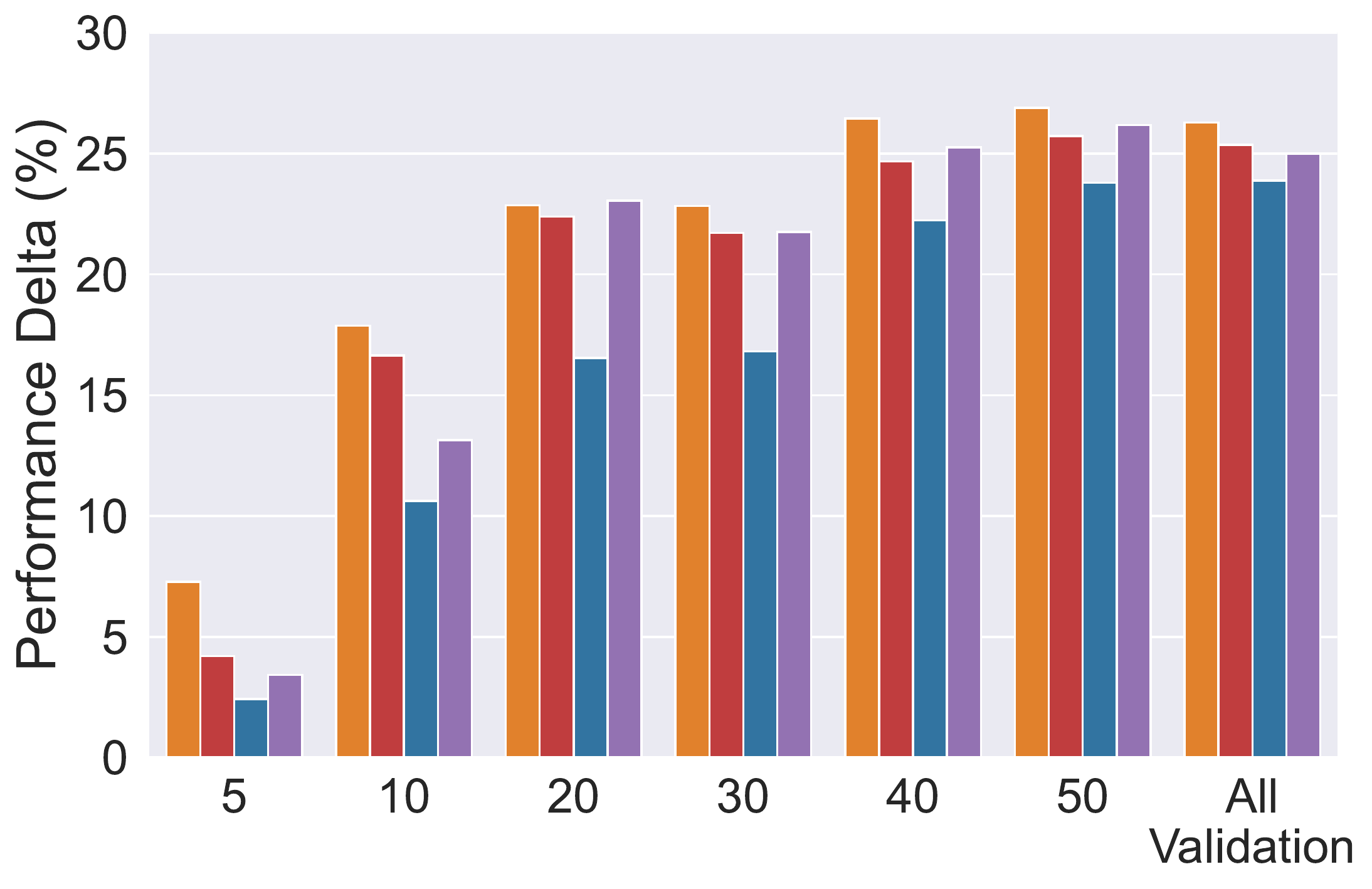}

     \end{subfigure}&%
          \begin{subfigure}[b]{0.45\columnwidth}
         \centering
         \includegraphics[width=\columnwidth]{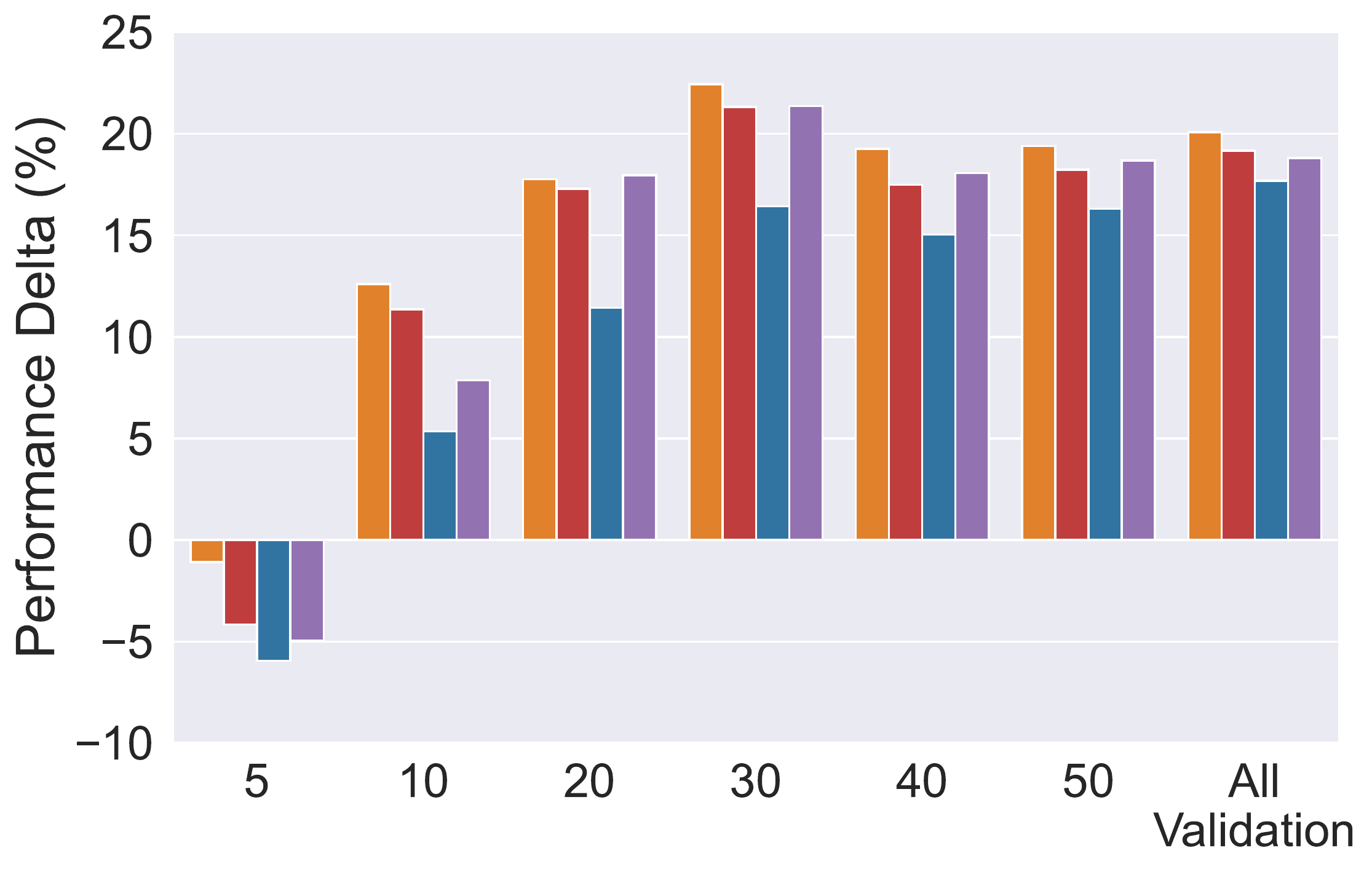}

     \end{subfigure}&%
          \begin{subfigure}[b]{0.45\columnwidth}
         \centering
         \includegraphics[width=\columnwidth]{figs/lcd/lcd_peft_on_small_validation_bars/trec_compare_with_COSINE.pdf}

     \end{subfigure} \\

        \raisebox{30pt}{\rotatebox[origin=c]{90}{SemEval}}&
     \begin{subfigure}[b]{0.45\columnwidth}
         \centering
         \includegraphics[width=\columnwidth]{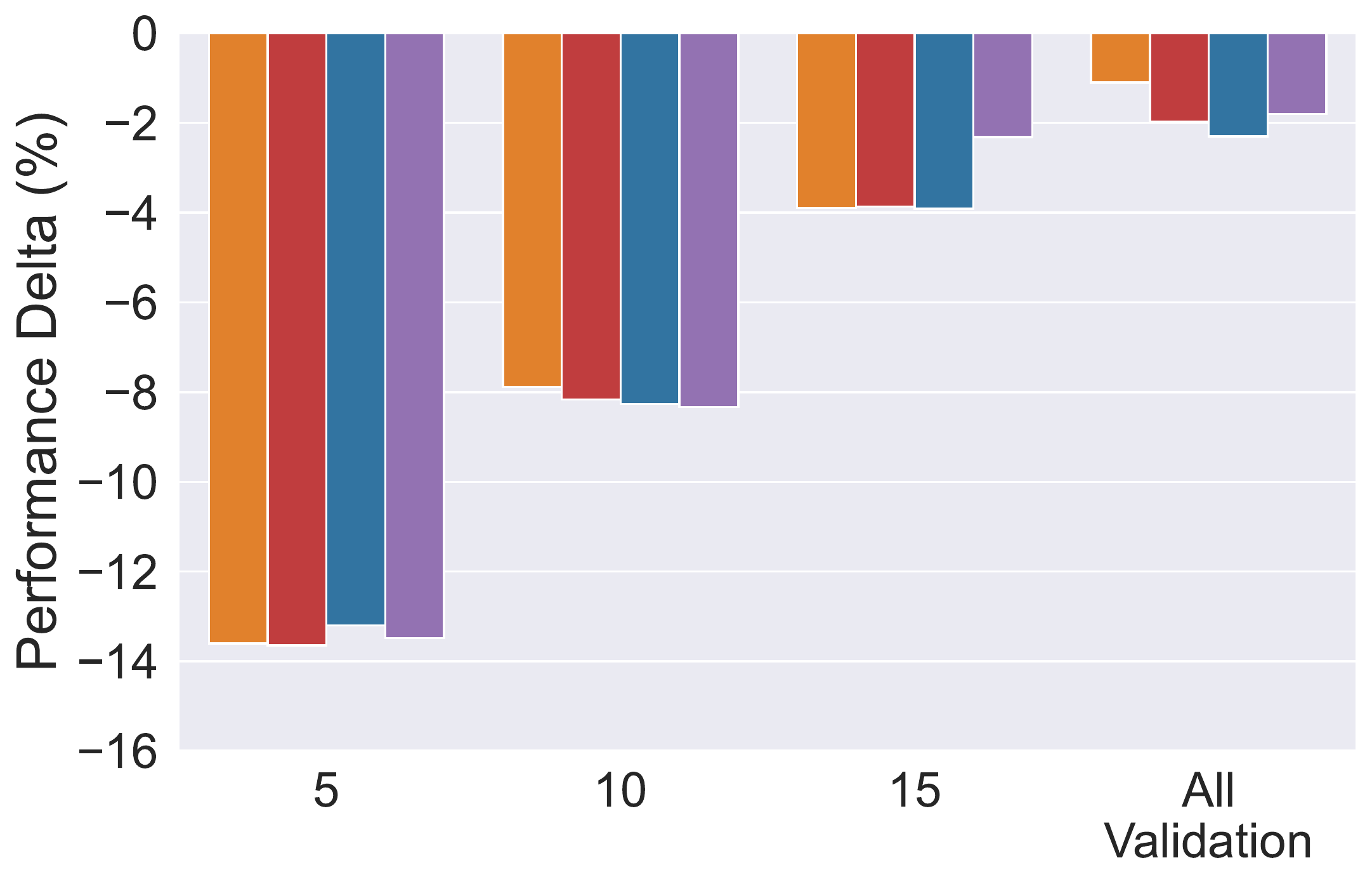}
     \end{subfigure}&%
          \begin{subfigure}[b]{0.45\columnwidth}
         \centering
         \includegraphics[width=\columnwidth]{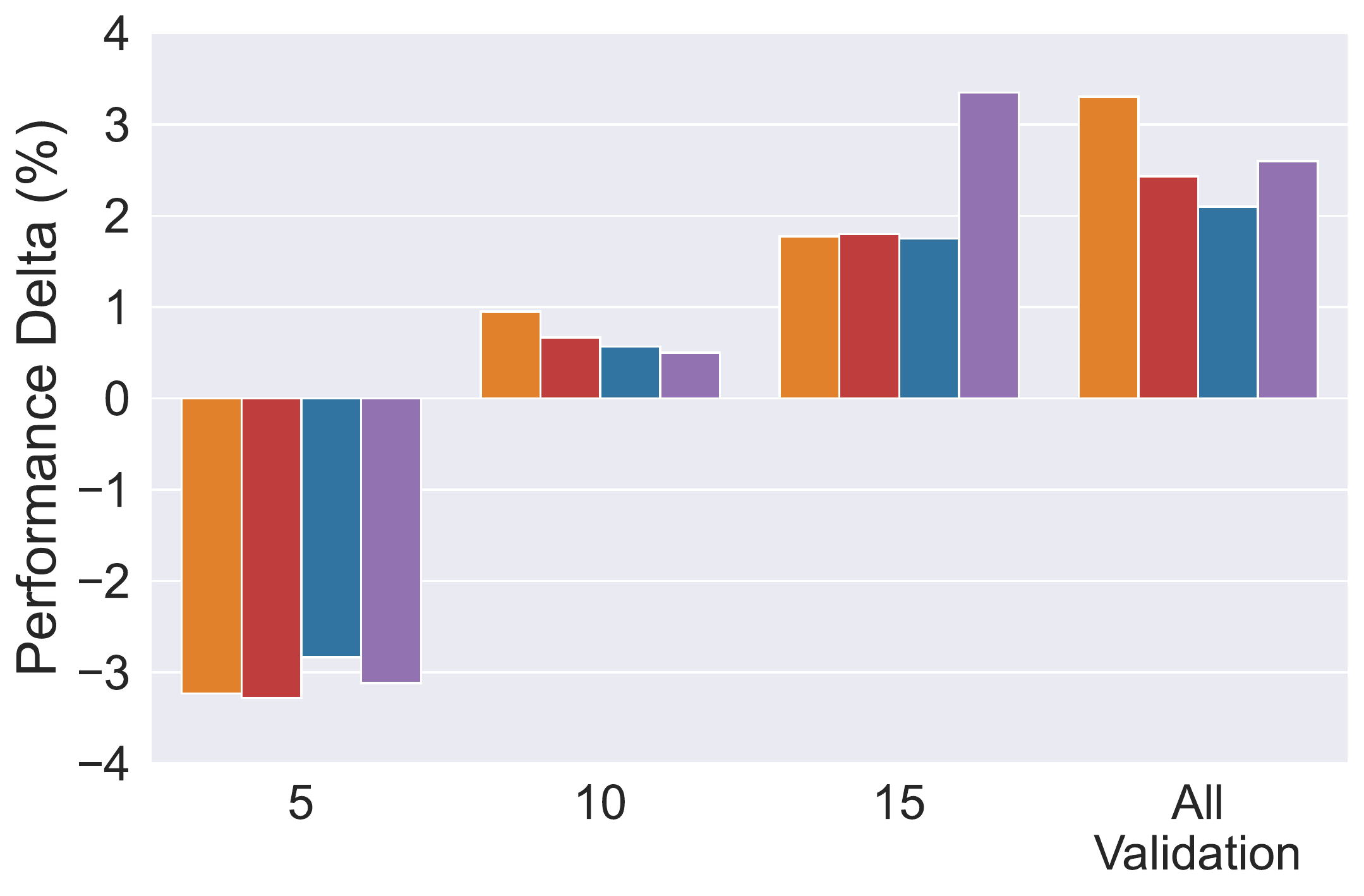}

     \end{subfigure}&%
          \begin{subfigure}[b]{0.45\columnwidth}
         \centering
         \includegraphics[width=\columnwidth]{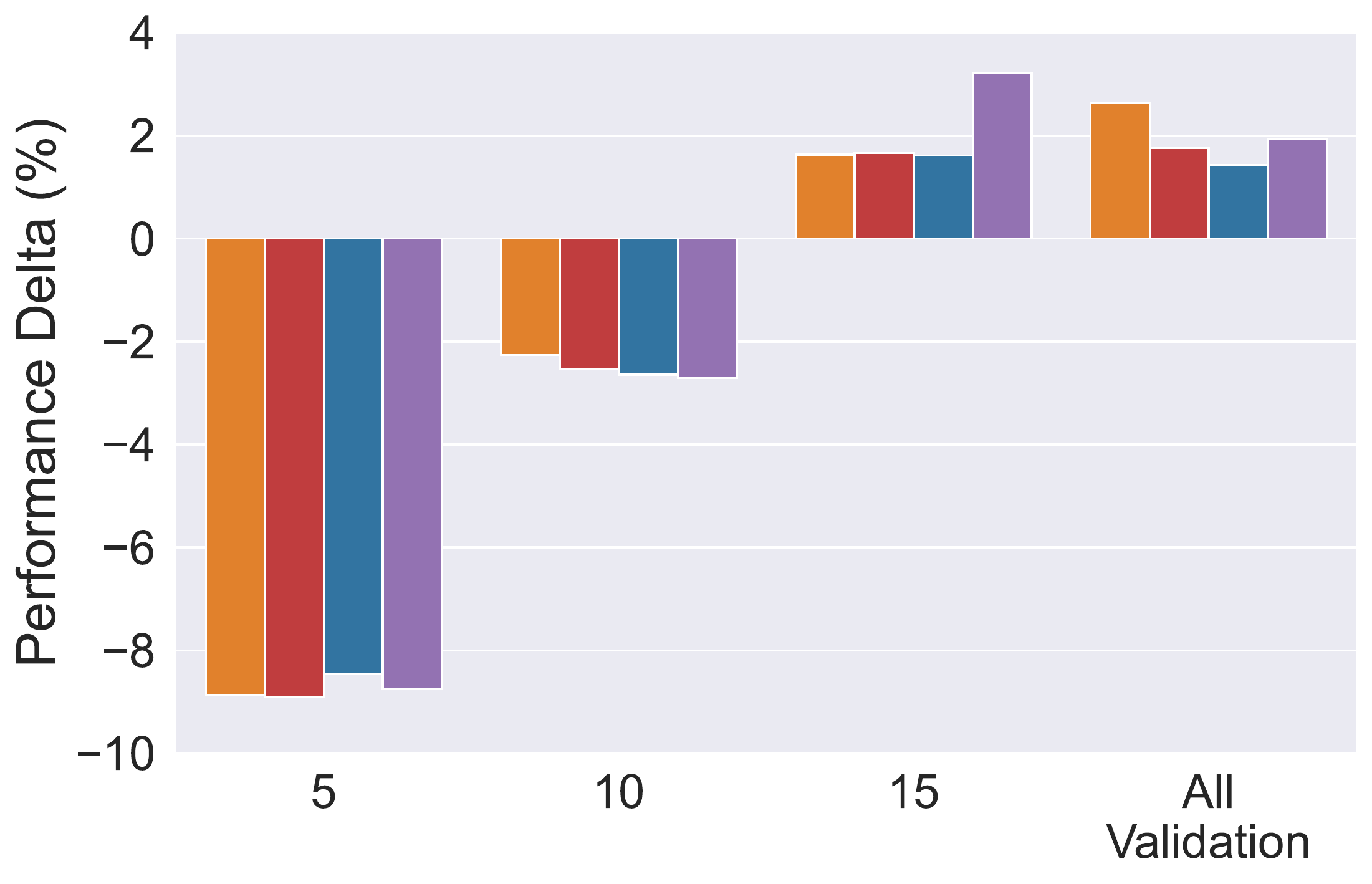}

     \end{subfigure}&%
          \begin{subfigure}[b]{0.45\columnwidth}
         \centering
         \includegraphics[width=\columnwidth]{figs/lcd/lcd_peft_on_small_validation_bars/semeval_compare_with_COSINE.pdf}

     \end{subfigure} \\

        \raisebox{30pt}{\rotatebox[origin=c]{90}{ChemProt}}&
     \begin{subfigure}[b]{0.45\columnwidth}
         \centering
         \includegraphics[width=\columnwidth]{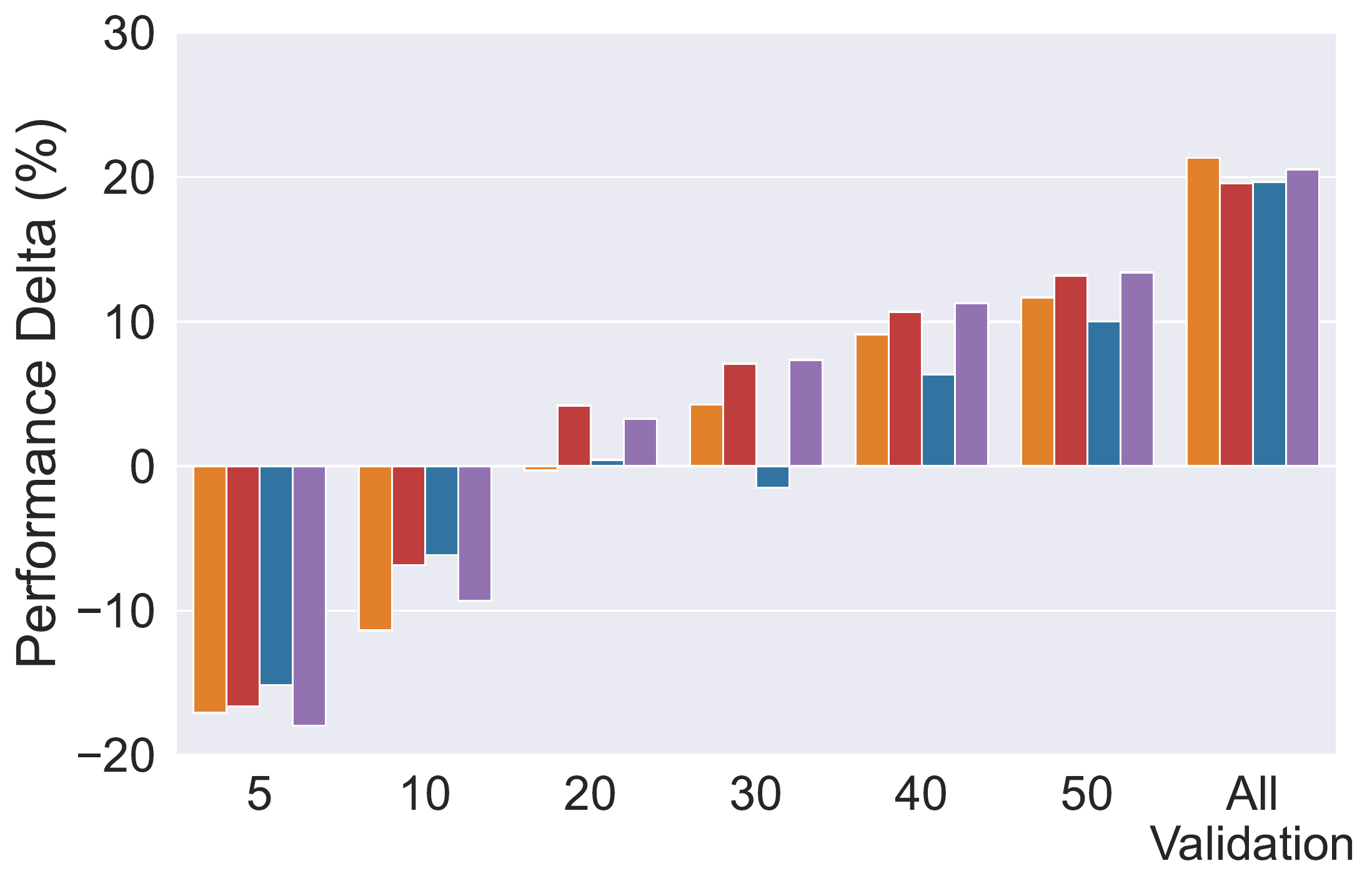}
     \end{subfigure}&%
          \begin{subfigure}[b]{0.45\columnwidth}
         \centering
         \includegraphics[width=\columnwidth]{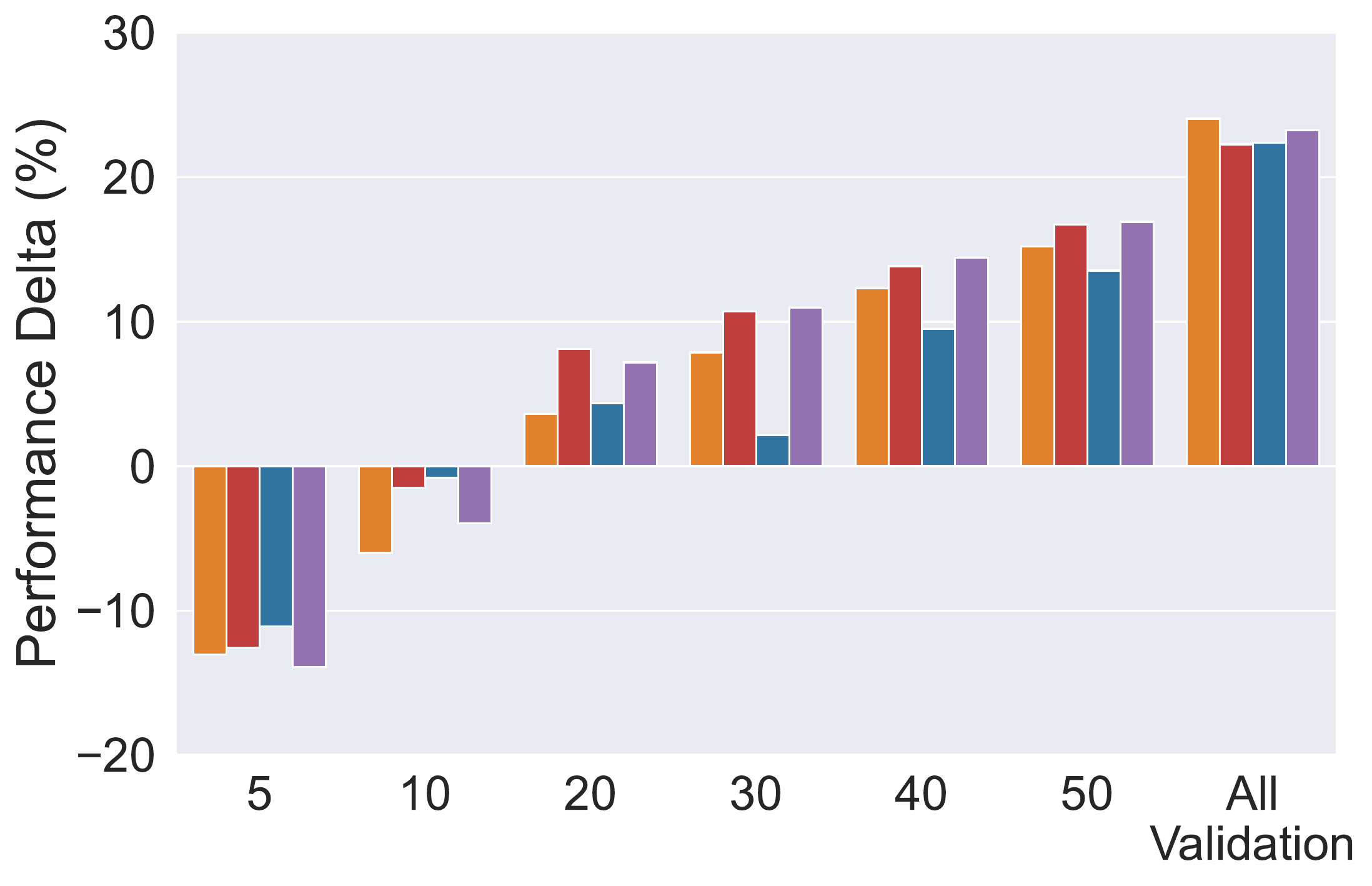}

     \end{subfigure}&%
          \begin{subfigure}[b]{0.45\columnwidth}
         \centering
         \includegraphics[width=\columnwidth]{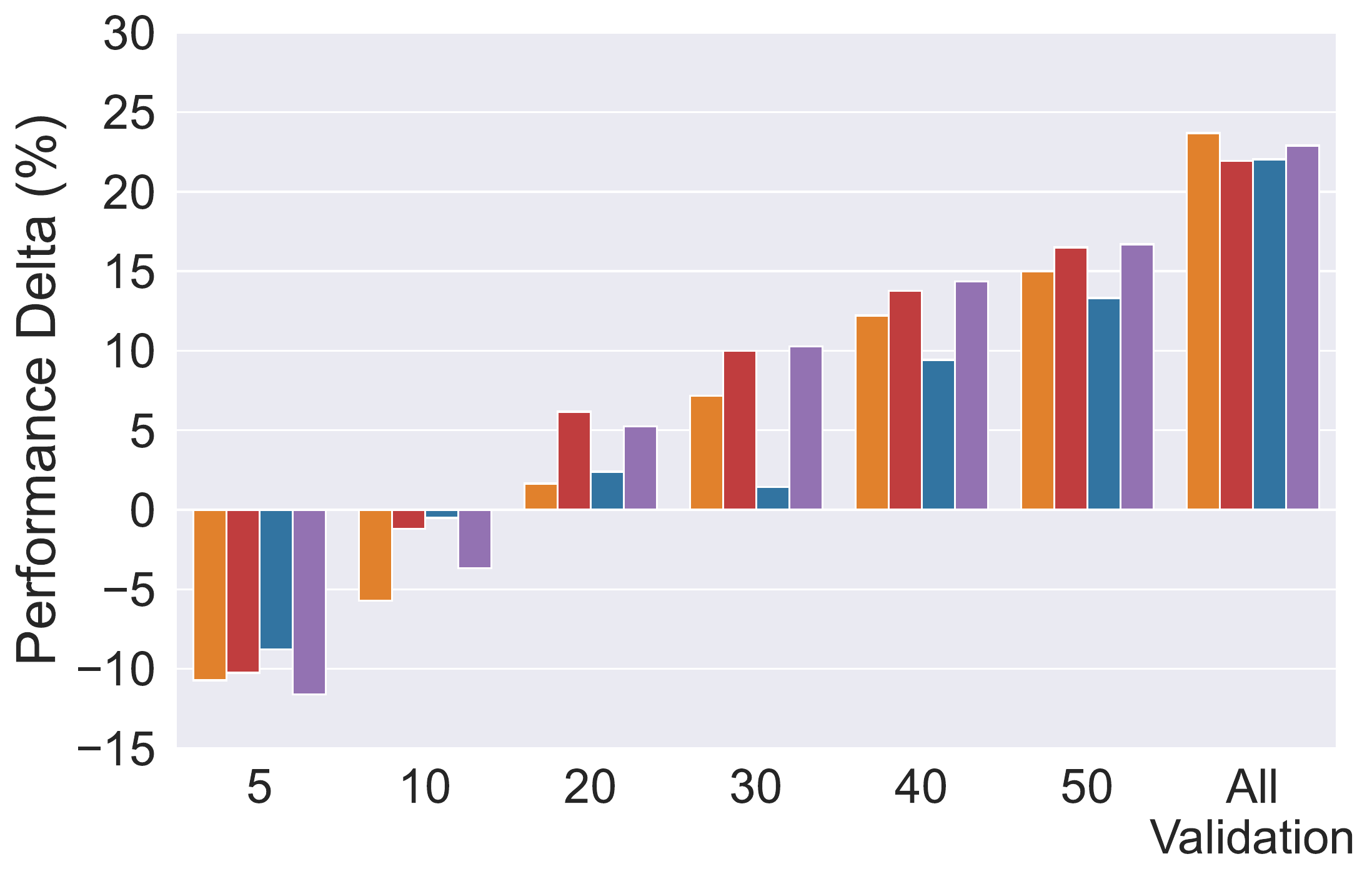}

     \end{subfigure}&%
          \begin{subfigure}[b]{0.45\columnwidth}
         \centering
         \includegraphics[width=\columnwidth]{figs/lcd/lcd_peft_on_small_validation_bars/chemprot_compare_with_COSINE.pdf}

     \end{subfigure} \\

        \raisebox{30pt}{\rotatebox[origin=c]{90}{CoNLL-03}}&
     \begin{subfigure}[b]{0.45\columnwidth}
         \centering
         \includegraphics[width=\columnwidth]{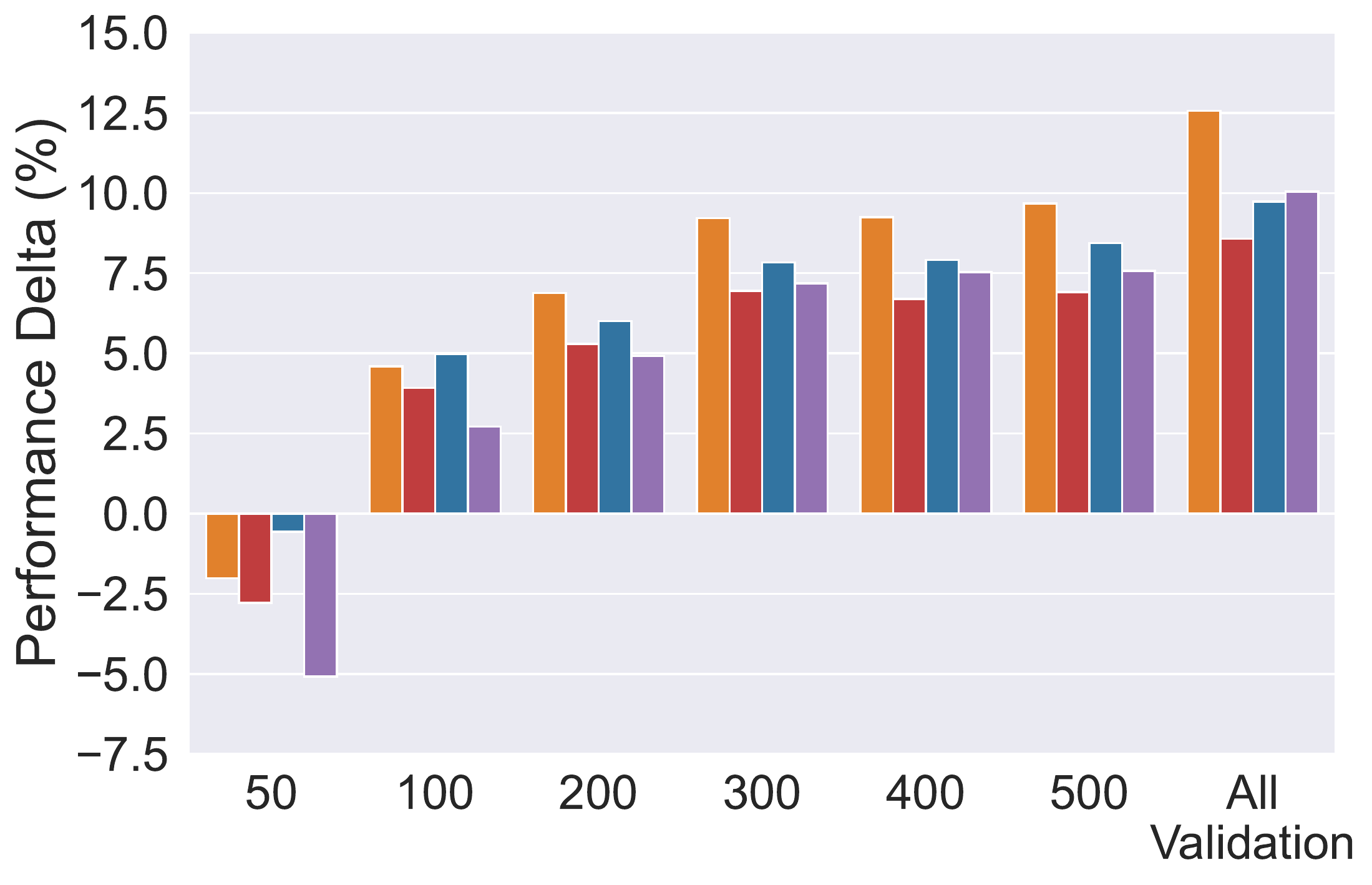}
     \end{subfigure}&%
          \begin{subfigure}[b]{0.45\columnwidth}
         \centering
         \includegraphics[width=\columnwidth]{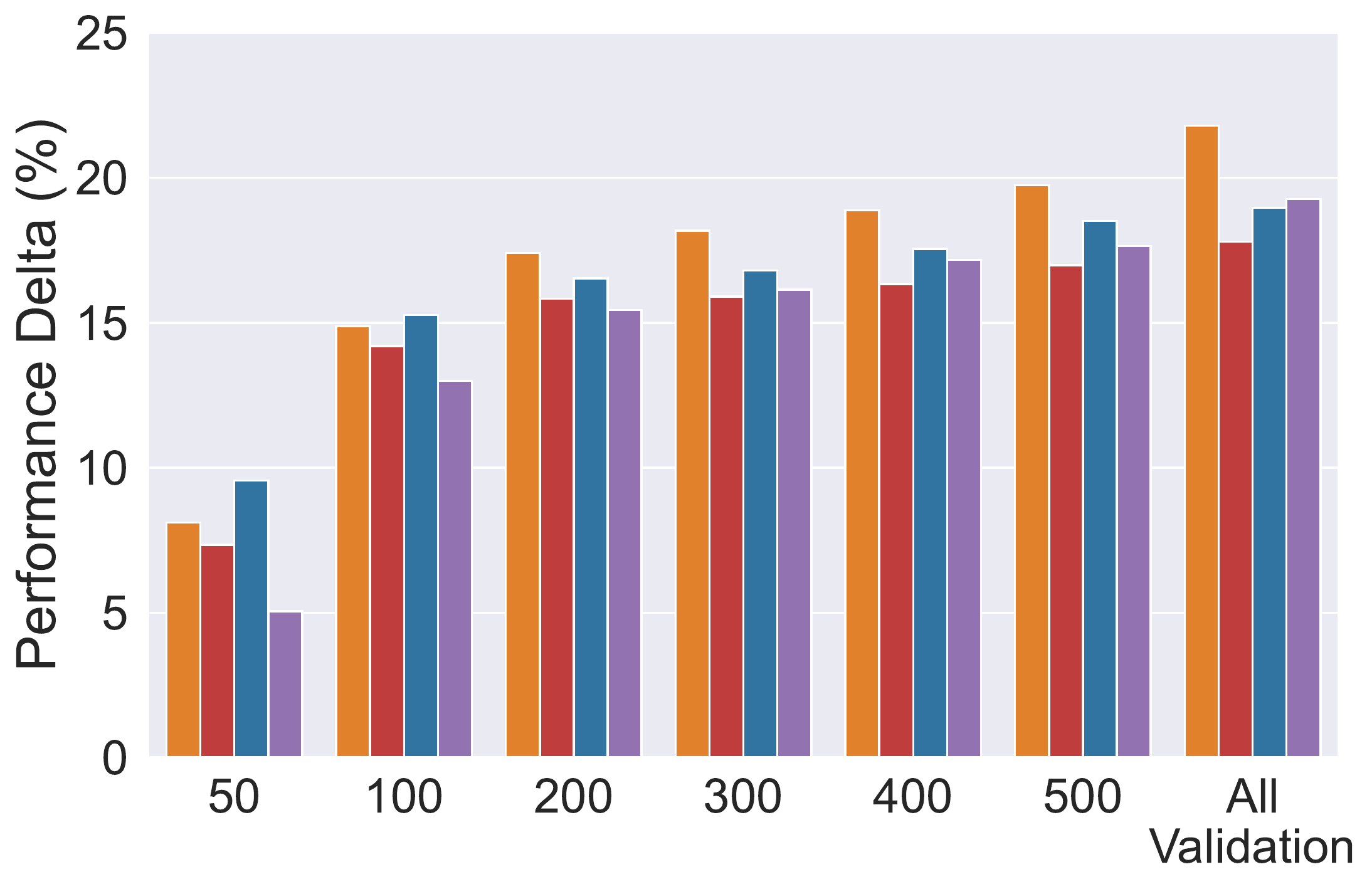}

     \end{subfigure}&%
          \begin{subfigure}[b]{0.45\columnwidth}
         \centering
         \includegraphics[width=\columnwidth]{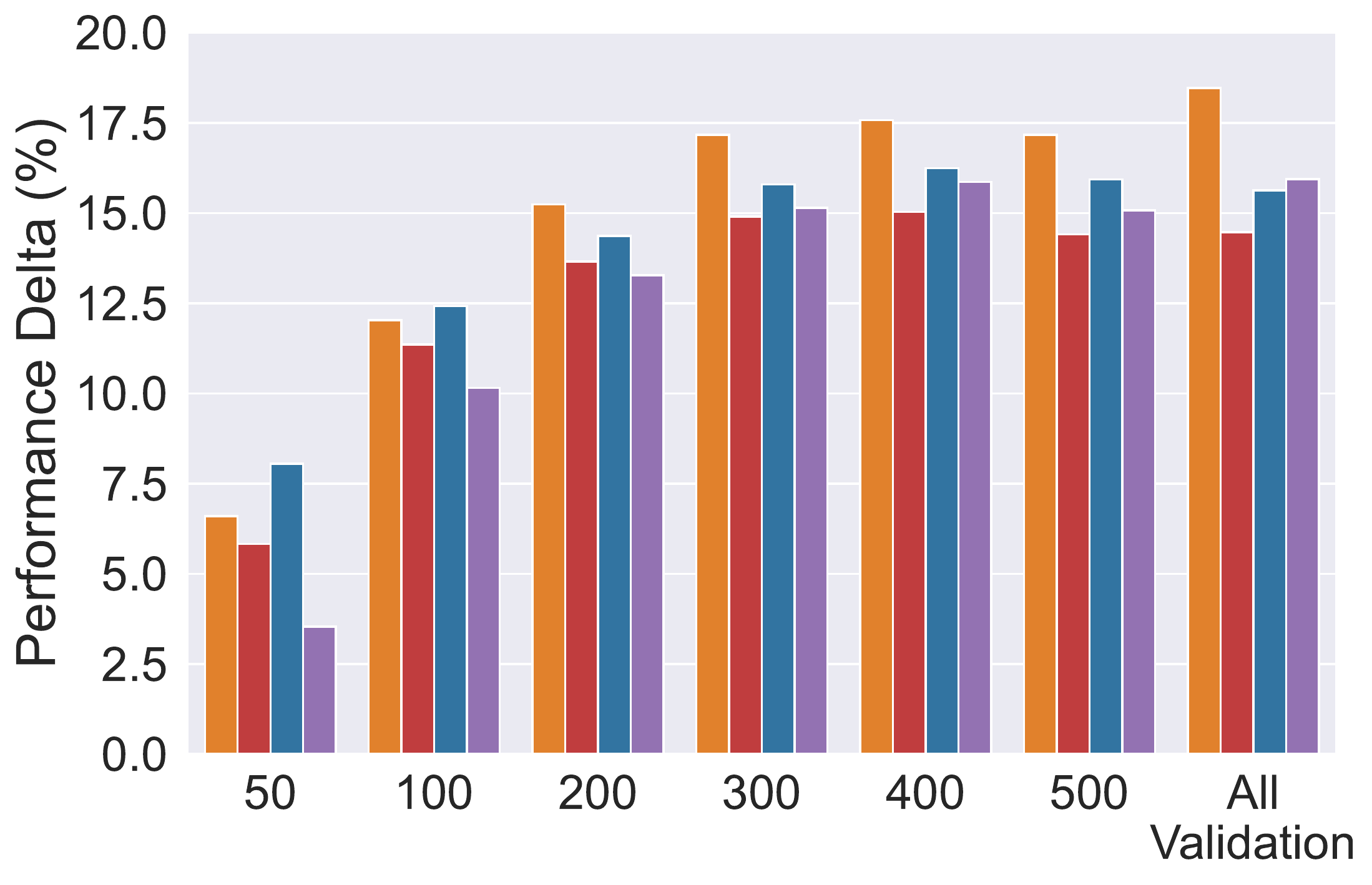}

     \end{subfigure}&%
          \begin{subfigure}[b]{0.45\columnwidth}
         \centering
         \includegraphics[width=\columnwidth]{figs/lcd/lcd_peft_on_small_validation_bars/conll_compare_with_COSINE.pdf}

     \end{subfigure} \\

        \raisebox{30pt}{\rotatebox[origin=c]{90}{OntoNotes 5.0}}&
     \begin{subfigure}[b]{0.45\columnwidth}
         \centering
         \includegraphics[width=\columnwidth]{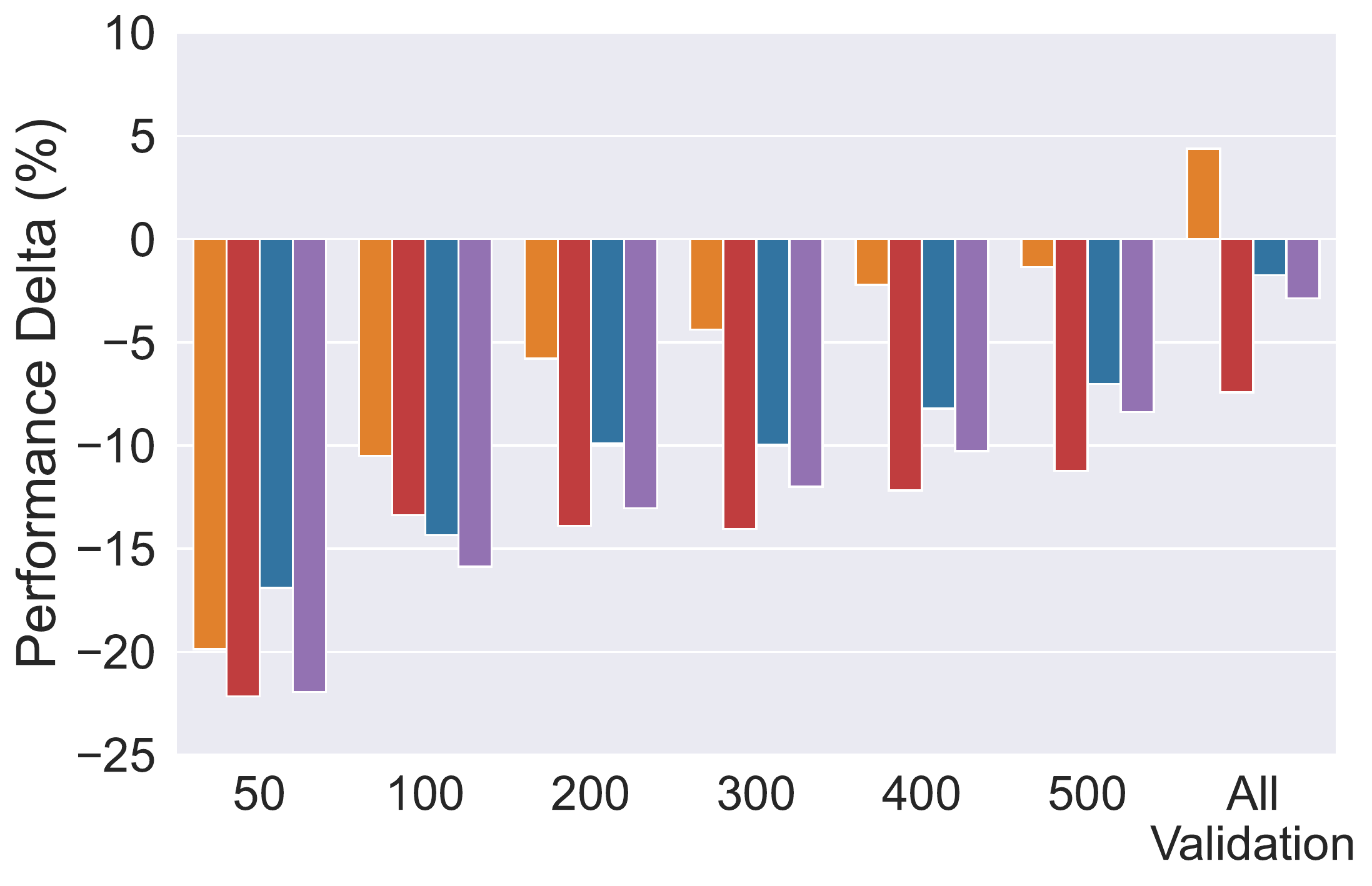}
     \end{subfigure}&%
          \begin{subfigure}[b]{0.45\columnwidth}
         \centering
         \includegraphics[width=\columnwidth]{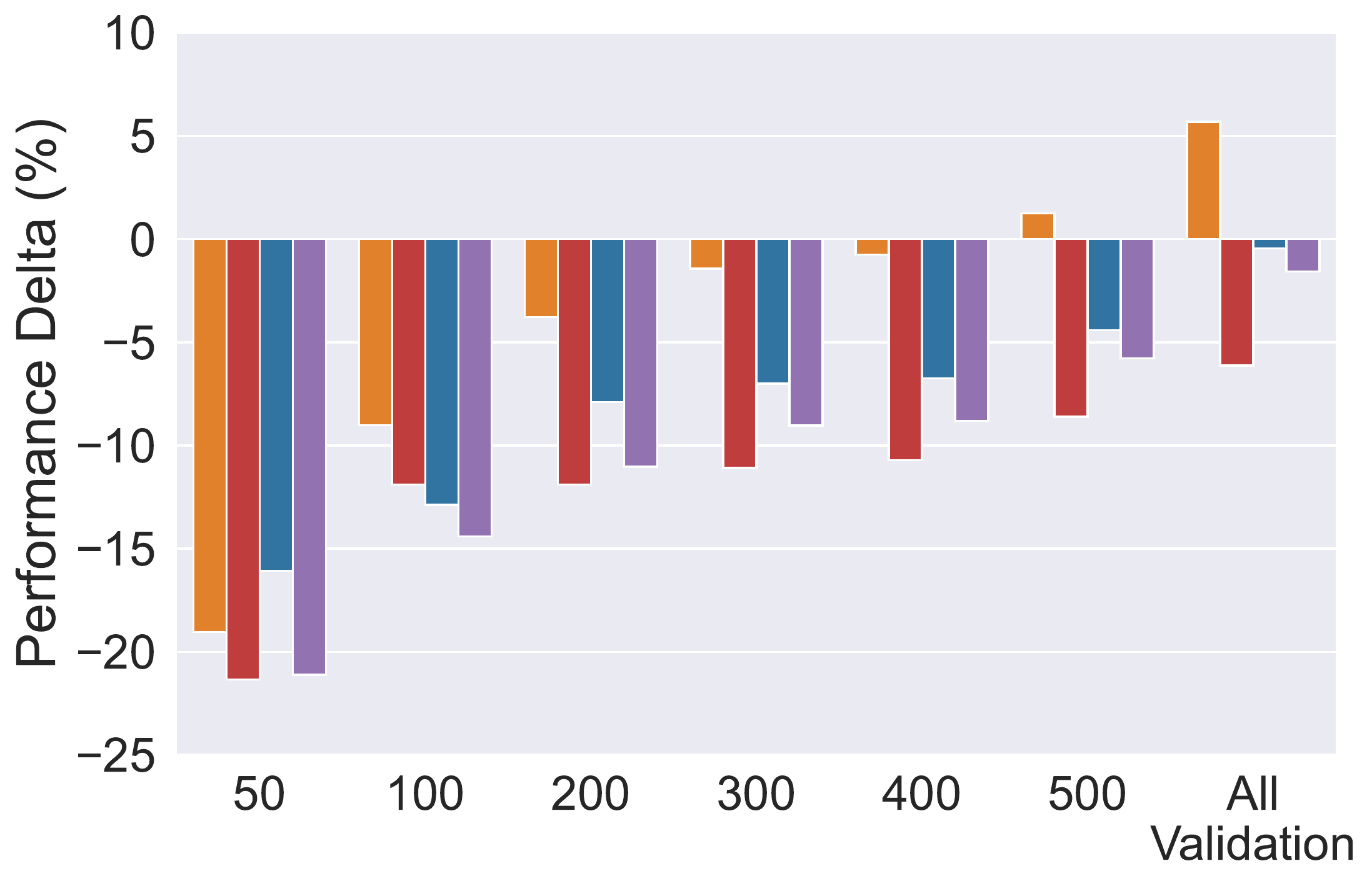}

     \end{subfigure}&%
          \begin{subfigure}[b]{0.45\columnwidth}
         \centering
         \includegraphics[width=\columnwidth]{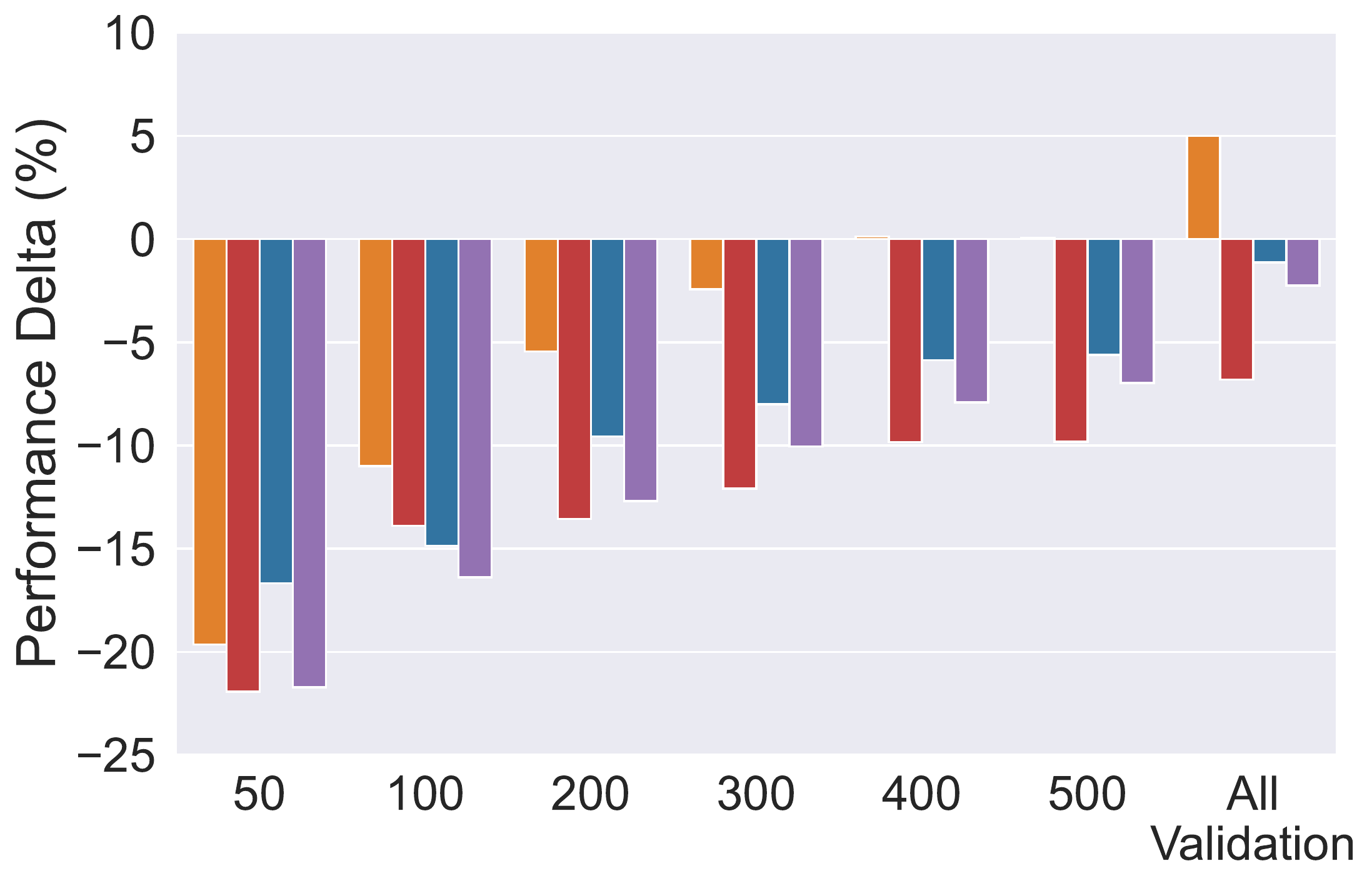}

     \end{subfigure}&%
          \begin{subfigure}[b]{0.45\columnwidth}
         \centering
         \includegraphics[width=\columnwidth]{figs/lcd/lcd_peft_on_small_validation_bars/ontonotes_compare_with_COSINE.pdf}

     \end{subfigure} \\
     \end{tabular}
        \caption{%
        Performance difference of (parameter-efficient) fine-tuning approaches (FT, LoRA, BitFit and Adapter) with WSL apporaches (L2R, MLC, BOND and COSINE), using varying amounts of clean data. We use the subscript ``C" (e.g., FT\textsubscript{C}) to indicate that the fine-tuning approaches are applied on clean data.}
        \label{fig:peft_vs_weak_full}
\end{figure*}

\section{Training with clean samples}
\label{sec:appendix:ft_on_clean_samples}
\subsection{Methods and implementation details}
\label{sec:appendix:methods_and_implementation_detail}
In Section \ref{sec:little_clean_data_for_peft}, we apply four (parameter-efficient) fine-tuning approaches to train models on clean validation sets. Since we do not have extra data for model selection, we choose a fixed set of hyperparameters for all datasets. In the following we briefly introduce the fine-tuning approaches, together with their hyperparameter configurations.
\begin{itemize}
  \item Vanilla fine-tuning \cite{devlin2019bert, liu2019roberta} is the standard fine-tuning approaches for pre-trained language models. It works by adding a randomly initialized classifier on top of the pre-trained model and training it together with all other model parameters. We use a fixed learning rate of $2e^{-5}$ in all experiments.
  \item Adapter-based fine-tuning \cite{houlsby2019parameter} adds additional feed-forward layers called adapters to each layer of the pre-trained language model. During fine-tuning, we only update the weights of these adapter layers and keep all other parameters \textit{frozen} at their pre-trained values. We use a fixed learning rate of $2e^{-5}$ in all experiments. The reduction factor is set to 16.
  \item BitFit \cite{zaken2022bitfit} updates only the bias parameters of every layer and keeps all other weights frozen. Despite its simplicity it has been demonstrated to achieve similar results to adapter-based fine-tuning. We use a fixed learning rate of $1e^{-4}$ in all experiments.
  \item LoRA \cite{hu2021lora} is a recently proposed adapter-based fine-tuning method which uses a low-rank bottleneck architecture in each of the newly added feed-forward networks. The motivation here is to perform a low rank update to the model during fine-tuning. We use a fixed learning rate of $2e^{-5}$ in all experiments. The $\alpha$ value used in LoRa is fixed to 16.
\end{itemize}
In all experiments, the batch size used in all fine-tuning approaches is 32. The optimizer is AdamW \cite{loshchilov2017decoupled}.

\subsection{Training on the full validation sets}
\label{sec:appendix:training_on_the_full_validation_sets}
In addition to training sets, the WRENCH \cite{zhang2021_wrench} benchmark provides a validation set for each of its tasks. The validation sets are cleanly annotated and typically range in size from 5\% to 25\% of the weakly annotated training sets. Although such validation size is reasonable for fully supervised learning, we suspect that it is exorbitant in the sense that it provides a significantly better training signal for models than the weakly annotated training set. Thus we compare the performance of recent WSL approaches that access both the training and validation sets with a model that is directly fine-tuned on the validation set. The following WSL methods are included in this experiment: L2R \cite{ren2018learning}, MetaWN \cite{shu2019meta}, BOND \cite{Liang2020_bond}, Denoise \cite{ren2020_denoising}, MLC \cite{zheng2021meta}, and COSINE \cite{Yu2021_cosine}. Following prior work, we select the best set of hyperparameters via the validation set when applying the WSL methods. Also, early-stopping based on the validation performance is applied. In contrast, the direct fine-tuning baseline uses a fixed set of hyperparameters across all datasets, and no early-stopping is applied (same configuration as in Appendix \ref{sec:appendix:methods_and_implementation_detail}). We train this baseline for 6000 steps. In all cases, the training losses converged much earlier than 6000 steps, but we deliberately kept training for longer to show that the good performance achieved by this baseline is not due to any fine-grained configurations. As shown in Figure \ref{fig:train_on_full_clean_validation}, this simple baseline outperforms all the WSL methods in all but one case.

\subsection{Extended comparison of training on clean data and validation for WSL approaches}

In Section \ref{sec:little_clean_data_for_peft}, standard fine-tuning (FT) and multiple parameter-efficient fine-tuning (PEFT) are compared with the competitive WSL method COSINE.
In this section, we provide additional plots which show the same comparison with the other WSL methods examined in this work, namely L2R, MLC, and BOND.
We report average performance (Acc. and F1 in \%) difference between (parameter-efficient) fine-tuning methods and the specific WSL method for varying number of clean samples. 
The overall tendency is consistent with the results in Section \ref{sec:little_clean_data_for_peft}: WSL methods perform well on a small amount of clean labeled data but PEFT outperforms WSL methods with an increasing amount of clean labeled data.

\begin{figure*}[t]
        \centering
        
     \begin{subfigure}[b]{0.5\columnwidth}
         \centering
         \includegraphics[width=\columnwidth]{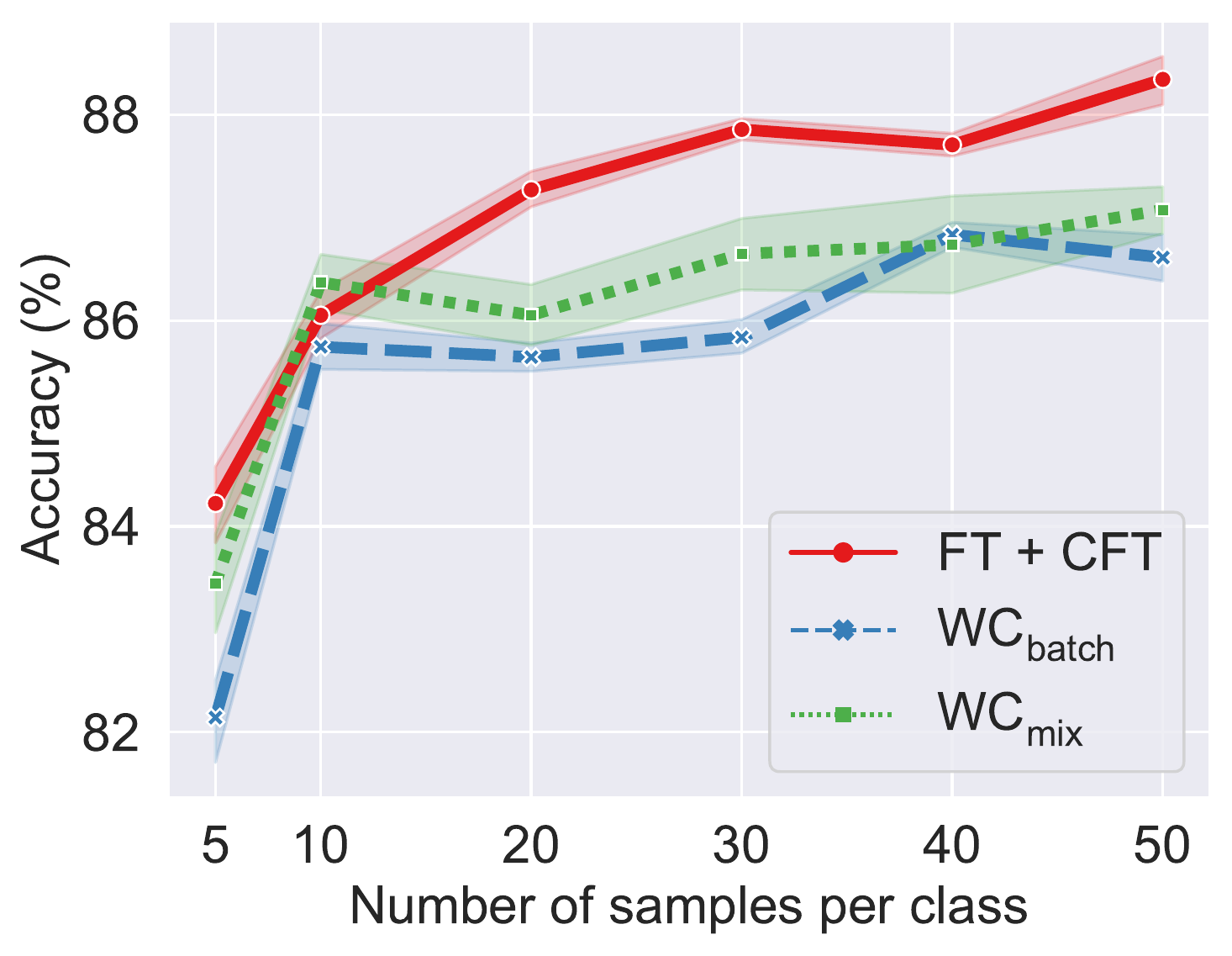}
         \caption{AGNews}
     \end{subfigure}\hfill
          \begin{subfigure}[b]{0.5\columnwidth}
         \centering
         \includegraphics[width=\columnwidth]{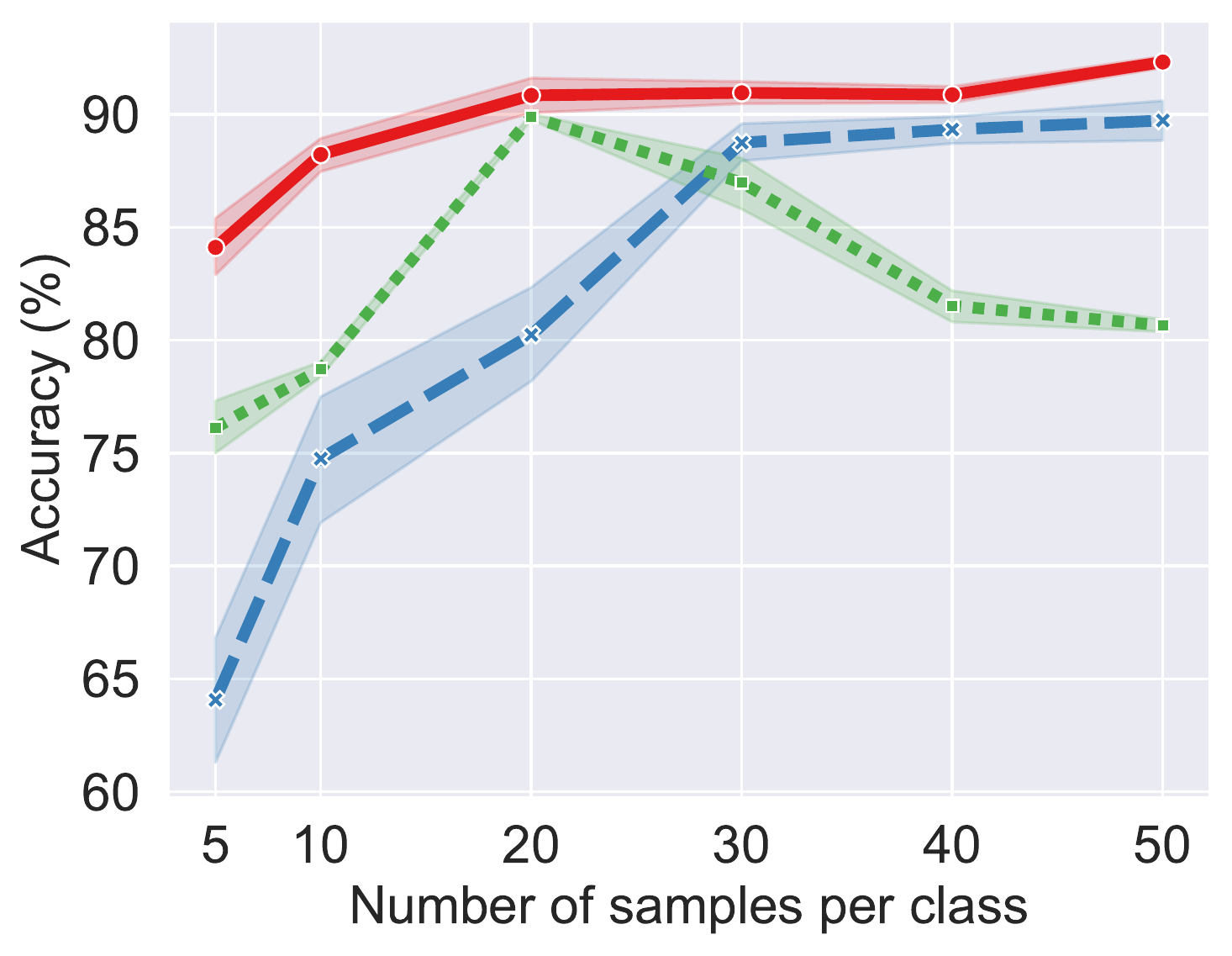}
         \caption{Yelp}

     \end{subfigure}\hfill
          \begin{subfigure}[b]{0.5\columnwidth}
         \centering
         \includegraphics[width=\columnwidth]{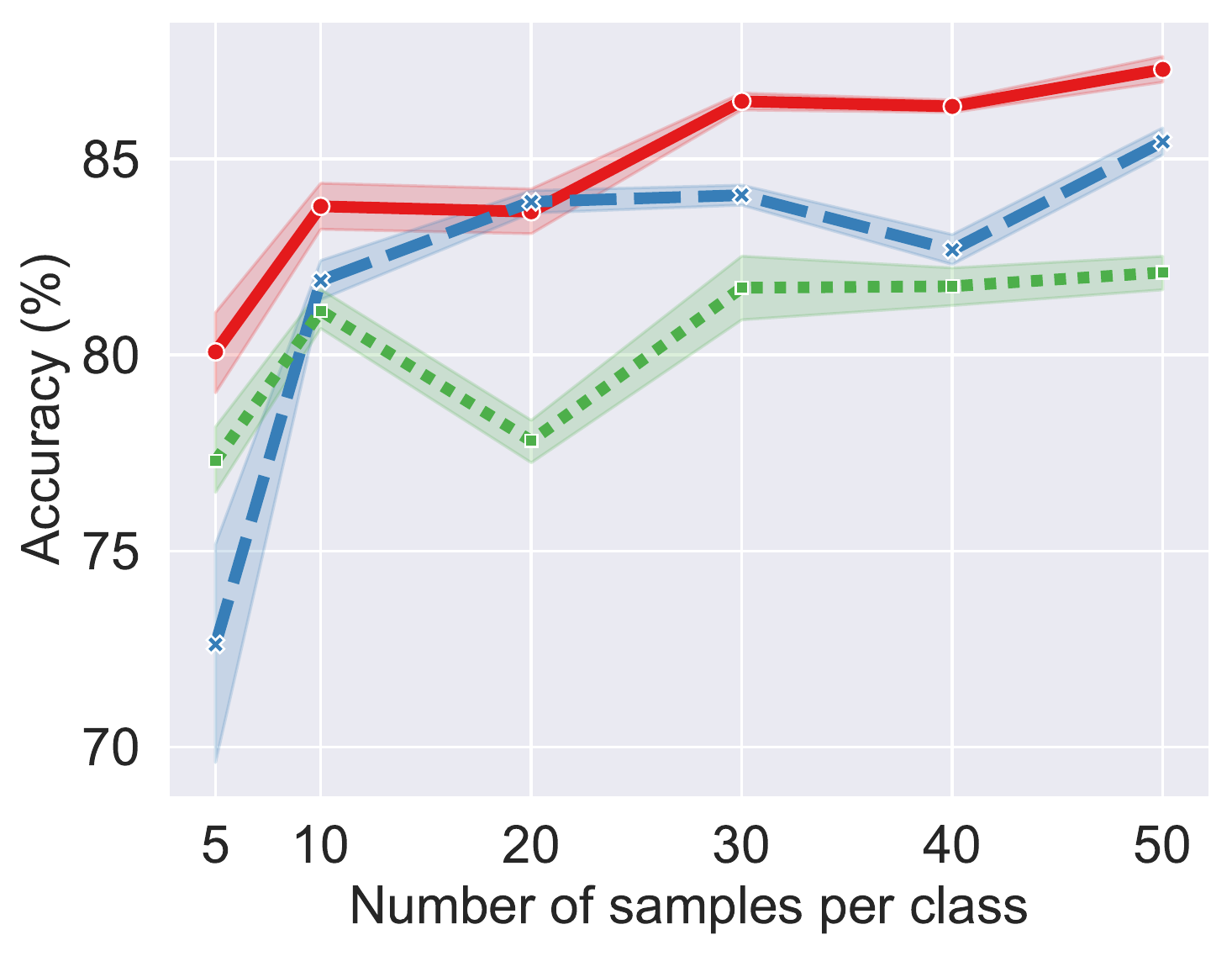}
         \caption{IMDb}

     \end{subfigure}\hfill
          \begin{subfigure}[b]{0.5\columnwidth}
         \centering
         \includegraphics[width=\columnwidth]{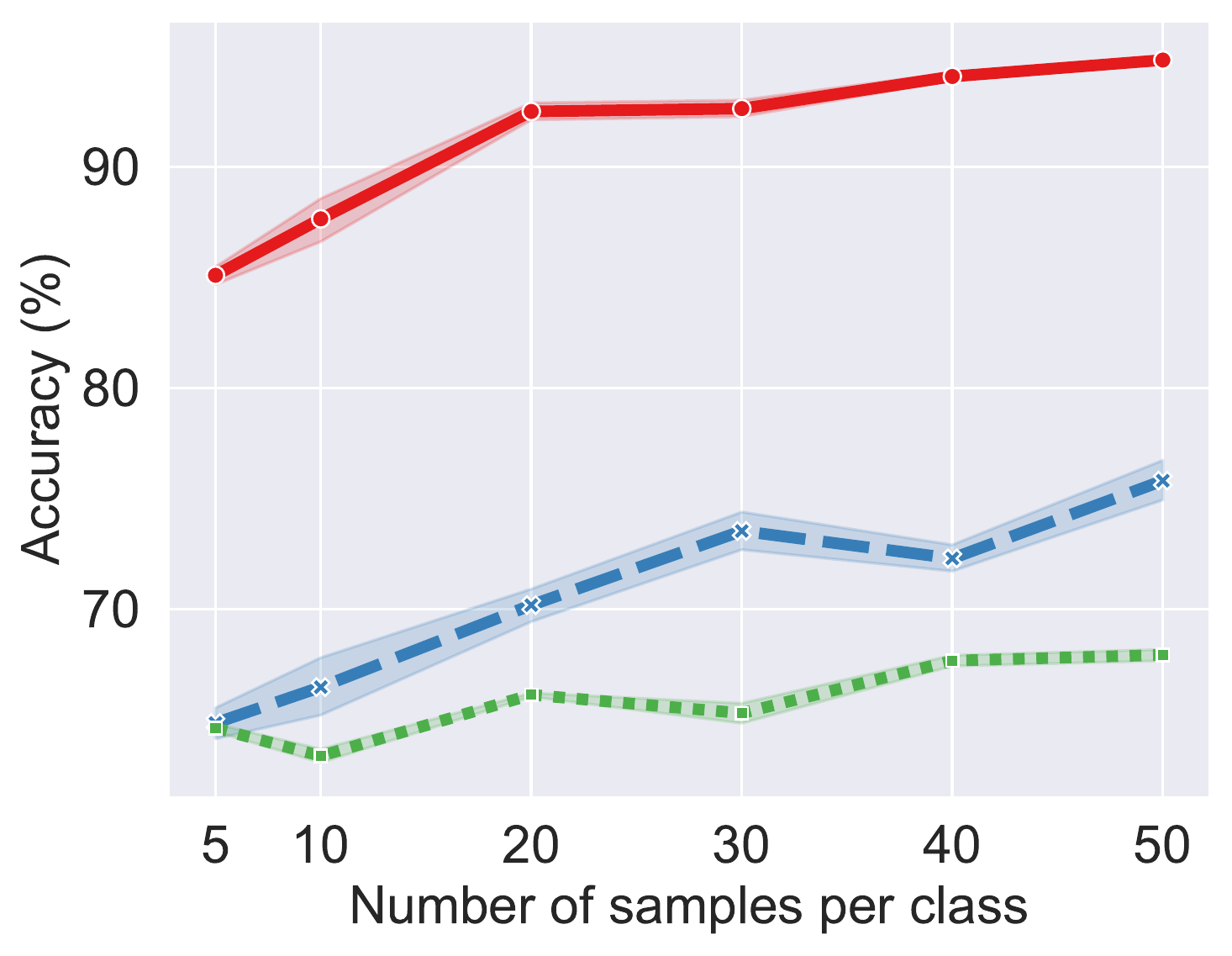}
         \caption{TREC}

     \end{subfigure}\hfill
     \begin{subfigure}[b]{0.5\columnwidth}
     \centering
     \includegraphics[width=\columnwidth]{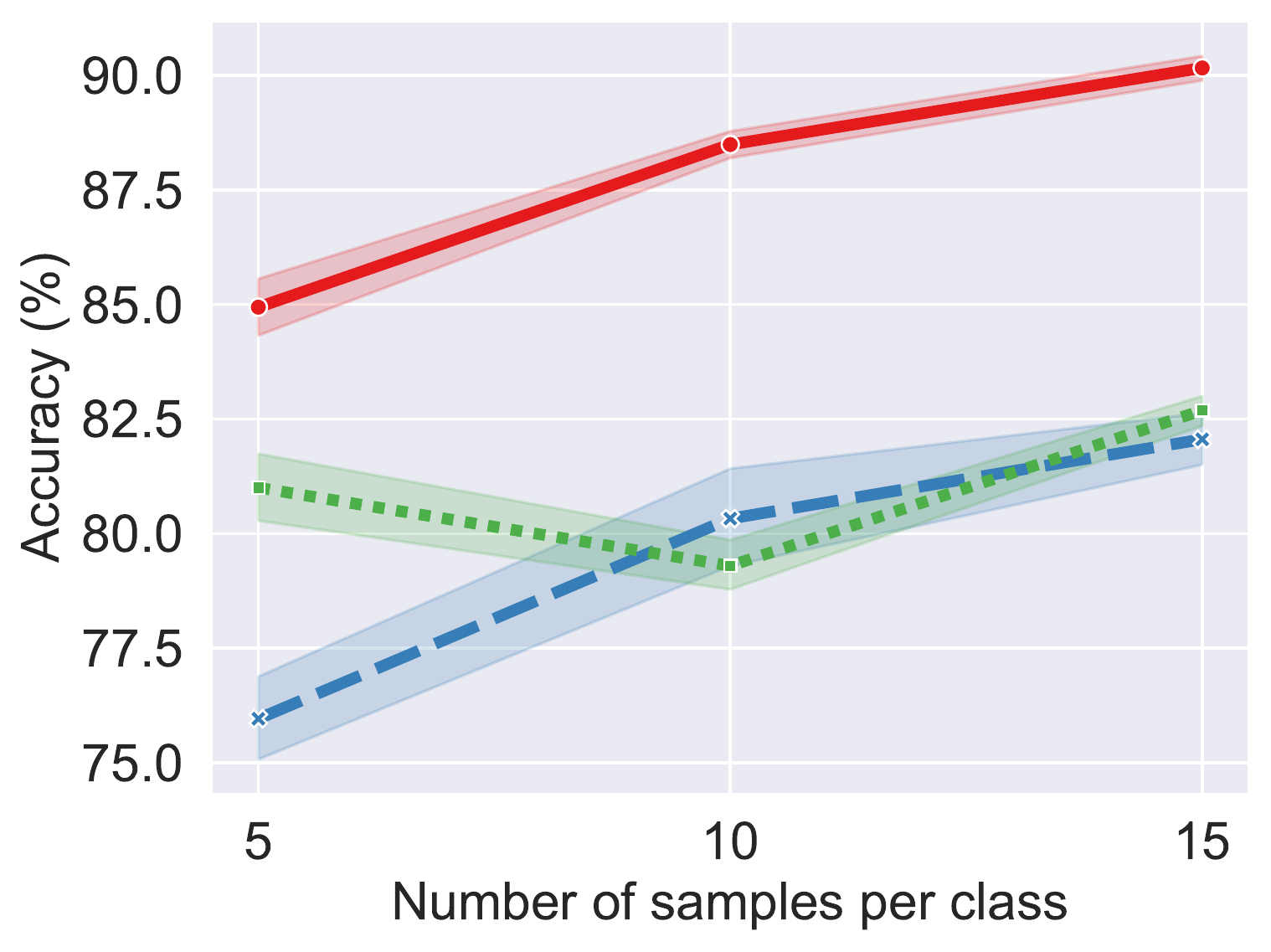}
     \caption{SemEval}

     \end{subfigure}\hfill
     \begin{subfigure}[b]{0.5\columnwidth}
     \centering
     \includegraphics[width=\columnwidth]{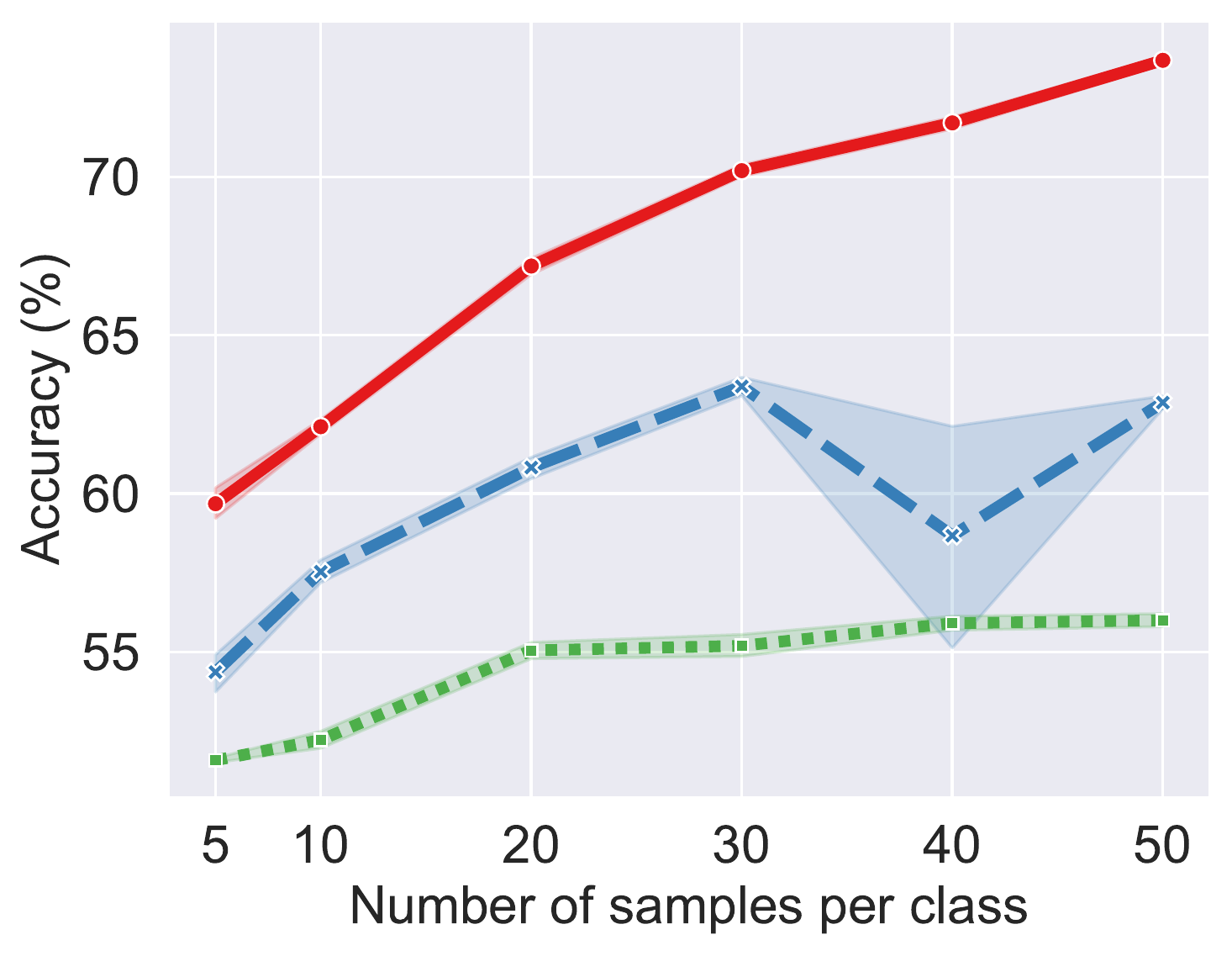}
     \caption{ChemProt}

     \end{subfigure}\hfill
     \begin{subfigure}[b]{0.5\columnwidth}
     \centering
     \includegraphics[width=\columnwidth]{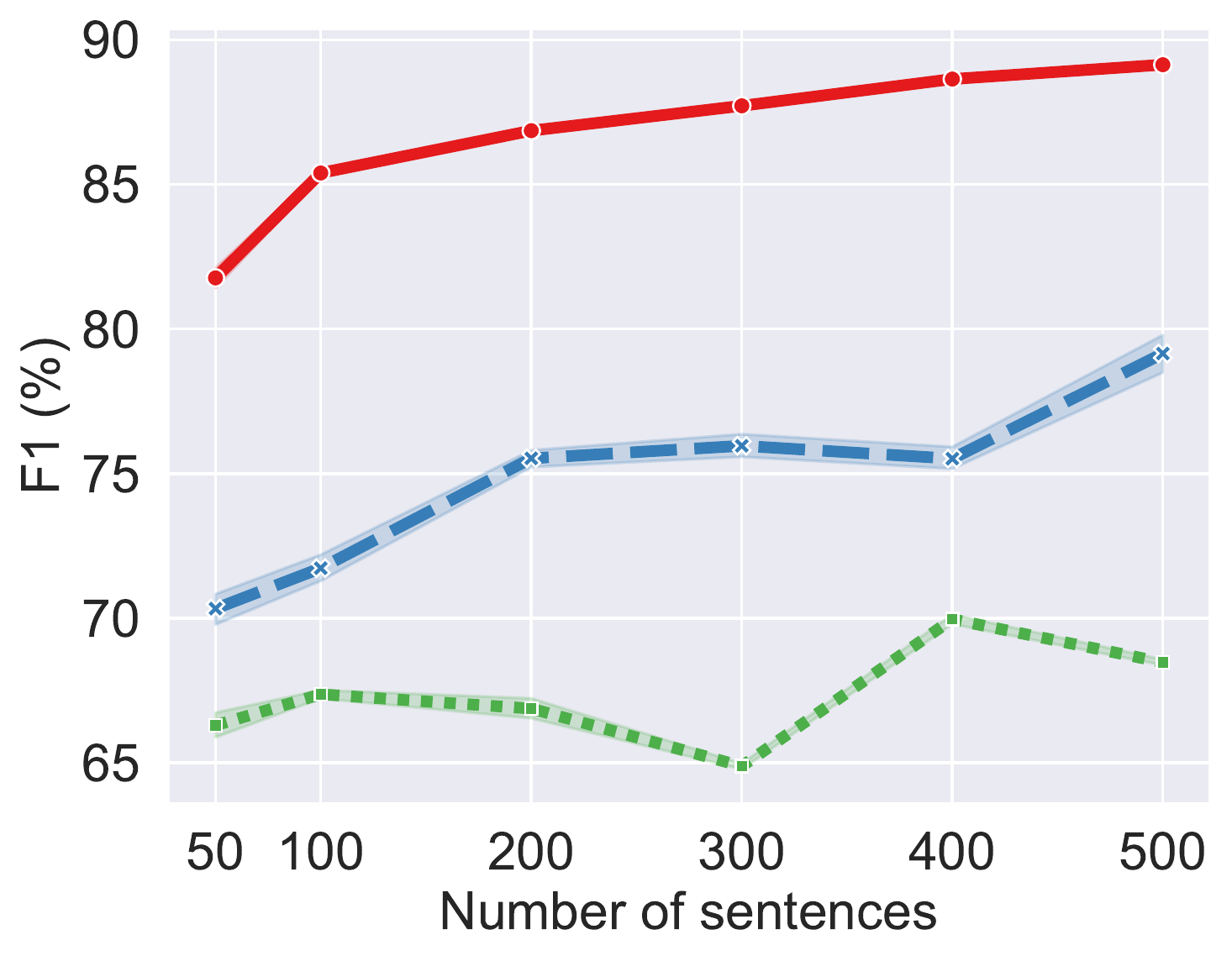}
     \caption{CoNLL-03}

     \end{subfigure}\hfill
      \begin{subfigure}[b]{0.5\columnwidth}
     \centering
     \includegraphics[width=\columnwidth]{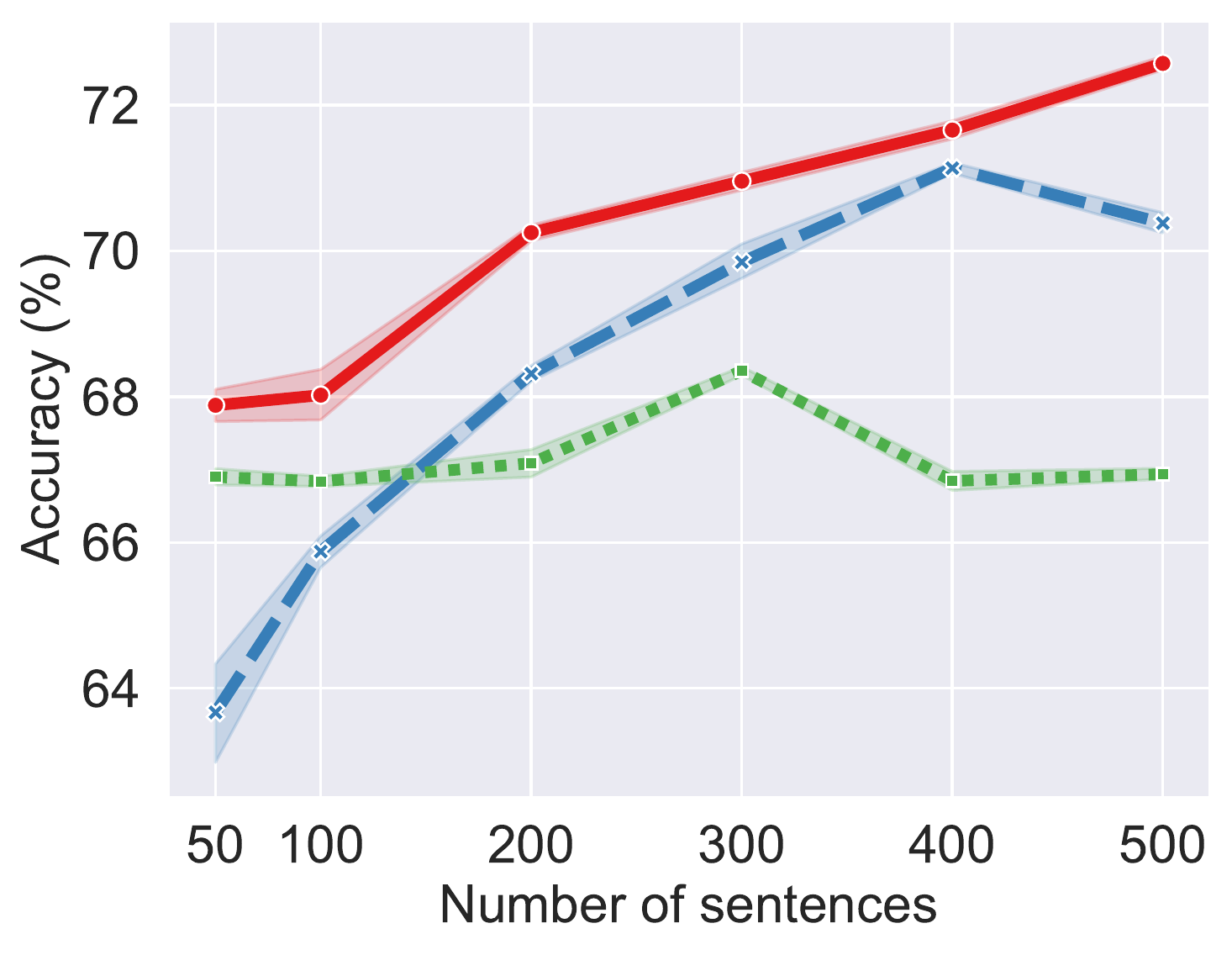}
     \caption{OntoNotes 5.0}

     \end{subfigure}\hfill
     
        \caption{Performance vs. number of clean samples. In most cases, CFT outperforms the other two baselines, WC\textsubscript{batch} and WC\textsubscript{mix}, by a considerable margin. }
        \label{fig:appendix:different_wc_methods}
\end{figure*}

\section{Additional baselines that combine weak and clean data during training}
\label{sec:appendix:further_combination_baselines}
Besides CFT we also explored two simple baselines that combine both the cleanly and weakly annotated data in training:
\begin{enumerate}
  \item \textbf{WC\textsubscript{mix}}: it mixes the clean data into the weakly labeled training set. We then fine-tune a PLM on this combined dataset.
  \item \textbf{WC\textsubscript{batch}}: in each batch, we mix the weakly and cleanly labeled data at a ratio of 50:50. This makes sure that the model can access clean samples in each batch.
\end{enumerate}
We compared these two baselines with CFT, the results are shown in Figure \ref{fig:appendix:different_wc_methods}. It can be seen that when the same amount of data is accessed, CFT outperforms the two baselines in most cases, sometimes by a large margin.

\section{Additional plots on CFT with different numbers of clean samples}
\label{sec:appendix:cft}

\begin{figure*}[ht!]
        \centering
        
     \begin{subfigure}[b]{1.99\columnwidth}
         \centering
         \includegraphics[width=\columnwidth]{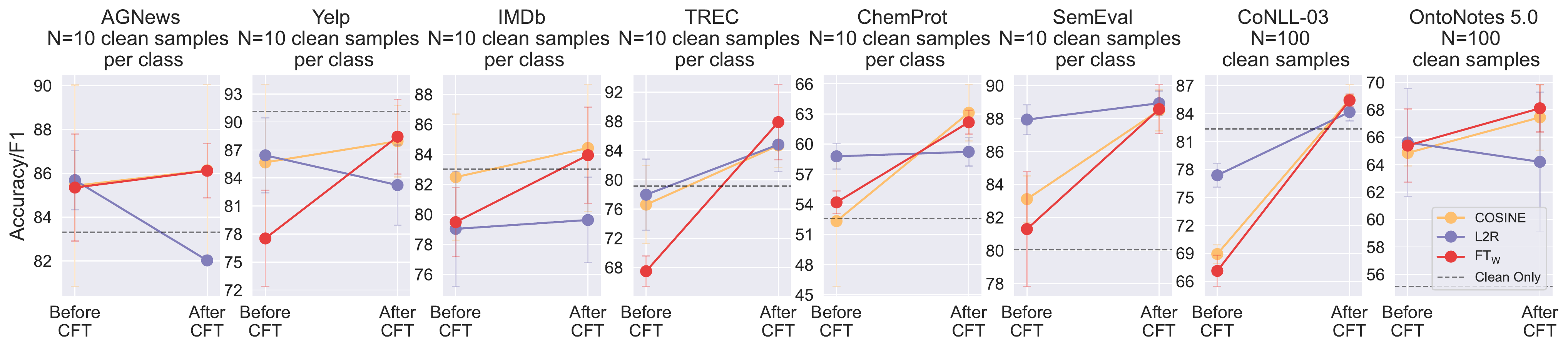}
         \caption{$N=10$ samples per class ($N=100$ sentences on NER)}
     \end{subfigure}\hfill
          \begin{subfigure}[b]{1.99\columnwidth}
         \centering
         \includegraphics[width=\columnwidth]{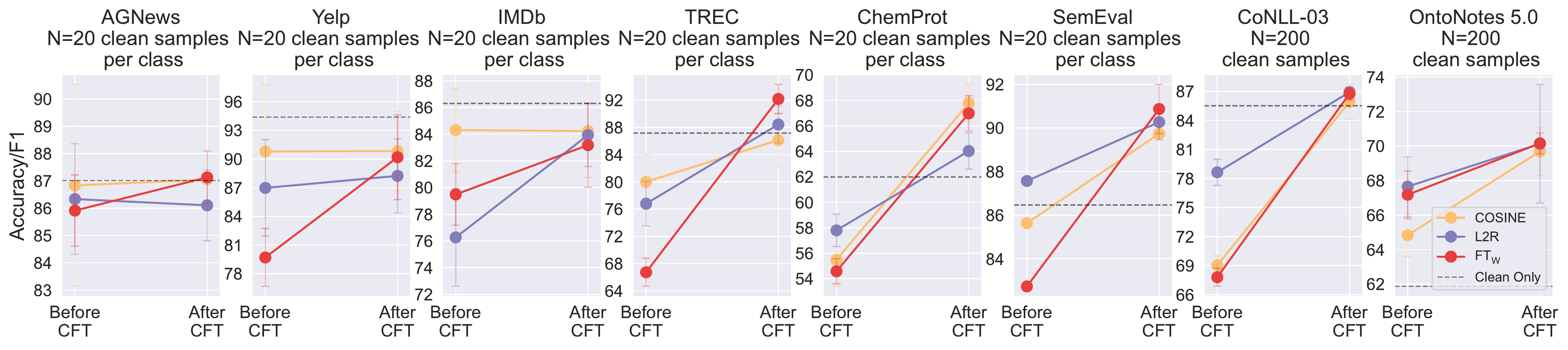}
         \caption{$N=20$ samples per class ($N=200$ sentences on NER)}

     \end{subfigure}\hfill
               \begin{subfigure}[b]{1.99\columnwidth}
         \centering
         \includegraphics[width=\columnwidth]{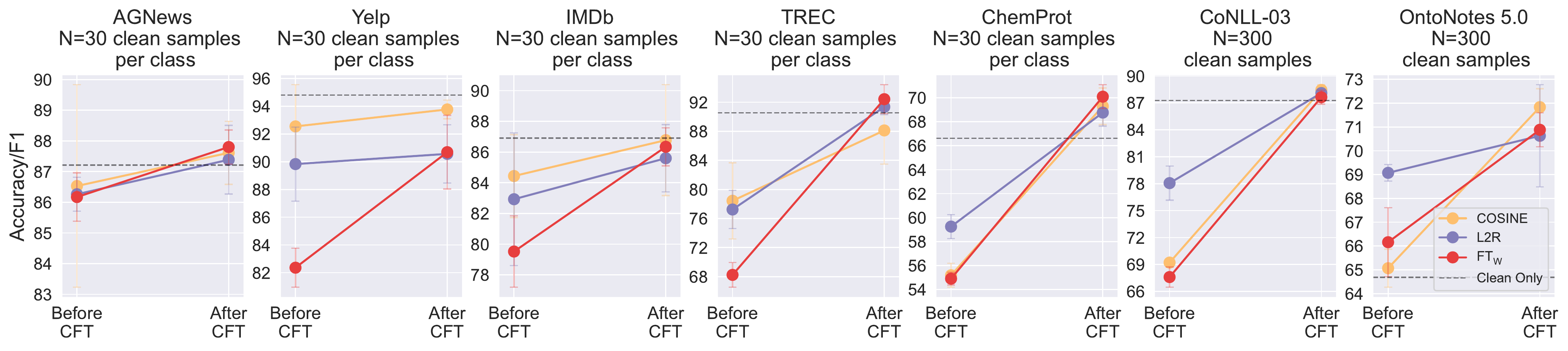}
         \caption{$N=30$ samples per class ($N=300$ sentences on NER)}

     \end{subfigure}\hfill

                \begin{subfigure}[b]{1.99\columnwidth}
         \centering
         \includegraphics[width=\columnwidth]{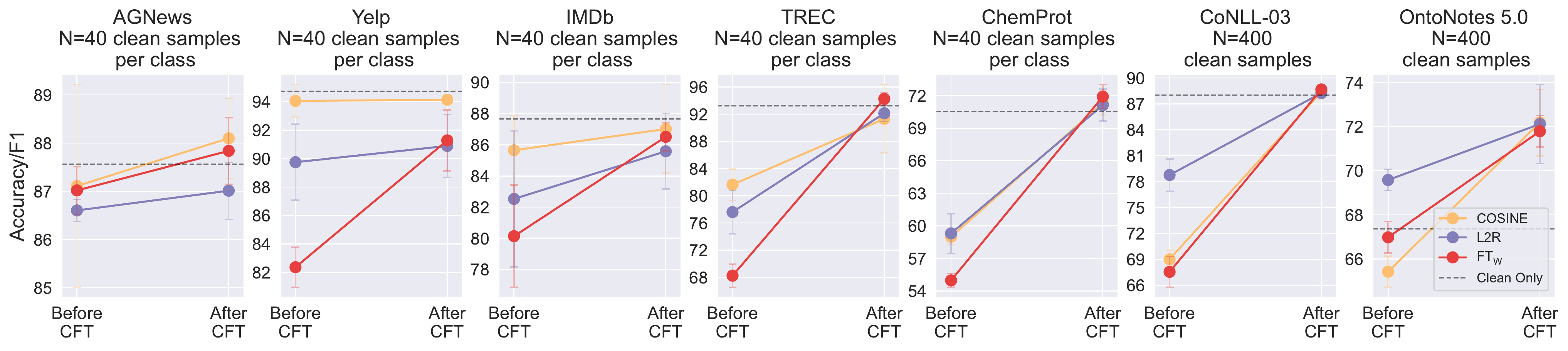}
         \caption{$N=40$ samples per class ($N=400$ sentences on NER)}

     \end{subfigure}\hfill
     
        \caption{Performance difference before and after applying CFT to WSL methods. For text classification and relation extraction tasks, we subsample $N\in \{5,10,20,30,40,50 \} $ examples from the validation set. For NER, we subsample $N\in \{50,100,200,300,400,500 \}$. On SemEval, the original validation set is small, and sampling more than $20$ samples per class is not possible. The figure shows that the performance gap between the simple baseline FT\textsubscript{W} and COSINE/L2R becomes much smaller after CFT, suggesting that we may not require sophisticated WSL methods to achieve good generalization.}
        \label{fig:apendix_wc_slope_plots}
\end{figure*}

We show further plots of experiments in Section \ref{sec:cft} with different numbers of clean samples in Figure \ref{fig:apendix_wc_slope_plots}. More specifically, it shows the results for selecting $N \in \{10, 20, 30, 40\}$ clean samples per class from the clean validation set for classification and $N \in \{100, 200,  300, 400\}$ for NER tasks. These results corroborate the analysis presented in Section \ref{sec:cft}.

\section{CFT with different PLMs and agreement ratios}
\label{sec:appendix:cft_abalation}
\begin{figure*}[ht!]
        \centering
        
     \begin{subfigure}[b]{1\columnwidth}
         \centering
         \includegraphics[width=\columnwidth]{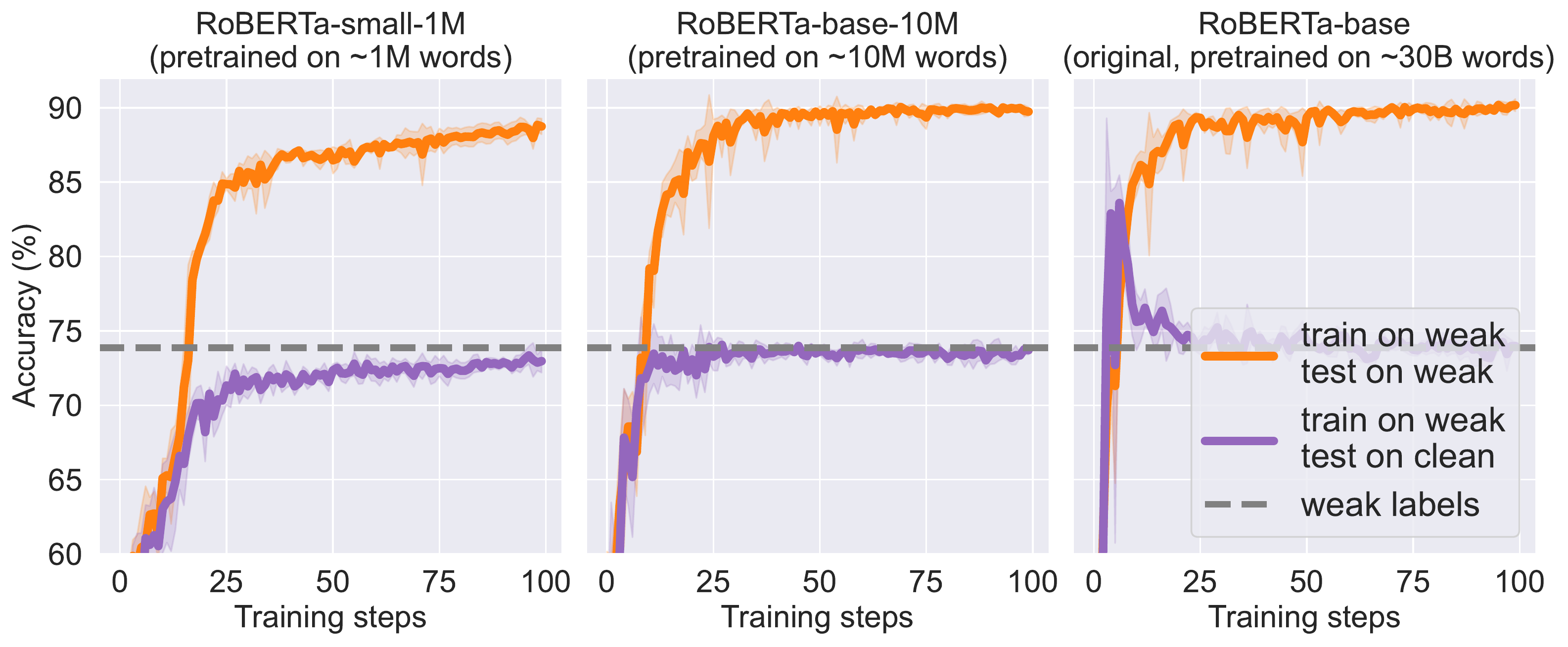}
         \caption{IMDb}
     \end{subfigure}\hfill
          \begin{subfigure}[b]{1\columnwidth}
         \centering
         \includegraphics[width=\columnwidth]{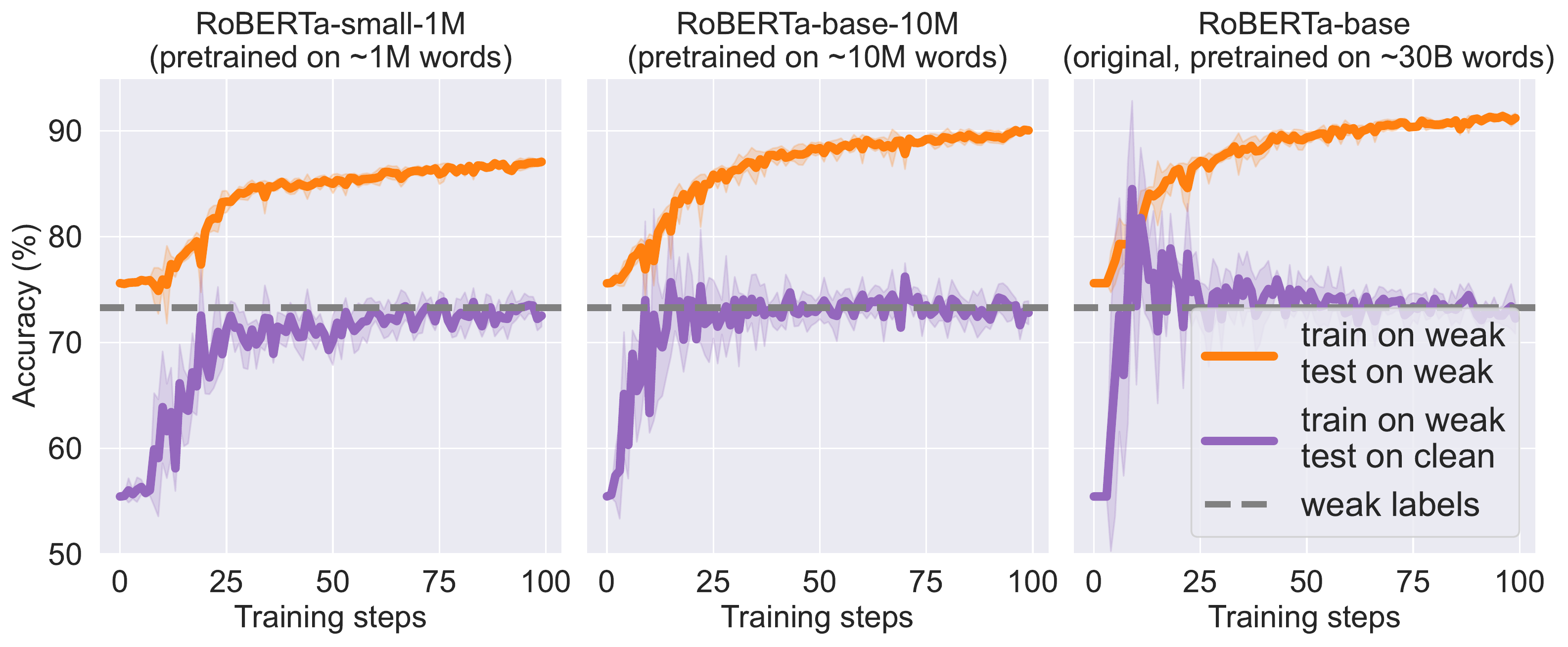}
         \caption{Yelp}
     \end{subfigure}\hfill
          \begin{subfigure}[b]{1\columnwidth}
         \centering
         \includegraphics[width=\columnwidth]{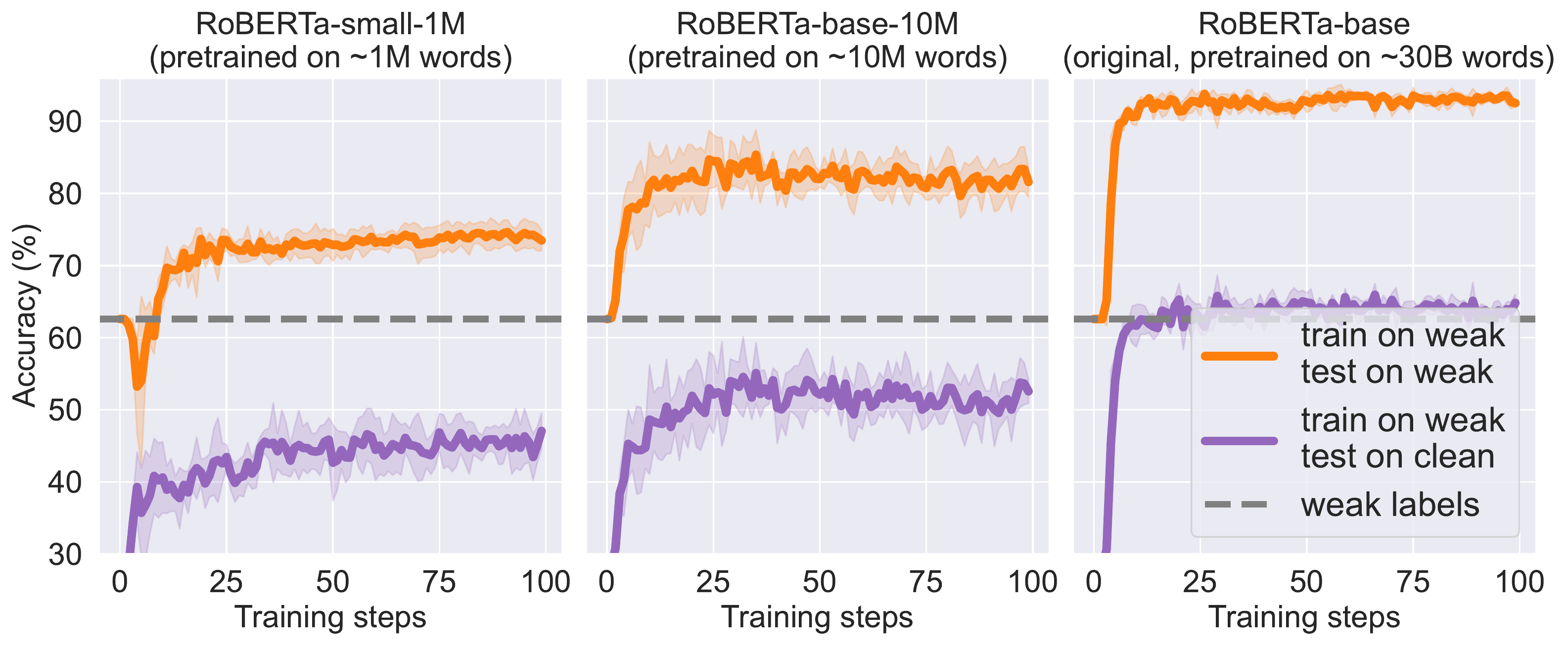}
         \caption{TREC}
     \end{subfigure}\hfill
          \begin{subfigure}[b]{1\columnwidth}
         \centering
         \includegraphics[width=\columnwidth]{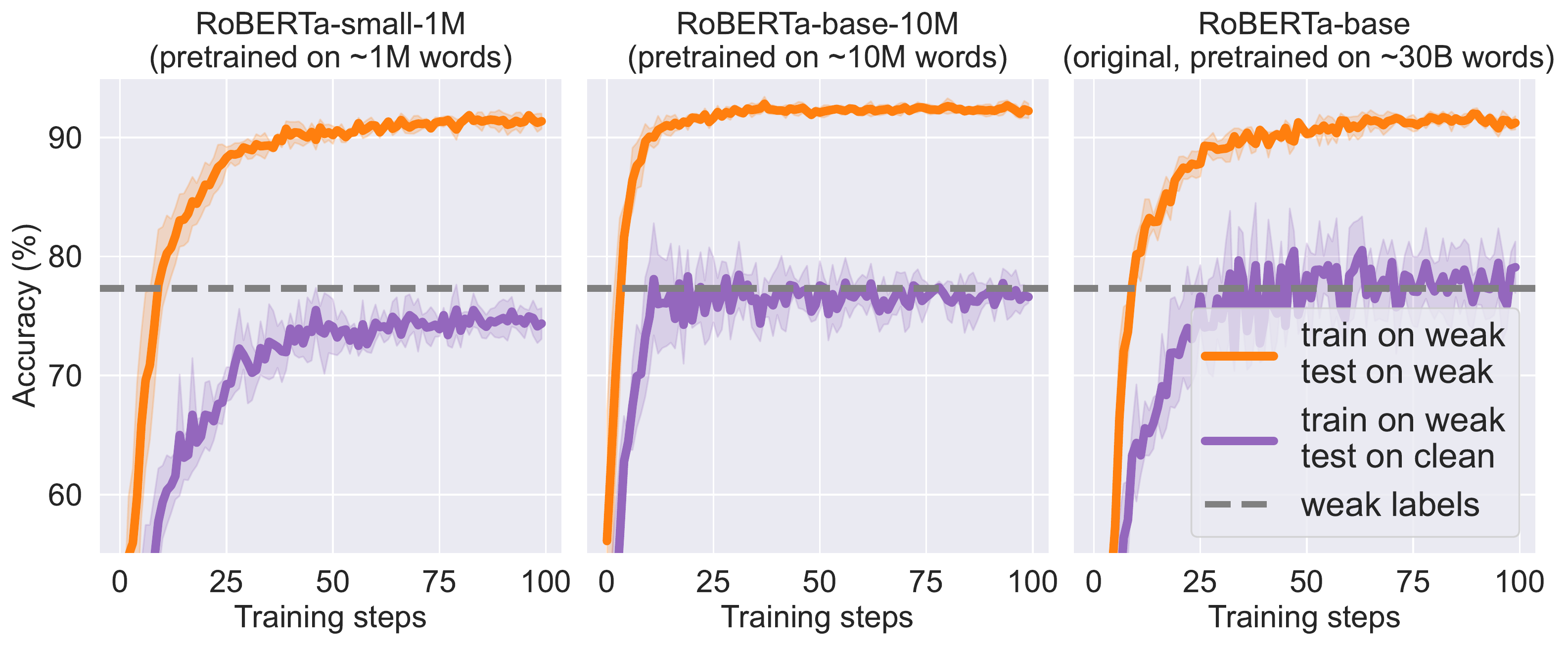}
         \caption{SemEval}
     \end{subfigure}\hfill
           \begin{subfigure}[b]{1\columnwidth}
         \centering
         \includegraphics[width=\columnwidth]{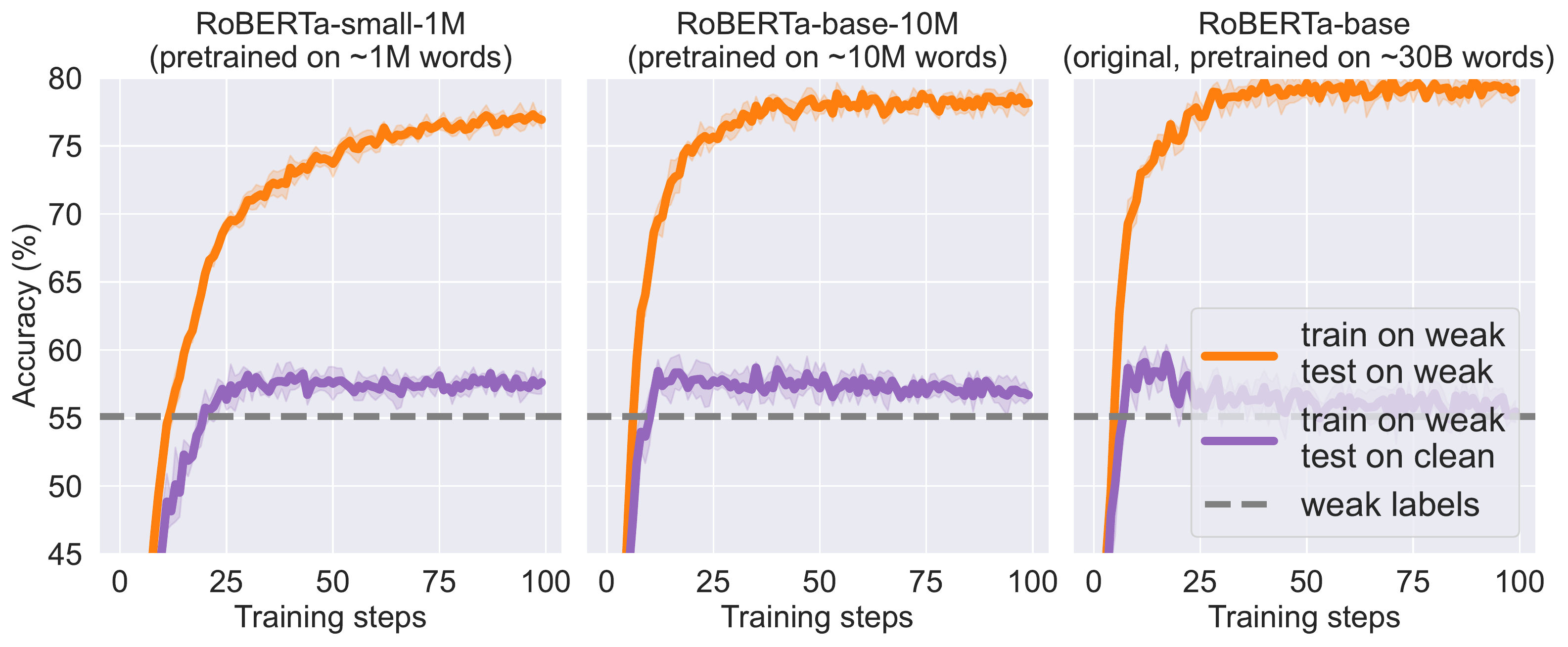}
         \caption{ChemProt}
     \end{subfigure}\hfill
        \begin{subfigure}[b]{1\columnwidth}
         \centering
         \includegraphics[width=\columnwidth]{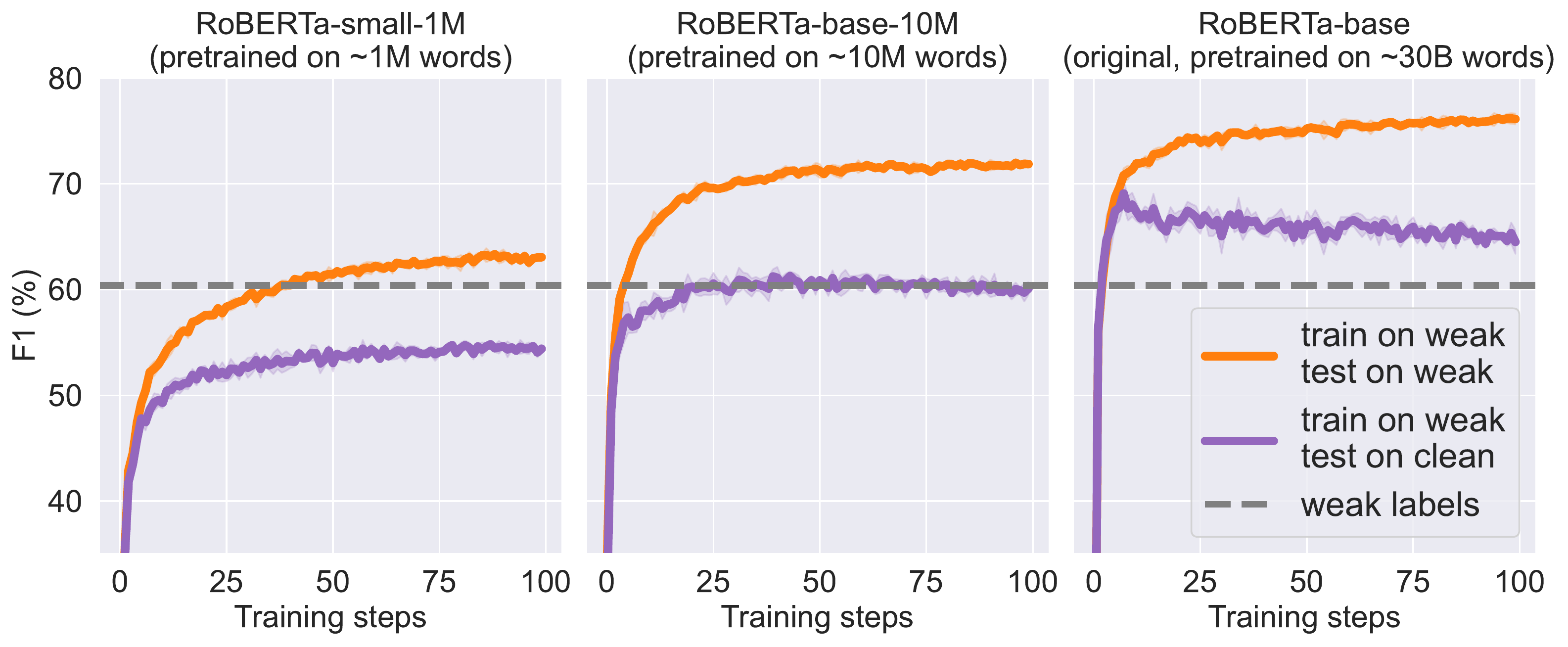}
         \caption{CoNLL-03}
     \end{subfigure}\hfill
     
        \caption{\textbf{Performance curves of different PLMs during training}. PLMs are trained on weak labels and evaluated on both clean and weakly labeled test sets. Pre-training on larger corpora improves performance on the clean distribution.}
        \label{fig:plm_dff_appendix}
\end{figure*}

\begin{figure}[ht!]
        
      \begin{subfigure}[b]{0.5\columnwidth}
     \centering
     \includegraphics[width=\columnwidth]{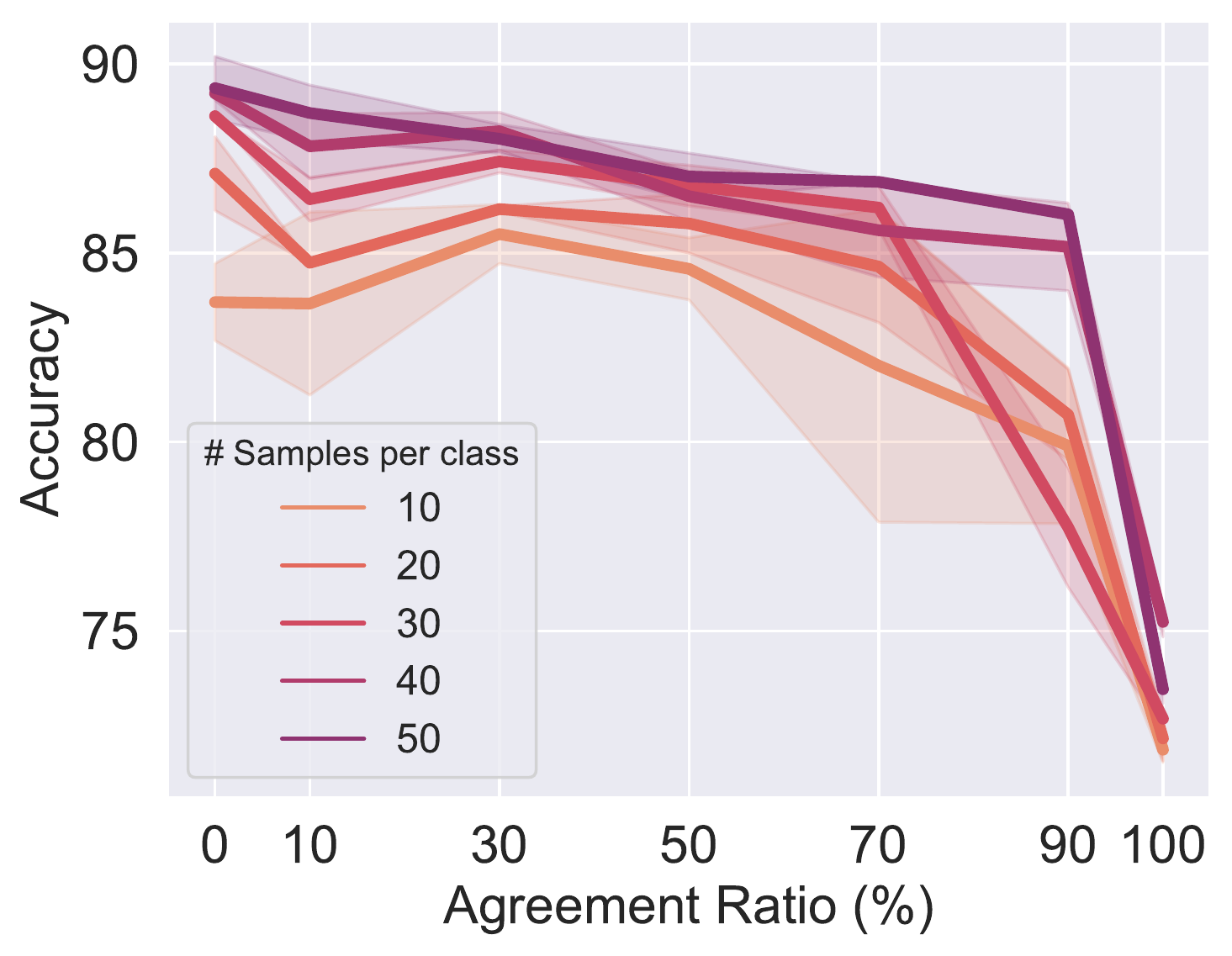}
     \caption{IMDb}

     \end{subfigure}\hfill
     \begin{subfigure}[b]{0.5\columnwidth}
     \centering
     \includegraphics[width=\columnwidth]{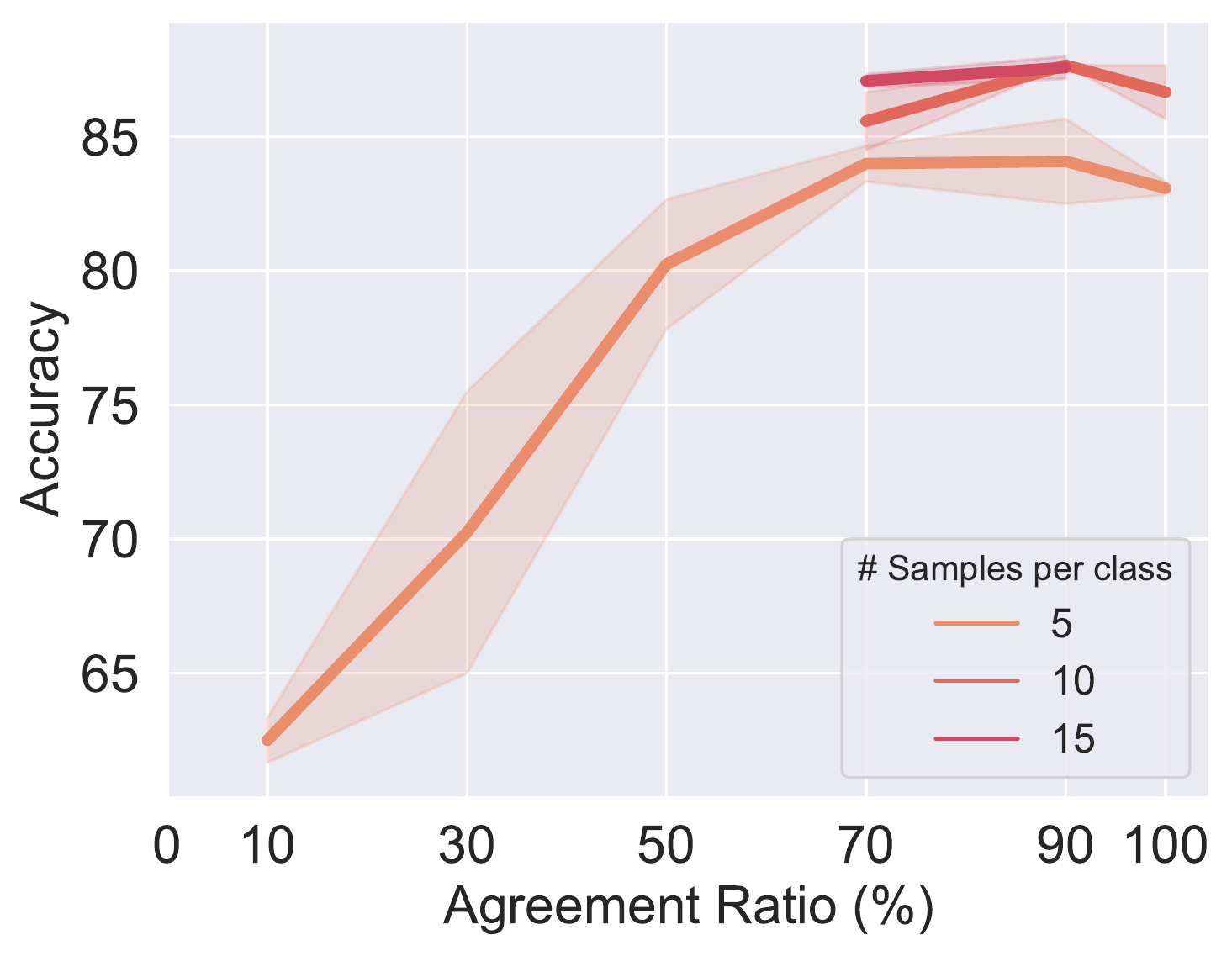}
     \caption{SemEval$^*$}

     \end{subfigure}\hfill
     \begin{subfigure}[b]{0.5\columnwidth}
     \centering
     \includegraphics[width=\columnwidth]{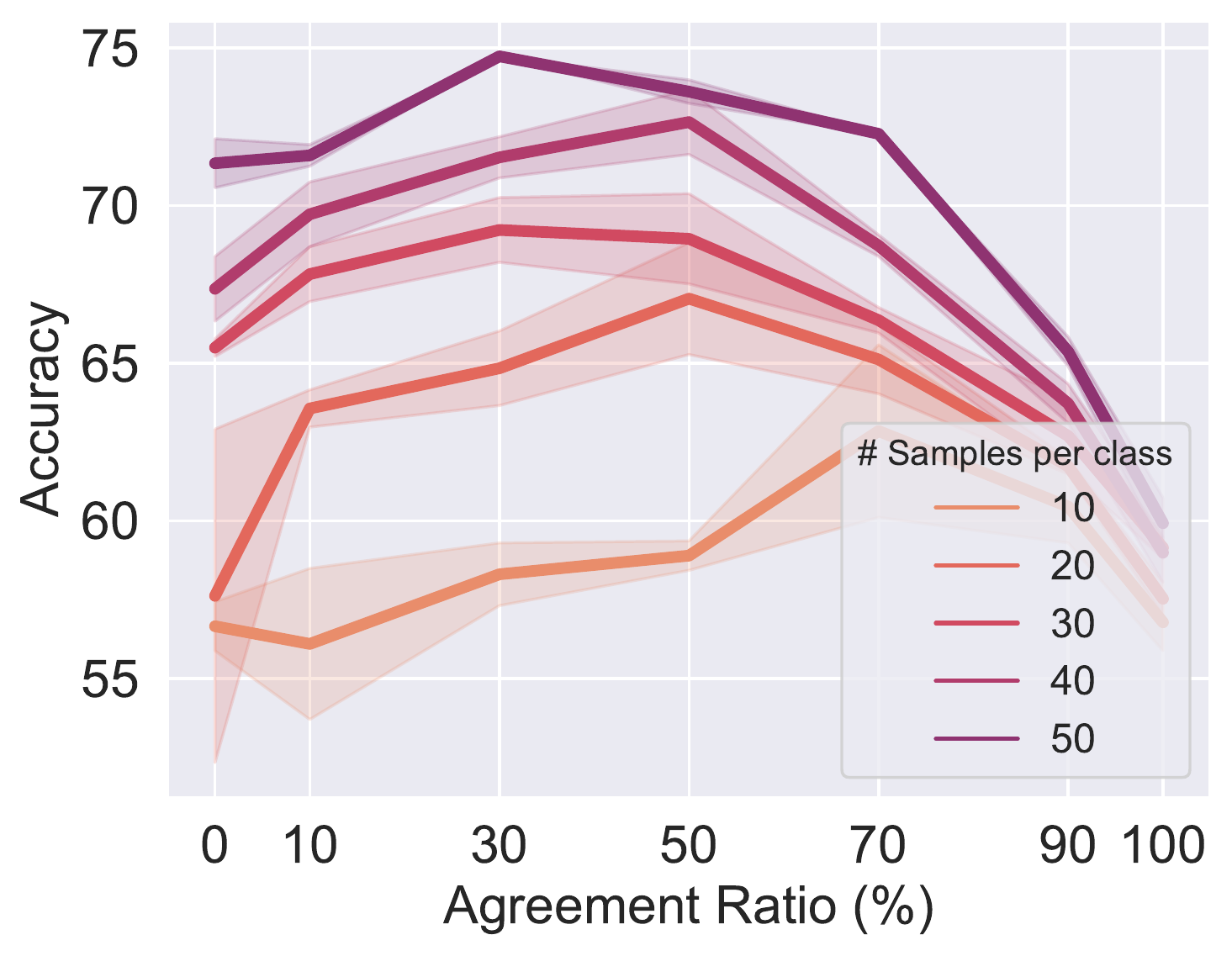}
     \caption{ChemProt}

     \end{subfigure}\hfill
     \begin{subfigure}[b]{0.5\columnwidth}
     \centering
    \includegraphics[width=\columnwidth]{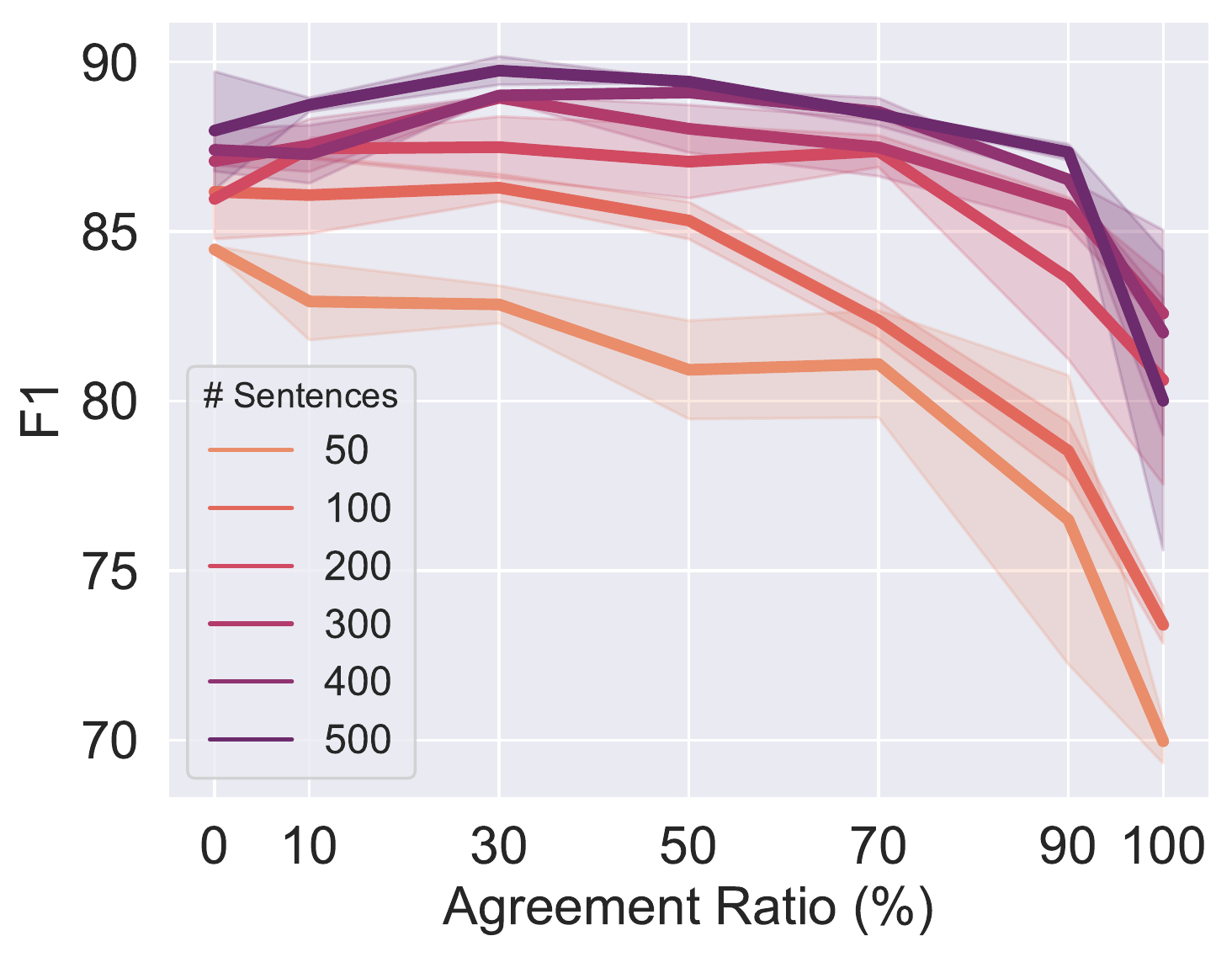}
     \caption{CoNLL-03}

     \end{subfigure}\hfill

        \caption{\textbf{Model performance varying the number of clean samples $N$ and agreement ratio $\alpha$}. Large values of $\alpha$ generally cause a substantial performance drop. $^*$: Certain combinations of $\alpha$ and $N$ are not feasible because the validation set lacks samples with clean and weak labels that coincide or differ.}
        \label{fig:cft_ablation_appendix}
\end{figure}

We provide additional plots of the experiments mentioned in Section \ref{sec:cft_ablation} on more datasets. Figure \ref{fig:plm_dff_appendix} shows the performance of CFT using different PLMs during training and Figure \ref{fig:cft_ablation_appendix} shows the performance when the number of clean samples and the agreement ratio is varied.

\end{document}